\documentclass[10pt,journal,compsoc]{IEEEtran}

\usepackage{ragged2e}
\usepackage{url}
\usepackage{subcaption}
\usepackage{graphicx}
\usepackage{multirow}
\usepackage{bbm}
\usepackage{amsmath}
\usepackage{xcolor}
\usepackage[pagebackref=true,breaklinks=true,colorlinks,urlcolor=blue,citecolor=blue,linkcolor=blue,bookmarks=false]{hyperref}

\makeatletter
\def\hlinew#1{%
	\noalign{\ifnum0=`}\fi\hrule \@height #1 \futurelet
	\reserved@a\@xhline}
\makeatother%

\ifCLASSOPTIONcompsoc
\usepackage[nocompress]{cite}
\else
\usepackage{cite}
\fi

\ifCLASSINFOpdf

\else

\fi

\begin{document}
	
	\title{Towards Unified Deep Image Deraining: \\A Survey and A New Benchmark}
	
	\author{Xiang Chen, Jinshan Pan,
		Jiangxin Dong, and Jinhui Tang
		\IEEEcompsocitemizethanks{\IEEEcompsocthanksitem X. Chen, J. Pan, J. Dong, and J. Tang are with the School of Computer Science and Engineering, Nanjing University of Science and Technology, Nanjing, 210094, China.\protect ~
			E-mail: chenxiang@njust.edu.cn, sdluran@gmail.com, dongjxjx@gmail.com, jinhuitang@njust.edu.cn.
		}
		\IEEEcompsocitemizethanks{\IEEEcompsocthanksitem Corresponding author: Jinshan Pan.
		}
	}
	
	\markboth{}%
	{Shell \MakeLowercase{\textit{et al.}}: Bare Demo of IEEEtran.cls for Computer Society Journals}
	
	\IEEEtitleabstractindextext{%
		\begin{abstract}
			\justifying
			Recent years have witnessed significant advances in image deraining due to the kinds of effective image priors and deep learning models.
			As each deraining approach has individual settings (e.g., training and test datasets, evaluation criteria), how to fairly evaluate existing approaches comprehensively is not a trivial task.
			Although existing surveys aim to review of image deraining approaches comprehensively, few of them focus on providing unify evaluation settings to examine the deraining capability and practicality evaluation.
			In this paper, we provide a comprehensive review of existing image deraining method and provide a unify evaluation setting to evaluate the performance of image deraining methods.
			We construct a new high-quality benchmark named HQ-RAIN to further conduct extensive evaluation, consisting of 5,000 paired high-resolution synthetic images with higher harmony and realism.
			We also discuss the existing challenges and highlight several future research opportunities worth exploring.
			To facilitate the reproduction and tracking of the latest deraining technologies for general users, we build an online platform to provide the off-the-shelf toolkit, involving the large-scale performance evaluation.
			This online platform and the proposed new benchmark are publicly available and will be regularly updated at \url{http://www.deraining.tech/}.
		\end{abstract}
		
		}
	
	\maketitle
	
	\IEEEdisplaynontitleabstractindextext
	
	\IEEEpeerreviewmaketitle
	
	\IEEEraisesectionheading{\section{Introduction}
		\label{sec:introduction}}
	
	\IEEEPARstart{R}{ain} is a challenging adverse weather condition which always obstructs or deforms the image background.
	It consequently degrades the visual quality of images and impairs the performance of outdoor vision-based systems, including video surveillance \cite{sultani2018real}, autonomous driving \cite{kiran2021deep} and person identification \cite{ye2021deep}, etc.
	Single image deraining, also known as rain removal, aims to recover the rain-free image from the observed rainy one.
	A large number of deraining methods and benchmark datasets have been proposed in recent years with demonstrated success.
	Since this topic booms with the deep learning techniques \cite{wang2019survey,li2019single,yang2020single,zhang2023data}, it is significant to timely and comprehensively evaluate the state-of-the-art approaches to demonstrate their strength and weakness, thus facilitating the future development of this research field for more robust algorithms.
	
	\begin{figure*}[!t]
		\centering 	
		\begin{subfigure}[t]{0.65\textwidth}
			\centering
			\includegraphics[width=\textwidth]{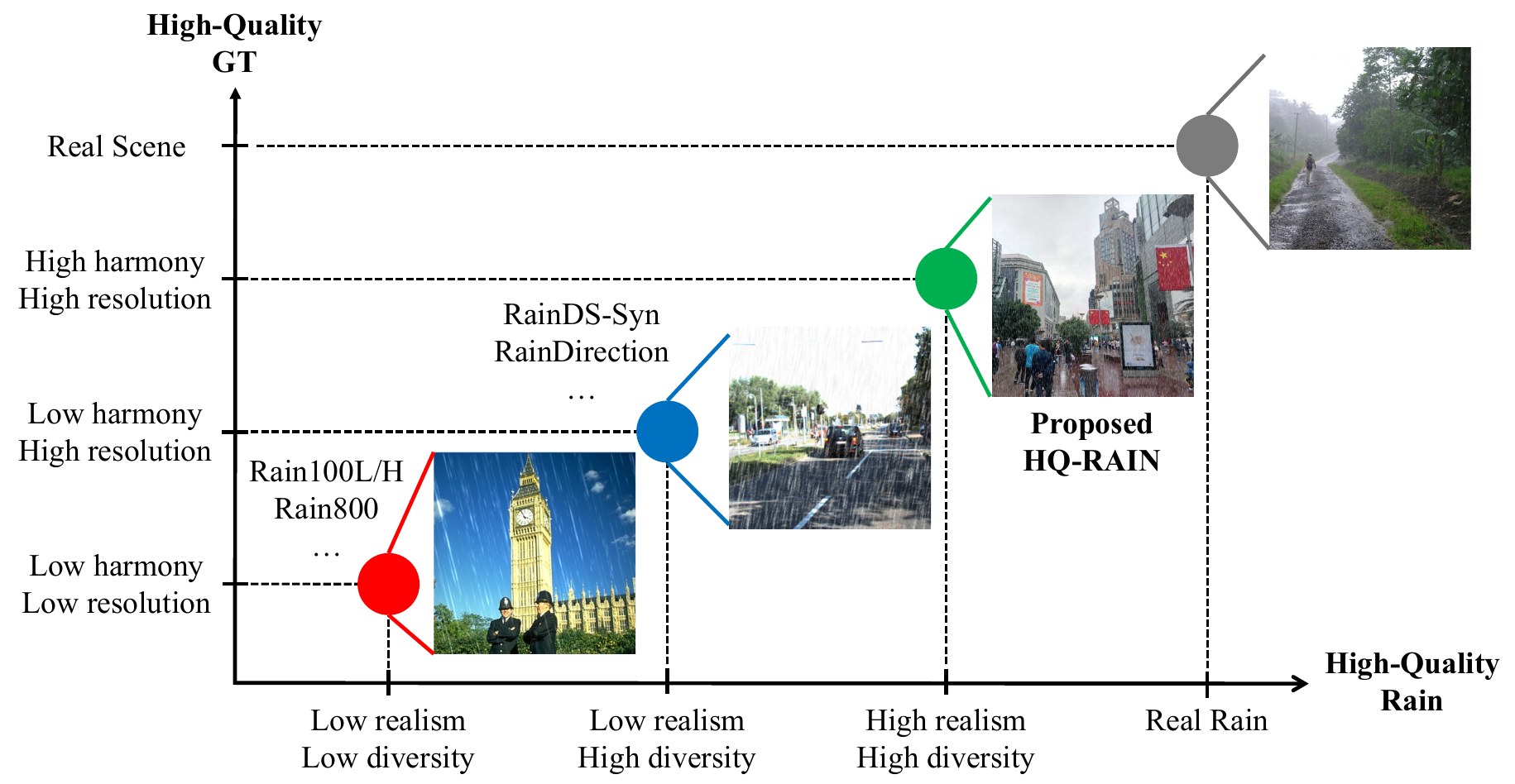}
			\caption{Analysis of Benchmark Characteristic}
		\end{subfigure}
		\begin{subfigure}[t]{0.34\textwidth}
			\centering
			\includegraphics[width=\textwidth]{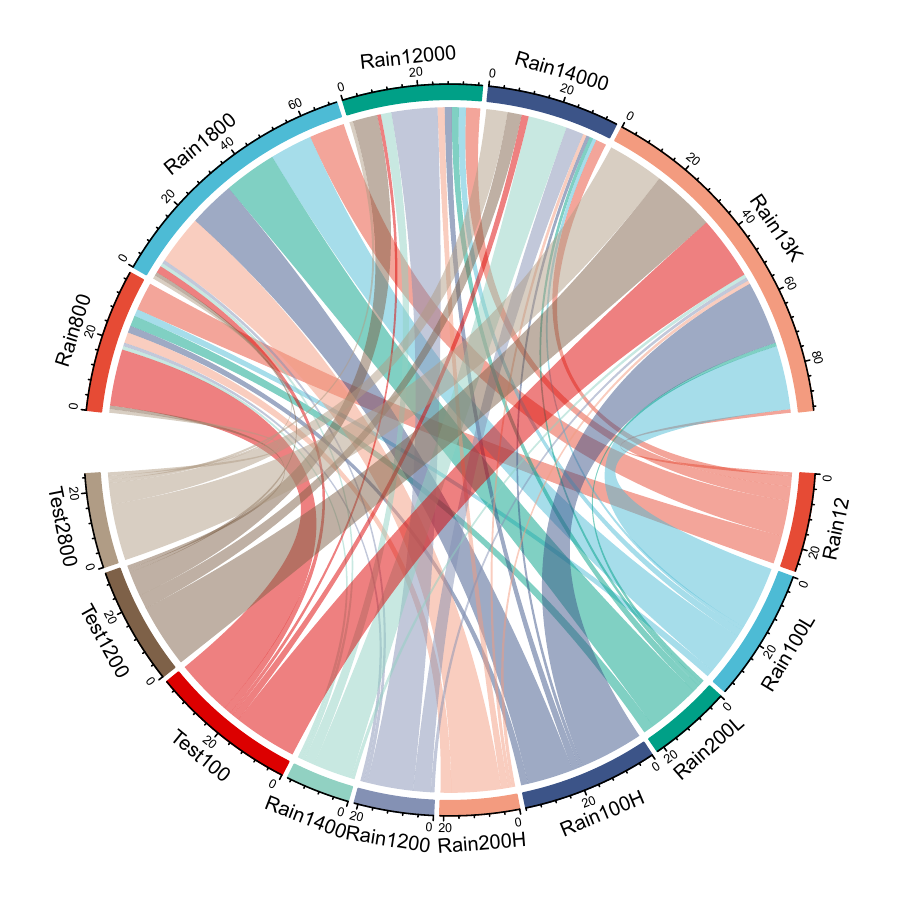}
			\caption{Analysis of Benchmark Usage}
		\end{subfigure}
		\caption{Analysis of commonly used synthetic benchmark characteristic and usage based on over a hundred literatures about image deraining. (a) Our proposed new benchmark narrows the domain gap between real rainy images, due to its properties of high-quality rain with higher realism and diversity, and high-quality GT with higher harmony and resolution. (b) Statistics on the usage relationship between training and test datasets in their experiments with existing methods. The upper semicircle means training datasets, and the lower semicircle means testing datasets.}
		\label{fig1}
	\end{figure*}
	
	To facilitate the performance comparison and analysis of existing methods, it is indispensable to collect a representative benchmark dataset.
	There exist several classical synthetic datasets for single image deraining task, such as the Rain100L \cite{yang2017deep}, Rain100H \cite{yang2017deep}, and Rain800 \cite{li2019single}.
	Although these synthetic datasets are prevalent in current studies \cite{chen2023learning,fu2023continual,xiao2022image,yi2021structure,fu2021rain,deng2020detail}, most approaches randomly add rain effects to any clear backgrounds to build the synthetic paired data, which inevitably leads to inadequate realism and harmony, especially in the sky region (see Figure~\ref{fig1}(a)).
	As a result, there is a large domain gap between synthetic and real data, which limits the ability of training better deep deraining models.
	Recently, significant efforts have been made to solve real-world image deraining, where the large-scale datasets are proposed, \emph{e.g}, SPA-Data \cite{wang2019spatial} and GT-RAIN \cite{ba2022not}.
	Unfortunately, these real datasets still have several shortcomings, including low resolution, spatial misalignment, limited rain appearances and background variation \cite{li2022toward}.
	In fact, obtaining strictly aligned rainy and clear image pairs in complex rainy environments is time-consuming and laborious.
	Returning to the problem itself, it is urgent and necessary to establish a high-quality benchmark and pave the way for future research in this field.
	
	Despite lots of efforts have been made on existing benchmarks\cite{wang2019survey,li2019single,yang2020single,li2022toward, zhang2023data}, there still exist several potential challenges.
	According to statistics in Figure~\ref{fig1}(b), existing methods usually use individual evaluation settings to evaluate the deraining performance. The evaluations based on different settings do not fully reflect performance to some extent.
	On one side, multiple training sets are trained in a mixed way, and a pretrained model is obtained to evaluate different testing sets.
	The typical mixed training based methods include MSPFN \cite{jiang2020multi}, MPRNet \cite{zamir2021multi}, Restormer \cite{zamir2022restormer}.
	On the other side, each training set is trained independently, and multiple pretrained models are obtained to evaluate the corresponding testing sets.
	The typical independent training based methods include SPDNet \cite{yi2021structure}, IDT \cite{xiao2022image}, DRSformer \cite{chen2023learning}.
	Under the setting of independent training, the usage relationships between training data and testing data are also quite diverse, especially for Rain100L/H and its new version, Rain200L/H \cite{yang2017deep}.
	As a result, this makes it difficult to make comparisons among the reported quantitative results.
	In addition, the validation metrics of the evaluated deraining algorithms are not the same, such as calculating PSNR and SSIM metric \cite{huynh2008scope} in different channels (RGB or Y), and converting YCbCr space by MATLAB or OpenCV tool in different ways.
	This naturally leads to incomparable and sometimes contradictory quantitative results reported in their literature.
	Hence, it is of importance to integrate them in a unified track for evaluation on a common criteria.

	Towards these goals, we first establish the latest development milestone of single image deraining field, especially with the increasing number of solutions in recent years.
	Note that this paper only focuses on image deraining, excluding video deraining.
	Based on this survey, we intend to summarize two tracks (\emph{i.e.}, mixed training and independent training) to report the performance of existing representative methods within the same evaluation criteria.
	It can provide a guideline on fair and reasonable comparisons of different methods in future work for researchers, especially for beginners who are new to this filed.
	We suggest that researchers can first try to categorize their methods into the corresponding track, and then pay special attention to evaluating and comparing previous methods that belong to the same track.
	Furthermore, we also construct a new high-quality image deraining benchmark, called HQ-RAIN, to further facilitate the performance comparison.
	To the best of our knowledge, this is the first attempt to improve the realism and harmony of synthetic rainy images, with the aim of reducing the domain gap between the synthetic and real data.
	Inspired by image harmonization \cite{cong2020dovenet}, we can treat a real rainy image as harmonized image, and further generate a synthetic rainy dataset by background collection, rain streaks synthesis and image blending, making it compatible with the background area in the synthetic image.
	In addition to reviewing and analyzing the existing
	methods, we also develop a unified online platform to facilitate the performance evaluation.
	To keep pace with the fast development, this platform releases a public repository to provide an up-to-date record of recent advancements, which can be served as a reference point for new researchers.
	The online platform and the proposed new benchmark
	can be found at \url{http://www.deraining.tech}.
	
	\subsection{Our Contributions}
	The contribution of this paper are mainly three-fold. (1) We present a systematic and comprehensive review of recent advancements in single image deraining field.
	(2) We construct a new high-quality image deraining banchmark (called HQ-RAIN) to inspire further researches for more robust algorithms.
	(3) We provide an online platform to offer a fair and unified performance evaluation for deep image deraining methods.
	
	\subsection{Relations with Other Surveys}
	Despite the fact that deep learning has dominated image deraining research, a timely and thorough survey on the latest technologies is lacking.
	There are two early surveys of image deraining \cite{wang2019survey, yang2020single} that mainly review prior-based and CNN-based methods.
	In comparison to recent survey \cite{zhang2023data} that reviews current methods from different perspectives, this paper seeks to provide much more than a summary of recent research progress.
	Although these surveys have provided comprehensive analyses of existing datasets, they still overlook two key factors when evaluting existing algorithms: synthetic dataset quality and common evaluation criteria.
	To the best of our knowledge, these two issues have not been well addressed in the literatures of image deraining.
	Our survey will fill the gap in this research.
	
	\subsection{Organization}
	The rest of this paper is structured as follows. In Section~\ref{sec:problem}, we present the problem formulations of some common rainy image models. Section~\ref{sec:review} provides a systematic review of representative image deraining methods. Section~\ref{sec:benchmark} introduces our proposed new benchmark. In Section~\ref{sec:performance}, we conduct extensive performance evaluations and analyses for representative methods. Section~\ref{sec:challengs} discusses the open challenges and new perspectives for future studies. Finally, the concluding remarks will be drawn in Section~\ref{sec:conclusion}.
	
	\section{Problem Formulation}
	\label{sec:problem}
	In this section, we review several representative rain models proposed in the existing methods. These different rain models can be used to simulate various types of rain degradation. It should be noted that all these models are heuristic, as they are only rough approximations of the real world.
	
	{\flushleft\textbf{Linear Superposition Model (LSM)}.}
	The most of the existing methods adpot simple and common rain model, \emph{i.e.}, linear superposition model (LSM), to achieve the separation of undesired rain streaks and latent clear background. The rainy image $\mathbf{R}$ degraded by rain streaks is formulated as:
	\begin{equation}
		\mathbf{R}=\mathbf{B}+\mathbf{S},
		\label{Eq1}
	\end{equation}
	where $\mathbf{B}$ and $\mathbf{S}$ represent the rain-free background layer and the rain streak layer, respectively. For convenience, we use the uniform mathematical symbols in the following.
	
	{\flushleft\textbf{Screen Blend Model (SBM)}.}
	Different from the LSM, Luo \emph{et al}. \cite{luo2015removing} formulate a non-linear composite model, \emph{i.e.}, screen blend model (SBM), for modeling rainy images. It considers that the rain streak layer and background layer may affect the apperance of each other, which can be expressed as:
	\begin{equation}
		\mathbf{R}=\mathbf{B}+\mathbf{S}-\mathbf{B} \odot \mathbf{S},
	\end{equation}
	where $\odot$ denotes the element-wise multiplication operator. According to \cite{luo2015removing}, these rainy images rendered by the SBM tend to look more realistic than those by the LSM.
	
	{\flushleft\textbf{Heavy Rain Model (HRM)}.}
	In the case of heavy rain, the accumulation of distant rain streaks and water particles in the atmosphere leads to the rain veiling effect. This visual effect of rain accumulation produces a mist (or haze)-like phenomenon in the image background \cite{yang2017deep}. Based on the joint observation of the rain streak model and the atmospheric scattering model \cite{mccartney1976optics}, the formulation of this heavy rain model (HRM) is defined as:
	\begin{equation}
		\mathbf{R}=\left(\mathbf{B}+\sum_{i=1}^n \mathbf{S}_i\right) \odot \mathbf{T}+(1-\mathbf{T})\odot \mathbf{A},
	\end{equation}
	where $\mathbf{T}$ and $\mathbf{A}$ refers to the transmisson map and the global atmospheric light, respectively. Each $\mathbf{S}_i$ is termed as the rain streak layer with the same direction. Here, $i$ is the number of the rain streak layers, and its maximum value is $n$.
	
	{\flushleft\textbf{Depth-guided Rain Model (DRM)}.}
	Hu \emph{et al}. \cite{hu2019depth} formulate a rain imaging model based on the visual effects of rain in relation to scene depth information. The visibility of distant objects varies with depth from the camera, and fog visibly obscures those distant objects more than rain streaks do. The mathematical model of the depth-guided rain model (DRM) is introducted as:
	\begin{equation}
		\mathbf{R}=\left(1-\mathbf{S}-\mathbf{F}\right)\odot\mathbf{B}+\mathbf{S}+\mathbf{F} \odot \mathbf{A},
	\end{equation}
	where $\mathbf{F} \in[0,1]$ represents the fog layer. $\mathbf{F}$ is further written as $\mathbf{F}=1-e^{-\beta d(x)}$, where $\beta$ determines the thickness of fog, and $d(x)$ denotes the scene depth.
	
	{\flushleft\textbf{Transmission Medium Model (TMM)}.}
	Based on the observation that rain streaks and vapors are entangled with each other, Wang \emph{et al}. \cite{wang2020rethinking} remodel rain imaging by formulating both rain streaks and vapors as transmission medium, which is formulated as:
	\begin{equation}
		\mathbf{R}=\left(\mathbf{T}_{\mathbf{s}}+\mathbf{T}_{\mathbf{v}}\right) \odot \mathbf{B}+\left[1-\left(\mathbf{T}_{\mathbf{s}}+\mathbf{T}_{\mathbf{v}}\right)\right] \odot \mathbf{A},
	\end{equation}
	where $\mathbf{T}_\mathbf{s}$ and $\mathbf{T}_\mathbf{v}$ represent the transmission map of rain streaks and vapors, respectively.
	
	{\flushleft\textbf{Raindrop Mask Model (RMM)}.}
	The clear background image may be obscured or blurred by persistent raindrops that cling to camera lenses or window glasses as they fall and flow \cite{you2015adherent}. Qian \emph{et al}. \cite{qian2018attentive} formulate a raindrop degraded image as the result of combining the raindrop effect and the background image:
	\begin{equation}
		\mathbf{R}=(1-\mathbf{M}) \odot \mathbf{B}+\mathbf{D},
	\end{equation}
	where $\mathbf{D}$ denotes the obstruction or blurry effect brought by the raindrops. $\mathbf{M}$ is the binary mask. If $\mathbf{M}(x)=1$, the pixel $x$ in the mask belongs to the raindrop region, otherwise it is a part of the background image.
	
	{\flushleft\textbf{Mixture of Rain Model (MRM)}.}
	During outdoor image capturing, rain streaks and raindrops may co-occur. The lighting conditions that have a significant impact on the transparency of the raindrops during image capture may alter as a result of rain streaks. As a result, removing rain streaks and raindrops cannot simply be construed as a rain streak removal and a raindrop removal combined. The mixture of rain model (MRM) \cite{quan2021removing} can be modeled as:
	\begin{equation}
		\mathbf{R}=(\mathbf{1}-\mathbf{M}) \odot(\mathbf{B}+\mathbf{S})+\alpha\mathbf{D},
	\end{equation}
	where $\alpha$ indicates the atmospheric lighting coefficient.
	
	\section{Technical Development Review}
	\label{sec:review}
	In this section, we first provide a latest milestone of single image deraining in Table \ref{table1}. Then, we analyze the technical characteristics of the deep learning-based methods from five main perspectives of the learning strategies, network structures, loss functions, experimental datasets, and evaluation metrics. In what follows, we select representative methods based on the following principles: publishing in well-known journals (\emph{e.g.}, TPAMI, IJCV) and conferences (\emph{e.g.}, CVPR, ICCV), providing the official codes to ensure credibility, preferring works with more citations and Github stars.
	
	\begin{table*}[!t]
		\centering
		\caption{Summarization of representative single image deraining methods. ``SL'', ``SSL'' and ``UL'' stand for supervised/semi-supervised/unsupervised learning-based methods. ``CE'', ``Adv'', ``KL'', ``TV'', ``Contra'' and ``QS'' denote cross entropy loss, adversarial loss, Kullback Leibler loss, total variation loss, contrastive loss and quasi-sparsity loss, respectively. ``self-selected'' denotes images selected by the author rather than common datasets. ``P'', ``R'', ``F'' and ``U'' represent parameter, run-time, FLOPs and user study, respectively.  ``-'' means this item is not available or not indicated in their paper.}
		\resizebox{1.0\textwidth}{!}{
			\begin{tabular}{l||lllllllll}
				\hline
				& \textbf{Method}     & \textbf{Category} & \textbf{Learning} & \textbf{Network Structure} & \textbf{Loss Function} & \textbf{Experimental Dataset}                 & \textbf{Evaluation Metric}           & \textbf{Platform} & \textbf{Venue} \\ \hline
				\multirow{3}{*}{\textbf{\rotatebox{90}{}}}
				& Image Decomposition \cite{kang2011automatic} & Prior             &  -                 &  -                          &  -                      & self-selected                                             & VIF \#R                              & MATLAB            & TIP                  \\
				& DSC \cite{luo2015removing}                & Prior             &  -                 &  -                          &  -                      & self-selected                                             & PSNR SSIM                            & MATLAB            & ICCV                 \\
				& GMM \cite{li2016rain}                & Prior             &  -                 &  -                          &  -                      & Rain12                                        & SSIM                                 & MATLAB            & CVPR                 \\ \hline
				\multirow{4}{*}{\textbf{\rotatebox{90}{2017}}}
				& JBO \cite{zhu2017joint}               & Prior             &  -                 &  -                          &  -                      & self-selected                                        & PSNR SSIM                                 & MATLAB            & ICCV                 \\
				& JCAS \cite{gu2017joint}               & Prior             &  -                 &  -                          &  -                      & Rain12                                        & SSIM                                 & MATLAB            & ICCV                 \\
				& DerainNet \cite{fu2017clearing}          & CNN               & SL                & three layers               & $L_2$                     & Rain12                                        & SSIM BIQE \#R                        & MATLAB            & TIP                  \\
				& DDN \cite{fu2017removing}                & CNN               & SL                & residual network           & $L_2$                     & Rain14000                                     & SSIM \#R                             & Caffe             & CVPR                 \\
				& JORDER \cite{yang2017deep}             & CNN               & SL                & multi-task network         & $L_2$+CE                  & Rain100L/H Rain12                             & PSNR SSIM \#R                        & Caffe             & CVPR                 \\ \hline
				\multirow{6}{*}{\textbf{\rotatebox{90}{2018}}}
				& DID-MDN \cite{zhang2018density}            & CNN               & SL                & multi-stream network       & $L_2$+CE                  & DID-Data                                      & PSNR SSIM \#R                        & PyTorch           & CVPR                 \\
				& AttentGAN \cite{qian2018attentive}          & GAN               & SL                & recurrent network          & $L_2$+Adv                 & RainDrop                                      & PSNR SSIM                            & PyTorch           & CVPR                 \\
				& DualCNN \cite{pan2018learning}            & CNN               & SL                & multi-branch network       & $L_2$                     & Rain800                                       & PSNR                                 & PyTorch           & CVPR                 \\
				& RESCAN \cite{li2018recurrent}             & CNN               & SL                & recurrent network          & $L_2$                     & Rain100H Rain800                              & PSNR SSIM                            & PyTorch           & ECCV                 \\
				& NLEDN \cite{li2018non}              & CNN               & SL                & non-local network          & $L_1$                     & Rain100L/H DNN-Data DID-Data                  & PSNR SSIM                            & PyTorch           & ACM MM               \\
				& ResGuideNet \cite{fan2018residual}        & CNN               & SL                & residual network           & $L_2$+SSIM                & Rain100/H Rain12                              & PSNR SSIM \#P \#R                    & TensorFlow        & ACM MM               \\ \hline
				\multirow{18}{*}{\textbf{\rotatebox{90}{2019}}}
				& GCANet \cite{chen2019gated}             & CNN               & SL                & gated network              & $L_2$                     & DID-Data                                      & PSNR SSIM                            & PyTorch           & WACV                 \\
				& PReNet \cite{ren2019progressive}             & CNN               & SL                & recursive network          & SSIM                   & Rain100/H Rain12                              & PSNR SSIM \#R                        & PyTorch           & CVPR                 \\
				& DAF-Net \cite{hu2019depth}            & CNN               & SL                & residual network           & $L_2$                     & RainCityscapes Rain100H                       & PSNR SSIM                            & Caffe             & CVPR                 \\
				& HeavyRainRemoval \cite{li2019heavy}   & GAN               & SL                & multi-stage network        & $L_1$+$L_2$+Adv              & NYU-Rain Outdoor-Rain                         & PSNR SSIM                            & PyTorch           & CVPR                 \\
				& SPANet \cite{wang2019spatial}             & CNN               & SL                & multi-stage network        & $L_1$+$L_2$+Adv              & SPA-Data DID-Data                             & PSNR SSIM                            & PyTorch           & CVPR                 \\
				& UMRL \cite{yasarla2019uncertainty}               & CNN               & SL                & multi-scale network        & $L_1$                     & DID-Data Rain800                              & PSNR SSIM                            & PyTorch           & CVPR                 \\
				& SIRR \cite{wei2019semi}               & CNN               & SSL               & two subnetworks            & $L_2$+KL+TV           & DDN-SIRR                                      & PSNR SSIM                            & TensorFlow        & CVPR                 \\
				& RaindropAttention \cite{quan2019deep}  & CNN               & SL                & U-Net like network         & $L_1$                     & RainDrop                                      & PSNR SSIM                            & TensorFlow        & ICCV                 \\
				& ERL-Net \cite{wang2019erl}            & CNN               & SL                & U-Net like network         & $L_1$                     & DDN-Data DID-Data                             & PSNR SSIM                            & TensorFlow        & ICCV                 \\
				& ID-CGAN \cite{zhang2019image}            & GAN               & SL                & two subnetworks            & $L_2$+Adv                 & Rain800                                       & PSNR SSIM UQI VIF \#R                & PyTorch           & TCSVT                \\
				& DTDN \cite{wang2019dtdn}               & GAN               & SL                & multi-task network         & $L_2$+Adv                 & DID-Data                                      & PSNR SSIM UQI                        & Keras             & ACM MM               \\
				& GraNet \cite{yu2019gradual}             & CNN               & SL                & multi-stage network        & $L_1$                     & Rain100L/H Rain12                             & PSNR SSIM                            & PyTorch           & ACM MM               \\
				& ReHEN \cite{yang2019single}              & CNN               & SL                & multi-stage network        & $L_2$+SSIM                & Rain100L/H Rain800 DNN-Data DID-Data          & PSNR SSIM                            & TensorFlow        & ACM MM               \\
				& LPNet \cite{fu2019lightweight}              & CNN               & SL                & pyramid network            & $L_1$+SSIM                & Rain100L/H Rain12                             & PSNR SSIM \#P \#R                    & TensorFlow        & TNNLS                \\
				& JORDER-E \cite{yang2019joint}           & CNN               & SL                & multi-task network         & $L_1$+$L_2$                  & Rain100L/H Rain800 Rain12                     & PSNR SSIM \#R                        & Caffe             & TPAMI                \\
				& RR-GAN \cite{zhu2019singe}             & GAN               & UL                & multi-scale network        & $L_2$+Adv                 & Rain800                                       & PSNR SSIM                            & PyTorch           & AAAI                 \\
				& UD-GAN \cite{jin2019unsupervised}             & GAN               & UL                & multi-task network         & $L_1$+Adv                 & Rain100L/H Rain800 Rain12                     & PSNR SSIM \#U \#R                    & PyTorch           & ICIP                 \\
				& DualResidualNetwork \cite{liu2019dual} & CNN               & SL                & residual network           & $L_1$+SSIM                & DDN-Data DID-Data                             & PSNR SSIM                            & PyTorch           & CVPR                 \\ \hline
				\multirow{12}{*}{\textbf{\rotatebox{90}{2020}}}
				& MSPFN \cite{jiang2020multi}              & CNN               & SL                & multi-scale network        & $L_1$                     & Rain13K                                       & PSNR SSIM FSIM NIQE SSEQ             & TensorFlow        & CVPR                 \\
				& DRD-Net \cite{deng2020detail}            & CNN               & SL                & two subnetworks            & $L_2$                     & Rain200/H Rain800                             & PSNR SSIM \#R                        & TensorFlow        & CVPR                 \\
				& RCDNet \cite{wang2020model}             & CNN               & SL                & multi-stage network        & $L_2$                     & Rain100L/H Rain12 DDN-Data SPA-Data           & PSNR SSIM                            & PyTorch           & CVPR                 \\
				& Syn2Real \cite{yasarla2020syn2real}           & CNN               & SSL               & U-Net like network         & $L_1$+$L_2$               & DDN-SIRR                                      & PSNR SSIM NIQE BRISQUE               & PyTorch           & CVPR                 \\
				& All-in-One \cite{li2020all}         & GAN               & SL                & architectural search       & $L_2$                     & Outdoor-Rain Raindrop                         & PSNR SSIM                            &  -                 & CVPR                 \\
				& CVID \cite{du2020conditional}               & CNN               & SL                & auto-encoder network       & $L_2$+KL                  & Rain100L/H DNN-Data DID-Data                  & PSNR SSIM NIQE \#R                   & TensorFlow        & TIP                  \\
				& Physics GAM \cite{pan2020physics}        & GAN               & SL                & two subnetworks            & $L_1$+Adv                 & self-selected                                             & PSNR SSIM                            & PyTorch           & TPAMI                \\
				& FBL \cite{yang2020towards}                & CNN               & SL                & recurrent network          & $L_1$                     & Rain100/H Rain800                             & PSNR SSIM                            & PyTorch           & AAAI                 \\
				& PRRNet \cite{zhang2020beyond}             & CNN               & SL                & multi-stage network        & $L_1$+$L_2$+CE               & RainCityscapes RainKITTI2012/2015             & PSNR SSIM                            & PyTorch           & ECCV                 \\
				& SNet \cite{wang2020rethinking}               & CNN               & SL                & multi-branch network       & $L_1$+$L_2$                  & self-selected                                             & PSNR SSIM FID \#R                    & PyTorch           & ECCV                 \\
				& DCSFN \cite{wang2020dcsfn}              & CNN               & SL                & multi-scale network        & SSIM                   & Rain100L Rain100H DID-Data                    & PSNR SSIM NIQE                       & PyTorch           & ACM MM               \\
				& JDNet \cite{wang2020joint}              & CNN               & SL                & non-local network          & SSIM                   & Rain100L Rain100H Rain12                      & PSNR SSIM                            & PyTorch           & ACM MM               \\ \hline
				\multirow{17}{*}{\textbf{\rotatebox{90}{2021}}}
				& DualGCN \cite{fu2021rain}            & CNN               & SL                & graph network              & $L_2$                     & Rain200L/H Rain12 DID-Data DDN-Data SPA-Data  & PSNR SSIM                            & TensorFlow        & AAAI                 \\
				& DerainCycleGAN \cite{wei2021deraincyclegan}     & GAN               & UL                & CycleGAN                   & $L_1$+$L_2$+Adv              & Rain100L Rain12 Rain800 SPA-Data              & PSNR SSIM                            & PyTorch           & TIP                  \\
				& IPT \cite{chen2021pre}                & Transformer       & SL                & multi-task network         & $L_1$+Contra              & Rain100L                                      & PSNR SSIM                            & PyTorch           & CVPR                 \\
				& MPRNet \cite{zamir2021multi}             & CNN               & SL                & multi-stage network        & $L_1$                     & Rain13K                                       & PSNR SSIM \#P \#R                    & PyTorch           & CVPR                 \\
				& CCN \cite{quan2021removing}                & CNN               & SL                & architectural search       & $L_1$+SSIM                & RainDS Rain200L/H RainDrop                    & PSNR SSIM                            & PyTorch           & CVPR                 \\
				& MOSS \cite{huang2021memory}               & CNN               & SSL               & U-Net like network         & $L_1$+TV                  & DDN-SIRR                                      & PSNR SSIM \#R                        & PyTorch           & CVPR                 \\
				& RLNet \cite{chen2021robust}              & CNN               & SL                & feedback network           & $L_1$+$L_2$+SSIM             & Rain100L Rain200H Rain800 DID-Data            & PSNR SSIM NIQE \#F \#R               & PyTorch           & CVPR                 \\
				& VRGNet \cite{wang2021rain}             & GAN               & SL                & four subnetworks           & Adv+KL                 & Rain100L/H DDN-Data                           & PSNR SSIM \#U                        & PyTorch           & CVPR                 \\
				& JRGR \cite{ye2021closing}               & GAN               & SL                & CycleGAN                   & $L_1$+$L_2$+Adv              & RainCityscape                                 & PSNR SSIM                            & PyTorch           & CVPR                 \\
				& RIC-Net \cite{ni2021controlling}            & GAN               & SL                & U-Net like network         & $L_1$+$L_2$+Adv              & RainLevel5 Rain800 Rain200H DID-Data SPA-Data & PSNR SSIM                            & PyTorch           & CVPR                 \\
				& QSMD \cite{wang2021multi}               & CNN               & SL                & multi-scale network        & $L_1$+$L_2$+QS               & self-selected                                             & PSNR SSIM \#R                        & PyTorch           & CVPR                 \\
				& SPDNet \cite{yi2021structure}             & CNN               & SL                & recursive network          & $L_2$                     & Rain200L/H Rain800 DID-Data SPA-Data          & PSNR SSIM \#P \#R                    & PyTorch           & ICCV                 \\
				& UDRDR \cite{liu2021unpaired}              & CNN               & SSL               & three subnetworks          & $L_2$+TV                  & Rain100L/H DID-Data RainDirection             & PSNR SSIM NIQE \#P \#R               & PyTorch           & ICCV                 \\
				& SPAIR \cite{purohit2021spatially}              & CNN               & SL                & U-Net like network         & $L_1$                     & Rain13K                                       & PSNR SSIM                            & PyTorch           & ICCV                 \\
				& UDGNet \cite{yu2021unsupervised}             & CNN               & UL                & multi-task network         & Adv+TV                 & RainCityscapes DDN-Data                       & PSNR SSIM NIQE \#U \#R               & PyTorch           & ACM MM               \\
				& EfficientDeRain \cite{guo2021efficientderain}    & CNN               & SL                & U-Net like network         & $L_1$+SSIM                & Rain100H DDN-Data SPA-Data RainDrop           & PSNR SSIM \#R                        & PyTorch           & AAAI                 \\
				& SWAED \cite{huang2021selective}              & CNN               & SL                & U-Net like network         & $L_1$                     & Rain100L/H Rain12 DDN-Data DID-Data           & PSNR SSIM \#R \#F                    & PyTorch           & IJCV                 \\ \hline
				\multirow{10}{*}{\textbf{\rotatebox{90}{2022}}}
				& DCD-GAN \cite{chen2022unpaired}            & GAN               & UL                & CycleGAN                   & $L_1$+$L_2$+Adv              & Rain800 DID-Data DDN-Data SPA-Data            & PSNR SSIM NIQE BRISQUE               & PyTorch           & CVPR                 \\
				& NLCL \cite{ye2022unsupervised}               & GAN               & UL                & two subnetworks            & $L_2$+Contra              & RainCityscapes SPA-Data                       & PSNR SSIM NIQE \#P \#R               & PyTorch           & CVPR                 \\
				& Restormer \cite{zamir2022restormer}          & Transformer       & SL                & U-Net like network         & $L_1$                     & Rain13K                                       & PSNR SSIM                            & PyTorch           & CVPR                 \\
				& Uformer \cite{wang2022uformer}            & Transformer       & SL                & U-Net like network         & Charbonnier            & SPA-Data                                      & PSNR SSIM                            & PyTorch           & CVPR                 \\
				& DGUNet \cite{mou2022deep}             & CNN               & SL                & multi-stage network        & $L_2$                     & Rain13K                                       & PSNR SSIM                            & PyTorch           & CVPR                 \\
				& MAXIM  \cite{tu2022maxim}             & Transformer       & SL                & multi-stage network        & $L_1$+Charbonnier         & Rain13K                                       & PSNR SSIM                            & PyTorch           & CVPR                 \\
				& AirNet \cite{li2022all}             & CNN               & SL                & two subnetworks            & $L_1$+Contra              & Rain100L                                      & PSNR SSIM                            & PyTorch           & CVPR                 \\
				& TransWeather \cite{valanarasu2022transweather}       & Transformer       & SL                & recurrent network          & $L_1$+$L_2$                  & Outdoor-Rain Raindrop                         & PSNR SSIM                            & PyTorch           & CVPR                 \\
				& SEIDNet \cite{lingenerative}       & CNN       & SL                & multi-branch network          & $L_2$+SSIM                  & Rain13K DDN-Data SPA-Data                         & PSNR SSIM                            & PyTorch           & NeurIPS                 \\
				& ELF \cite{jiang2022magic}                & Transformer       & SL                & two subnetworks            & Charbonnier            & Rain13K                                       & PSNR SSIM FSIM NIQE SSEQ \#P \#R \#F & PyTorch           & ACM MM               \\
				& IDT \cite{xiao2022image}                & Transformer       & SL                & U-Net like network         & SSIM                   & Rain200L/H DID-Data DDN-Data SPA-Data RainDS  & PSNR SSIM \#P \#R \#F                & PyTorch           & TPAMI                \\ \hline
				\multirow{5}{*}{\textbf{\rotatebox{90}{2023}}}
				& HCT-FFN \cite{chen2023hybrid}            & Transformer       & SL                & multi-stage network        & $L_2$+SSIM                & Rain100L/H Rain12 RainDS                      & PSNR SSIM NIQE BRISQUE               & PyTorch           & AAAI                 \\
				& HCN \cite{fu2023continual}                & CNN               & SL                & U-Net like network         & $L_1$                     & Rain200L/H Rain12 DID-Data DDN-Data SPA-Data  & PSNR SSIM                            & PyTorch           & TPAMI                \\
				& WeatherDiffusion \cite{ozdenizci2023restoring}                & Diffusion               & SL                & U-Net like network         & $L_1$                     & Outdoor-Rain Raindrop   & PSNR SSIM NIQE                           & PyTorch           & TPAMI                \\
				& DRSformer \cite{chen2023learning}               & Transformer       & SL                & U-Net like network         & $L_1$                     & Rain200L/H DID-Data DDN-Data SPA-Data         & PSNR SSIM NIQE BRISQUE               & PyTorch           & CVPR                 \\
				& SmartAssign \cite{wang2023smartassign}               & CNN       & SL                & multi-task network         & $L_2$+QS+Contra                     & DID-Data       & PSNR SSIM              & -           & CVPR                 \\ \hline
			\end{tabular}
		}	
		\label{table1}
	\end{table*}
	
	\subsection{Method Categories}
Based on the problem formulations, image deraining methods can be categorized into model-based methods and data-driven methods \cite{wang2019survey, yang2020single}.
	Early model-based approaches \cite{he2012guided,xu2012removing,xu2012improved,kim2013single,zheng2013single,ding2016single} usually formulate the image deraining as a filtering problem and restore the rain-free image by kinds of filters. Although rain streaks can be removed, some important strictures are also removed during the filtering process.
	As image deraining is ill-posed, lots of model-based approaches develops kinds of image priors based on the statistical properties of rain streaks and clear images, including image decomposition \cite{kang2011automatic}, low-rank representation \cite{chen2013generalized}, discriminative sparse coding \cite{luo2015removing}, and Gaussian mixture model \cite{li2016rain}, etc.
Although better results have been achieved, these approaches cannot handle complex and varying rainy scenes because the hand-crafted rain models used these approaches do not hold.
In addition, most algorithms \cite{gu2017joint,zhu2017joint} still need time-consuming iterative computations due to the complexity of the optimization procedure.
	
Data-driven methods are able to learn models to better describe the properties of rain streaks and clear images.
As one of the representative data-driven approach, image deraining based on deep learning approaches have achieved significant performance.
	Since the representative methods \cite{fu2017clearing,fu2017removing,yang2017deep} have shown that using deep learning help rain removal, lot of approaches based on the convolutional neural networks (CNNs) \cite{yang2020single,fu2017clearing,fu2017removing,zhang2018density,yang2019joint,jiang2020multi} have been developed to solve image deraining and achieve decent performance in kinds of image deraining benchmarks~\cite{wang2019survey,li2019single,yang2020single,li2022toward}.
	To restore realistic images, lots of approaches employ generative adversarial networks (GANs) \cite{goodfellow2020generative} to solve image deraining. With guidance of the adversarial learning strategy, the GAN-based methods~\cite{qian2018attentive, zhang2019image, li2019heavy, pan2020physics} are able to restore clear images with better perception quality.
	
	However, as the basic operation in CNNs or GANs, the convolution operation is spatial invariant and has limited receptive field, which cannot model the spatial variant property and global structures of clear images well.
	To solve this problem, Transformers \cite{vaswani2017attention} have been applied to image deraining and achieve significant advancements as they can model the non-local information for better image reconstruction.
	Motivated by the success of recent diffusion model \cite{ho2020denoising} in generating high-quality images, diffusion-based approaches \cite{ozdenizci2023restoring} achieve promising results in kinds of image restoration problems, which may facilitate  the image deraining problem.

	In the following, we will briefly review the CNN-based, GAN-based, and Transformer-based representative deraining methods in details.
	
	\subsection{Learning Strategies}
	Along this research line, we categorize existing deep image deraining methods into supervised learning, unsupervised learning and semi-supervised learning. Based on the statistic analysis in Table \ref{table1}, supervised learning (86.2\%) is the most widely used approach. There are emerging approaches starting in 2019 that make use of unsupervised learning (8.3\%) and semi-supervised learning (5.5\%). Next, we review some representative methods of each strategy.
	
	{\flushleft\textbf{Supervised Learning}.}
	The classic deep learning-based image deraining baseline DerainNet \cite{fu2017clearing} adopts a simple CNN to learn the nonlinear mapping function from paired clear and rainy images.
	Later, diverse network structures and designs have been extensively utilized to further enhance the ability of end-to-end learning, such as multi-scale fusion \cite{wang2020dcsfn, wang2020joint} and multi-stage progressive \cite{jiang2020multi, zamir2021multi}.
	To better represent the rain distribution, several studies take rain characteristics such as rain density \cite{zhang2018density}, location \cite{yang2017deep, yang2019joint}, depth \cite{hu2019depth}, sparsity \cite{wang2021multi}, non-local \cite{li2018non}, veiling effect \cite{li2019heavy, wang2020rethinking} into account.
	Recently, Transformers-based models have achieved significant performance in single image deraining due to their ability to model non-local information required for accurate image reconstruction.
	Xiao \emph{et al.} \cite{xiao2022image} designed an image deraining Transformer (IDT) with spatial-based and complementary window-based Transformer modules.
	Very recently, Chen \emph{et al.} \cite{chen2023learning} developed a Sparse Transformer (DRSformer) to achieve state-of-the-art (SOTA) performance in the category of supervised learning methods.
	Although impressive performance has been attained, these fully-supervised approaches require paired synthetic data, which poorly mimics the degradation that occurs in the real world. To this end, the semi-supervised and unsupervised learning have been proposed for image deraining \cite{chen2022unpaired,liu2022unsupervised}.
	
	{\flushleft\textbf{Unsupervised Learning}.}
	To alleviate the pressure of collecting large-scale paired training data, unsupervised learning has been proposed for image deraining.
	In fact, learning a deep deraining network from unpaired data is more challenging compared to fully-supervised models.
	Zhu \emph{et al.} \cite{zhu2019singe} first proposed an end-to-end CycleGAN-based deraining model, RainRemoval-GAN
	(RR-GAN) without using paired training data.
	Jin \emph{et al.} \cite{jin2019unsupervised} presented an unsupervised deraining generative adversarial network (UD-GAN) by introducing self-supervised constraints from the intrinsic statistics of unpaired rainy and clear images.
	Another CycleGAN-based deraining network, termed as DerainCycleGAN \cite{wei2021deraincyclegan}, was designed using an unsupervised rain attentive detector.
	By modelling of the structure discrepancy between the clear image and rain distribution, Yu \emph{et al.} \cite{yu2021unsupervised} developed an unsupervised directional gradient based optimization model (UDGNet) for real rain streaks removal.
	Recently, Chen \emph{et al.} \cite{chen2022unpaired} leveraged dual contrastive learning to encourage the model to explore the mutual features of rainy images and clean ones in the deep feature space, where the features from the unpaired clear exemplars can facilitate rain removal.
	Likewise, Ye \emph{et al.} \cite{ye2022unsupervised} incorporated non-local self-similarity into contrastive learning to further improve unsupervised image deraining performance.
	
	{\flushleft\textbf{Semi-supervised Learning}.}
	Semi-supervised learning has recently been offered as a way to combine the advantages of supervised learning and unsupervised learning.
	Wei \emph{et al.} \cite{wei2019semi} proposed the first  semi-supervised image deraining (SSIR) framework, treating rain removal as domain adaption problem.
	The work of Syn2Real \cite{yasarla2020syn2real} was a Gaussian process based semi-supervised image deraining method by jointly modeling the labeled and unlabeled latent space.
	Huang \emph{et al.} \cite{huang2021memory} presented a memory-oriented semi-supervised (MOSS) framework to adaptively record various appearances of rain degradations.
	In \cite{liu2021unpaired}, the semi-supervised learning and knowledge distillation components were combined to formulate a rain direction regularizer, which can preserve the sharpness and structures of the rain maps.
	For unsupervised/semi-supervised rain removal, continued research in this direction is highly encouraged.
	
	\subsection{Network Architectures}
	Significant progress has been made due to the development of kinds of network architectures in image deraining. Next, we review the representative network designs in this field.

	{\flushleft\textbf{CNN-based Network Design}.}
	Eigen \emph{et al.} \cite{eigen2013restoring} designed the first CNN-based method by constructing a three-layer CNN that learns the mapping between corrupted image patches and their corresponding clean counterparts.
	Inspired by the deep residual network (ResNet) \cite{he2016deep}, Fu \emph{et al.} \cite{fu2017removing} utilized the ResNet structure to integrate the high-frequency detail layer content of an image and perform regression on the negative residual rain information.
	Yang \emph{et al.} \cite{yang2017deep} proposed a multi-task deep learning architecture that initially detects rain regions, which inherently provides supplementary information for the task of rain removal.
	To better characterize rain streaks of varying scales and shapes, a multi-stream densely connected framework was designed by \cite{zhang2018density}, consisting of two main branches: rain-density classification and rain streak removal.
	Due to the overlapping nature of rain streak layers, removing rain in a single stage is challenging. Thus, Li \emph{et al.} \cite{li2018recurrent} decomposed the rain removal into multiple stages by introducing the recurrent neural network to retain useful information from previous stages and facilitate rain removal in subsequent stages.
	By formulating a rain imaging model based on the visual effects of rain in relation to scene depth information, Hu \emph{et al.} further \cite{hu2019depth} proposed the depth-guided attention network to remove rain streaks and fog.
	Instead of deeper and complex networks, Ren \emph{et al.} \cite{ren2019progressive} discovered that a simple combination of ResNet with multi-stage recursion yields favorable deraining performance.
	Considering the role of spatial attention is to selectively focus on specific regions or locations in an image, Wang \emph{et al.} \cite{wang2019spatial} presented a spatial attention network designed to learn discriminative rain information in a local-to-global manner.
	Yasarla \emph{et al.} \cite{yasarla2019uncertainty} incorporated uncertainty-guided learning into the network to learn the residual maps and corresponding confidence maps at various scales. These maps are subsequently propagated back to guide the network in subsequent layers.
	By incorporating the Gaussian-Laplacian image pyramid decomposition technique, Fu \emph{et al.} \cite{fu2019lightweight} proposed a lightweight pyramid network to simplify the learning problem.
	Inspired by the success of multi-scale strategies in vision tasks, Jiang \emph{et al.} \cite{jiang2020multi} proposed a multi-scale progressive fusion network that better characterizes rain streaks across multiple scales through the use of image pyramid representation.
	Wang \emph{et al.} \cite{wang2020model} proposed a rain convolutional dictionary network that leverages the inherent dictionary learning mechanism to accurately encode rain shapes.
	Deng \emph{et al.} \cite{deng2020detail} constructed a two-branch parallel network consisting of a detail repair sub-network and a rain residual sub-network based on squeeze-and-excitation mechanism.
	To better compute channel-wise correlation and spatial-wise coherence, Fu \emph{et al.} \cite{fu2021rain,fu2021successive} introduced a dual graph convolutional architecture to help image restoration and facilitate rain removal.
	Under the iterative guidance of the residue channel prior, Yi \emph{et al.} \cite{yi2021structure} proposed a structure preserving deraining architecture to reconstruct high-quality rain-free images.
	Drawing on the concept of closed-loop control, Li \emph{et al.} \cite{chen2021robust} designed a robust representation learning network structure by incorporating feedback mechanism into the CNN.
	Recently, Fu \emph{et al.} \cite{fu2023continual} developed a patch-wise hypergraph convolutional network architecture to help the model to explore non-local content of the images.
	
	{\flushleft\textbf{GAN-based Network Design}.}
	Driven by GAN in the image generation task \cite{goodfellow2020generative}, Qian \emph{et al.} \cite{qian2018attentive} incorporated a GAN-based architecture, where the generative network employs an attentive-recurrent network to generate an attention map. This attention map, along with the input image, is then utilized in a contextual autoencoder to generate a rain-free result.
	Afterwards, Zhang \emph{et al.} \cite{zhang2019image} proposed a conditional GAN-based architecture with a densely-connected generator and a multi-scale discriminator.
	To achieve heavy rain image restoration, the method of Li \emph{et al.} \cite{li2019heavy} consists of a two-stage network architecture: an initial physics-based sub-network followed by a depth-guided GAN refinement sub-network.
	Based on the consistency between the estimated results of the physical model and the observed image, Pan \emph{et al.} \cite{pan2020physics} proposed a GAN-based network constrained by a physics model to remove rain.
	Motivated by the image disentanglement strategy \cite{chen2022unpaired2}, Ye \emph{et al.} \cite{ye2021closing} presented a CycleGAN-based joint rain generation and removal framework by performing the translations on the simpler rain space.
	Ni \emph{et al.} \cite{ni2021controlling} put forward a rain intensity controlling GAN, which comprises three sub-networks: a main controlling network, a high-frequency rain-streak elimination network, and a background extraction network, which enables seamless control over rain intensities by leveraging interpolation techniques within the deep feature space.
	With the popularity of generative models \cite{shaham2019singan}, Wang \emph{et al.} \cite{wang2021rain} introduced a variational rain generation network to implicitly infer the underlying statistical distribution of rain.
	
	{\flushleft\textbf{Transformer-based Network Design}.}
	Inspired by the success of Transformer-based networks in the NLP and high-level vision field \cite{han2022survey}, image processing Transformer (IPT) \cite{chen2021pre} was first designed with multi-heads and multi-tails shared transformer body to solve a host of low-level vision tasks, including image deraining.
	In order to consider the high compatibility of Transformers and CNNs in the image deraining task, Jiang \emph{et al.} \cite{jiang2022magic} integrated the benefits of self-attention and CNNs into an association learning-based network for efficient rain removal.
	Similarly, Chen \emph{et al.} \cite{chen2023hybrid} formulated a hybrid stage-by-stage progressive network structure that combines the features by CNN and Transformer to help image restoration.
	UNet \cite{ronneberger2015u}, as a classic network architecture design, excels in image restoration tasks due to its ability to capture both high-level and low-level features through the use of skip connections.
	Wang \emph{et al.} \cite{wang2022uformer} built a U-shaped hierarchical network architecture with locally-enhanced window Transformer blocks.
	To reduce computational complexity, Zamir \emph{et al.} \cite{zamir2022restormer} proposed an efficient Transformer consisting of a multi-Dconv head transposed attention and a gated-Dconv feed-forward network.
	To compensate for the shortcomings of position encoding, Xiao \emph{et al.} \cite{xiao2022image} developed a relative position enhanced multi-head self-attention to improve the representation
	ability.
	Instead of vanilla self-attention in Transformers \cite{vaswani2017attention}, a top-\emph{k} sparse attention \cite{chen2023learning} was introduced into the model to maintain the most relevant features for boosting deraining performance.

	\subsection{Loss Functions}
	It is vital to choose suitable loss functions to constrain the model for better removing rain. The choice of loss functions is highly diverse among the existing approaches. We detail representative loss functions as follows.

	{\flushleft\textbf{Content-based Loss Function}.}
	The commonly used loss functions are based on the image content, which is usually defined as: $\mathcal{L}_1$ (Mean Absolute Error, MAE) and $\mathcal{L}_2$ (Mean Squared Error, MSE) losses can be written as:
	\begin{equation}
		\mathcal{L}_{content}=\frac{1}{H W C} \sum_{i=1}^H \sum_{j=1}^W \sum_{c=1}^C\rho(\widehat{\mathbf{B}}-\mathbf{B}),
	\end{equation}
	where $\widehat{\mathbf{B}}$ and $\mathbf{B}$ denote the derained image and ground-truth; $\rho(\cdot)$ denotes the robust function.
	$H$, $W$ and $C$ represent the height, width, and number of channels of the image, respectively. The commonly used ones include:
	\begin{align}
		\begin{split}
			&\rho(\widehat{\mathbf{B}}-\mathbf{B})=\|\widehat{\mathbf{B}}-\mathbf{B}\|^p,\\
			&\rho(\widehat{\mathbf{B}}-\mathbf{B})=\sqrt{\left\|\widehat{\mathbf{B}}-\mathbf{B}\right\|^2+\epsilon^2}, \\
		\end{split}
	\end{align}
	where $p$ and $\epsilon$ denote the positive scalars.

	{\flushleft\textbf{Structure-based Loss Function}.}
	The structural similarity index measure (SSIM) is used to measure the quality of restored images. It is defined as:
	\begin{equation}
		\mathcal{L}_{\mathrm{SSIM}}=1-\frac{2 \mu_B \mu_{\widehat{B}}+C_{1}}{\mu_{B}^{2}+\mu_{\widehat{B}}^{2}+C_{1}} \cdot \frac{2 \sigma_{B\widehat{B}}+C_{2}}{\sigma_{B}^{2}+\sigma_{\widehat{B}}^{2}+C_{2}},
	\end{equation}
	where $\mu_{B}$ and $\mu_{\widehat{B}}$ are the average of $\mathbf{B}$ and $\widehat{\mathbf{B}}$ over pixels, $\sigma_{B}$ and $\sigma_{\widehat{B}}$ are the variances of $\mathbf{B}$ and $\widehat{\mathbf{B}}$, $\sigma_{B\widehat{B}}$ is the covariance between $\sigma_{B}$ and $\sigma_{\widehat{B}}$. $C_{1}$ and $C_{2}$ are two fixed constants.

	{\flushleft\textbf{Adversarial Loss Function}.}
	The adversarial loss is derived from the minimax game between the generator $G$ and discriminator $D$, which usually is formulated as:
	\begin{equation}
		\min _G \max _{D} \mathbbm{E}_{R \sim S}[\log (1-D(G(\mathbf{R}))]+\mathbbm{E}_{B \sim T}[\log (D(\mathbf{B}))],
	\end{equation}
	where the rainy images $\mathbf{R}$ are sampled from the source manifold $S$ while the clear images $\mathbf{B}$ are sampled from the target manifold $T$. Here, $G$ minimizes the loss to produce clear images that resemble real samples. In contrast, $D$ maximizes the loss to discriminate between generated clear images and real clear images. To optimize the generator, we can define it as:
	\begin{equation}
		\mathcal{L}_{\mathrm{G}}=\log (1-D(G(\mathbf{R}))).
	\end{equation}
	
	Meanwhile, the loss function for optimizing the discriminator is minimized by:
	\begin{equation}
		\mathcal{L}_{\mathrm{D}}=-\log (1-D(G(\mathbf{R})))-\log (D(\mathbf{B})).
	\end{equation}
	
	{\flushleft\textbf{Cross Entropy Loss Function}.}
	The cross entropy loss for rain density classification \cite{yang2017deep, zhang2018density, zhang2020beyond} is designed to take advantage of the rain density information to guide the deraining process:
	\begin{equation}
		\mathcal{L}_{\mathrm{CE}}=-\sum_{i=1}^n p\left(x_i\right) \log \left(q\left(x_i\right)\right),
	\end{equation}
	where $x_i$ denotes the data sample, $p(x_i)$ and $q(x_i)$ are two probability distributions on random variables.
	
	{\flushleft\textbf{Kullback-Leibler Loss Function}.}
	The Kullback-Leibler (KL) divergence is used to measure the similarity of different distributions. recent studies \cite{du2020conditional, wang2021rain} employ this metric to measure the distribution similarity between the ground-truth and derained image:
	\begin{equation}
		\mathcal{L}_{\mathrm{KL}}=\sum_{i=1}^n p(x_i) \log \frac{p(x_i)}{q(x_i)}.
	\end{equation}
	
	{\flushleft\textbf{Total Variation Loss Function}.}
	The original total variation is mainly regarded as an effective image prior to for image restoration~\cite{chan2005recent}.
	Given its decent performance in image restoration, several approaches\cite{wei2019semi,huang2021memory,liu2021unpaired,yu2021unsupervised} use it to constrain the network training for better performance. This constraint is defined as:
	\begin{equation}
		\mathcal{L}_{\mathrm{TV}}=\left\|\nabla_h \widehat{\mathbf{B}}\right\|_1+\left\|\nabla_v \widehat{\mathbf{B}}\right\|_1,
	\end{equation}
	where $\nabla_h$ and $\nabla_v$ represent the horizontal and vertical gradient operators, respectively.
	
	{\flushleft\textbf{Contrastive Loss Function}.}
	Contrastive loss function is mainly used to improve the perceptual quality of derained images by positive and negative examples\cite{chen2022unpaired, ye2022unsupervised}. This constraint is usually defined as:
	\begin{equation}
		\mathcal{L}_{\mathrm{Cont}}=-\log \frac{\exp \left(v \cdot v^{+} / \tau\right)}{\sum_{i=0}^n \exp \left(v \cdot v_i^{-} / \tau\right)},
	\end{equation}
	where $v$ denotes a query representation, $v^{+}$ and $v_i^{-}$ are corresponding positive and negative samples. $n$ is the number of negatives, $\tau$ denotes a temperature parameter, and $\cdot$ denotes the inner product operation.
	
	{\flushleft\textbf{Quasi Sparsity Loss Function}.}
	The quasi sparsity loss function is mainly used to intrinsic sparsity prior of clear images~\cite{wang2021multi}:
	\begin{equation}
		\mathcal{L}_{\mathrm{QS}}=\sum_{t=1}^N \sum_{i, k}\left\|\omega_{i, k} * \widehat{\mathbf{B}}\right\|_1+\left\|\omega_{i, k} *\left[\mathbf{R}-\widehat{\mathbf{B}}\right]\right\|_1,
	\end{equation}
	where $*$ is the convolution operation. $\omega_{i, k}$ denotes the $k^{t h}$ filter centered at the $i^{t h}$ pixel.
	
	\begin{table*}[!t]
		\centering
		\caption{Summarization of public datasets for the single image deraining task. ``Syn'' and ``Real'' denote the synthetic and real-world rainy datasets. ``RS'', ``RD'', and ``RA'' represent the rain streak, raindrop and rain accumulation effect, respectively. Note that DDN-Data \cite{fu2017removing} and DID-Data \cite{zhang2018density} are also termed as Rain1400 and Rain1200 in some papers.}
		\resizebox{1.0\textwidth}{!}{
			\begin{tabular}{lllllll}
				\hline
				\textbf{Dataset} & \textbf{Train/Test} & \textbf{Syn/Real} & \textbf{Category} & \textbf{Rain Model} & \textbf{Venue} & {\textbf{Download Link}}                            \\ \hline
				Rain12 \cite{li2016rain}          & 0/12                & Syn               & RS                & LSM                 & CVPR 2016            & http://yu-li.github.io/paper/li\_cvpr16\_rain.zip                     \\
				Rain100L \cite{yang2017deep}        & 1,800/100            & Syn               & RS                & LSM                 & CVPR 2017            & https://www.icst.pku.edu.cn/struct/Projects/joint\_rain\_removal.html \\
				Rain100H \cite{yang2017deep}        & 1,800/100            & Syn               & RS                & LSM                 & CVPR 2017            & https://www.icst.pku.edu.cn/struct/Projects/joint\_rain\_removal.html \\
				Rain200L \cite{yang2017deep}        & 1,800/200            & Syn               & RS                & LSM                 & CVPR 2017            & https://www.icst.pku.edu.cn/struct/Projects/joint\_rain\_removal.html \\
				Rain200H \cite{yang2017deep}        & 1,800/200            & Syn               & RS                & LSM                 & CVPR 2017            & https://www.icst.pku.edu.cn/struct/Projects/joint\_rain\_removal.html \\
				DDN-Data \cite{fu2017removing}        & 12,600/1,400          & Syn               & RS                & LSM                 & CVPR 2017            & https://xueyangfu.github.io/projects/cvpr2017.html                    \\
				DID-Data \cite{zhang2018density}        & 12,000/1,200          & Syn               & RS                & LSM                 & CVPR 2018            & https://github.com/hezhangsprinter/DID-MDN                            \\
				RainDrop \cite{qian2018attentive}        & 861/249             & Syn               & RD                & RMM                 & CVPR 2018            & https://github.com/rui1996/DeRaindrop                                 \\
				Rain800 \cite{li2019single}         & 700/100             & Syn               & RS                & LSM                 & TCSVT 2019           & https://github.com/hezhangsprinter/ID-CGAN                            \\
				SPA-Data \cite{wang2019spatial}        & 638,492/1,000         & Real              & RS                & LSM                 & CVPR 2019            & https://github.com/stevewongv/SPANet                                  \\
				MPID \cite{li2019single}            & 3,961/419            & Syn+Real          & RS+RD+RA          & LSM+HRM             & CVPR 2019            & https://github.com/panda-lab/Single-Image-Deraining                   \\
				RainCityscapes \cite{hu2019depth}  & 9,432/1,188           & Syn               & RS+RA             & DRM                 & CVPR 2019            & https://github.com/xw-hu/DAF-Net                                      \\
				Outdoor-Rain \cite{li2019heavy}    & 9,000/1,500           & Syn               & RS+RA             & HRM                 & CVPR 2019            & https://github.com/liruoteng/HeavyRainRemoval                         \\
				Rain13K \cite{jiang2020multi}   & 13,712/4,298           & Syn               & RS                & LSM                 & CVPR 2020            & https://github.com/kuijiang94/MSPFN                     \\	
				RainKITTI2012 \cite{zhang2020beyond}   & 4,062/4,085           & Syn               & RS                & LSM                 & ECCV 2020            & https://github.com/HDCVLab/Stereo-Image-Deraining                     \\
				RainKITTI2015 \cite{zhang2020beyond}   & 4,200/4,189           & Syn               & RS                & LSM                 & ECCV 2020            & https://github.com/HDCVLab/Stereo-Image-Deraining                     \\
				RainDS \cite{quan2021removing}          & 3,450/900            & Syn+Real          & RS+RD             & LSM+RMM+MRM         & CVPR 2021            & https://github.com/Songforrr/RainDS\_CCN                              \\
				RainDirection \cite{liu2021unpaired}   & 2,920/430            & Syn               & RS                & LSM                 & ICCV 2021            & https://github.com/Yueziyu/RainDirection-and-Real3000-Dataset         \\
				GT-RAIN \cite{ba2022not}         & 28,217/2,100          & Real              & RS                & LSM                 & ECCV 2022            & https://github.com/UCLA-VMG/GT-RAIN                                   \\ \hline
			\end{tabular}
		}	
		\label{table2}
	\end{table*}
	
	\subsection{Deraining Datasets}
	To better evaluate the image deraining methods, lots of image deraining datasets have been proposed.
	Table \ref{table2} presents an overview of the existing datasets for single image deraining, including synthetic and real-world datasets.
	
	{\flushleft\textbf{Rain12}} \cite{li2016rain} is only a test dataset that contains 12 synthesized images with a single sort of rain streak.
	
	{\flushleft\textbf{Rain100L} and \textbf{Rain100H}}  \cite{yang2017deep} contain 1,800 synthetic image pairs for training and 100 ones for test, where the clear images are selected from the BSD dataset \cite{arbelaez2010contour}. These rain streaks are created either by adding simulated line streaks or by using photorealistic rendering techniques \cite{garg2006photorealistic}.
	
	{\flushleft\textbf{Rain200L} and \textbf{Rain200H}}
	\cite{yang2017deep} are created based on the original \textbf{Rain100L} and \textbf{Rain100H} datasets by filtering out duplicate background images. Among them, there are 1,800 synthetic training pairs and 200 test images.
	
	{\flushleft\textbf{DDN-Data}} \cite{fu2017removing}, also known as \textbf{Rain1400}, contains 12,600 image pairs for training and 1,400 ones for test, where the clear images are selected from the BSD dataset \cite{arbelaez2010contour}, UCID dataset \cite{schaefer2003ucid} and Google image search. Each clear image is used to generate 14 synthetic images with different rain directions and density levels.
	
	{\flushleft\textbf{DID-Data}} \cite{zhang2018density}, also known as \textbf{Rain1200}, consists of 12,000 synthetic training pairs and 1,200 test pairs with three rain density levels (\emph{i.e.}, light, medium, and heavy).
	
	{\flushleft\textbf{RainDrop}} \cite{qian2018attentive} is the first raindrop removal dataset, which conains 1,119 pairs of raindrop images with varied backgrounds using a camera with two aligned pieces of glass (one sprayed with water, and the other is left clean).
	
	{\flushleft\textbf{Rain800}} \cite{li2019single} includes 700 image pairs for training and 100 for test, where the clear images are randomly chosen from the BSD dataset \cite{arbelaez2010contour} and UCID dataset \cite{schaefer2003ucid}.
	
	{\flushleft\textbf{SPA-Data}} \cite{wang2019spatial} is the first paired real-world dataset which utilizes the human-supervised percentile video filtering to obtain the ground turths. It contains 638,492 rainy/clear image patches for training and 1,000 testing ones.
	
	{\flushleft\textbf{MPID}} \cite{li2019single} serves both machine and human vision by incorporating a considerably wider variety of rain models, including both synthetic and real-world images. There are three different forms of rain in it: rain streak, raindrops, and rain mist. The training set contains 2,400, 861, and 700 image pairs, whereas the test set has 200, 149, and 70 image pairs.
	
	{\flushleft\textbf{RainCityscapes}} \cite{hu2019depth} is made up of 262 training images and 33 test images from Cityscape's training and validation sets \cite{cordts2016cityscapes}, which are chosen as clear background images. The authors simulate rain and fog on the photographs using the camera settings and scene depth information.
	
	{\flushleft\textbf{Outdoor-Rain}} \cite{li2019heavy} contains 9,000 training samples and 1,500 test samples, where the clear backgrounds are obtained from \cite{qian2018attentive}. It also considers depth information to synthesize rain accumulation by using~\cite{godard2017unsupervised}.
	
	{\flushleft\textbf{Rain13K}} \cite{jiang2020multi} is the mixed datasets collected from multiple previous datasets, which consists of 13,712 image pairs for training and 4,298 test images. There are five test sets, i.e., Rain100H \cite{yang2017deep}, Rain100L \cite{yang2017deep}, Test100 \cite{li2019single}, Test2800 \cite{fu2017removing}, and Test1200 \cite{fu2017removing}.
	
	{\flushleft\textbf{RainKITTI2012} and \textbf{RainKITTI2015}} \cite{zhang2020beyond} are two stereo image deraining datasets synthesized based on the KITTI stereo 2012 dataset and KITTI stereo 2015 dataset \cite{geiger2013vision}. RainKITTI2012 and RainKITTI2015 contains 4,062 and 4,200 training pairs with various scenarios. There are 4,085 and 4,189 rainy images for test.
	
	{\flushleft\textbf{RainDS}} \cite{quan2021removing}
	is divided into two subsets: RainDS-Syn and RainDS-Real. Based on first-person view driving scenarios from autonomous driving datasets, RainDS-Syn is a synthetic dataset made up of 3,600 image pairs corrupted by raindrops and rain streaks. By manually mimicking rainfall with a sprinkler, RainDS-Real is a collection of 750 real-world images degraded by raindrops and rain streaks.
	
	{\flushleft\textbf{RainDirection}} \cite{liu2021unpaired} contains 2,920 high-resolution synthetic image for training and 430 ones for test, where the clear images are selected from the Flick2K and DIV2K dataset \cite{timofte2017ntire}. Each rainy image is assigned with a direction label.
	
	{\flushleft\textbf{GT-RAIN}} \cite{ba2022not} is the large-scale dataset with real paired data captured diverse rain effects. It contains 31,524 rainy and clear frame pairings, which are divided into 26,124 training frames, 3,300 validation frames, and 2,100 test frames.
	
	\subsection{Evaluation Metrics}
	There are several ways to evaluate the performance of image deraining models. One common way is to use some evaluation metrics including full and/or non-reference image quality assessment (IQA) and human-based evaluation (\emph{e.g.}, user study) to evaluate the quality of the derained images. In addition to the quality of the restoration results, the model complexity is also an important factor. Moreover, given the image deraining can be regarded as a pre-processing step, whether the derained images facilitate the following tasks is another commonly used evaluation metric. In the following, we provide details about these aforementioned evaluation metrics.

	\begin{figure*}[t]
	\centering
	\includegraphics[width=1.0\textwidth]{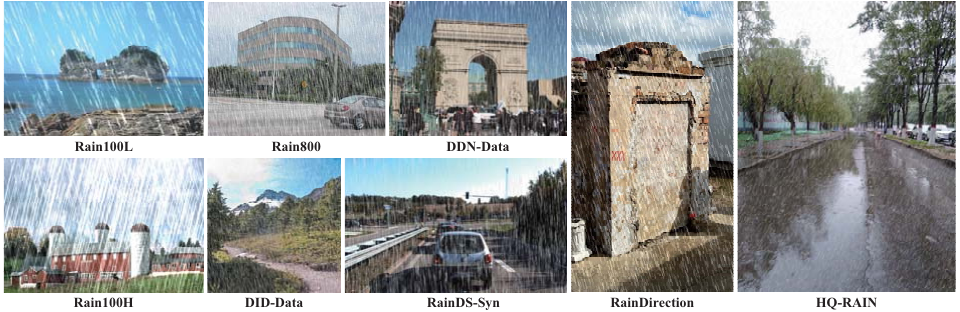}
	\caption{Example images from previous representative datasets \cite{yang2017deep, fu2017removing, zhang2018density, li2019single, quan2021removing, liu2021unpaired} and our proposed HQ-RAIN.}
	\label{fig2}
    \end{figure*}
	
	{\flushleft\textbf{Full-Reference Metrics}}.
	The commonly used full-reference IQA metrics include Peak Signal-to-Noise Ratio (PSNR) \cite{huynh2008scope}, Structure Similarity Index (SSIM) \cite{wang2004image}, Feature Similarity (FSIM) \cite{zhang2011fsim}, and Learned Perceptual Image Patch Similarity (LPIPS) \cite{zhang2018unreasonable}. The PSNR assesses the pixel-level similarity between two images, whereas the SSIM measures similarity according to structure information. The FSIM and LPIPS measure similarity at the feature level between image pairs for quality evaluation. Note that, higher PSNR, SSIM, FSIM, and lower LPIPS indicate better image visual quality.
	
	{\flushleft\textbf{Non-Reference Metrics}}.
	In terms of the rainy images without ground truth images, the non-reference IQA indicators are used for quantitatively evaluate the restoration performance, including Natural Image Quality Evaluator (NIQE) \cite{mittal2012no}, Blind/Referenceless Image Spatial Quality Evaluator (BRISQUE) \cite{mittal2012making}, and Spatial-Spectral Entropy-based Quality (SSEQ) \cite{liu2014no}. The smaller scores of NIQE, BRISQUE and SSEQ indicate clearer contents and better perceptual quality.
	
	{\flushleft\textbf{User Study}}.
	User study is a representative subjective evaluation method. Users can choose the image with the best deraining performance from a group of images. The premise is to anonymize the method and randomly sort the images in each group to ensure fairness. In general, the user study score is the mean opinion score (MOS) from a group of participants. A high MOS indicates superior perceptual quality from human perspectives.
	
	{\flushleft\textbf{Model Efficiency}}.
	Model efficiency is critical driver for the practical application of deep image deraining algorithms. The evaluation metrics for model efficiency typically include the numbers of parameters (\#Params), inference running time, and Floating Point Operations (FLOPs). It is widely known that a shorter running time and a smaller \#Params and FLOPs means better model efficiency.
	
	{\flushleft\textbf{Application-based Evaluations}}.
	One of the goals of image deraining, in addition to enhancing the visual quality, is to serve high-level vision tasks, such as object recognition \cite{liu2020deep} and segmentation \cite{wojna2019devil}. Thus, to verify the performance of different methods, the effects of image deraining on outdoor vision based applications are further investigated.
	
	{\flushleft\textbf{Robust Analyses on Adversarial Attacks}}.
	Rain highly degrades in a variety of ways, and the deraining model may suffer from a performance loss due to scene inconsistencies.  Adversarial attacks try to degrade the deraining algorithms' output by adding a small quantity of visually undetectable disturbances to the input rainy images. Recent studies \cite{yu2022towards, liu2023adversarial} have investigated various types of adversarial attacks focusing on rain degradation problems and their impact on both human and machine vision tasks.
		
	\begin{figure*}[t]
	\centering
	\includegraphics[width=1.0\textwidth]{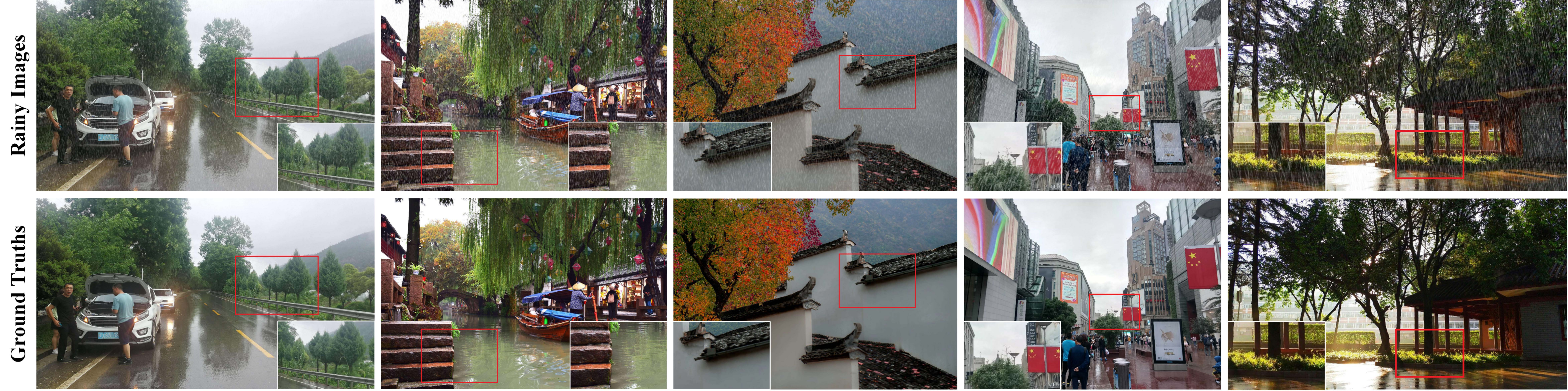}
	\caption{More example image pairs sampled from the proposed HQ-RAIN. Best viewed by zooming in the figures.}
	\label{fig3}
	\end{figure*}

	\section{HQ-RAIN: A New Benchmark}
	\label{sec:benchmark}
	Although lots of single image deraining datasets have been proposed, the quality of most synthetic rain dataset is sub-optimal in terms of realism and harmony. As a result, a large domain gap exists between synthetic and real data, thereby limiting the capacity to train better deep deraining models. In the following, we first describe the construction of the new benchmark and then compare it with other commonly used image deraining datasets.

	\subsection{Dataset Construction}
	Existing approaches usually add rain streaks into the clear images to obtain rainy images. However, this will leads to unnatural results as shown in Figure \ref{fig2}, especially in the sky region.
	Instead of simply adding rain streaks into the clear images, we develop an effective image synthesization approach to obtain more realistic datasets.
	Our method include the background collection, rain streaks synthesis, and image blending, which will be presented in the follows.
	
	{\flushleft\textbf{Background Collection}}.
	The quality of clear backgrounds (\emph{i.e.}, ground-truth images) is equally significant for constructing paired datasets, which was not taken into account in previous studies.
	In other words, the existing synthetic datasets \cite{yang2017deep, zhang2018density, li2019single, jiang2020multi, liu2021unpaired} only focus on the synthesized rain while ignoring the high-quality backgrounds that we also need.
	On the one hand, the ground-truths of these datasets have some unexpected problems about the images: low resolution, watermark, compression artifact, which may interfere with the quality of model learning and high-quality image reconstruction.
	On the other hand, they often overlook the basic fact that rainy weather mostly occurs under cloudy or low background brightness imaging conditions.
	When revisiting the rainy images in the existing datasets, we find that there are very obvious rain streaks in the clear blue sky region, which makes them look incompatible.
	%
	
	To ensure more realistic and harmonious synthetic rainy images for the next step, our selection of ground-truths is based on a strict set of collection criteria.
	Specifically, we first collect clear and rain-free scenes using a Python program based on Scrapy to download images from Google search.
	In addition, we also elaborately select several backgrounds covering abundant scenes from the DPED \cite{ignatov2017dslr} and RainDS \cite{quan2021removing} datasets.
	Low quality images that contain poor resolution, website watermark, compression artifact and blur are filtered out.
	All clear blue sky regions are filtered out as well to ensure appropriate background composition.
	Here, we tend to choose suitable rain-free backgrounds based on human visual perception of real rainy days by considering sky, illumination, and ground conditions.
	Overall, our ground-truths covers a large variety of typical daylight and night scenes from urban locations (\emph{e.g.}, streets, buildings, cityscapes) to natural scenery (\emph{e.g.}, hills, lakes, vegetations).
	Similar to \cite{ba2022not}, despite the fact that our collection of ground-truths is reliant on streamers, Google Image's fair use clause permits for its distribution to the academic community.
	
	{\flushleft\textbf{Rain Streaks Synthesis}}.
	The fidelity and diversity of rain are two key factors in the rain streaks synthesis step.
	For convenience, most synthesis methods \cite{li2019single} adopt Photoshop software \footnote[1]{The PhotoShop implementation of the rain streaks synthesis method. Please refer to \url{https://www.photoshopessentials.com/photo-effects/photoshop-weather-effects-rain/}.} to render the streaks.
	However, this manual synthesis based method is time-consuming and labor-intensive.
	Inspired by \cite{garg2007vision, wang2020rain}, we model the generation of rain streak layers as the motion blur process, which naturally takes advantage of two crucial aspects of rain streaks: repeatability and directionality. Mathematically, it can be expressed as:
	\begin{equation}
		\mathbf{S}=\mathbf{K}(l, \theta, w) * \mathbf{N}(n),
	\end{equation}
	where $\mathbf{N}$ denotes the rain mask generated by random noise $n$. Here, we use uniform random numbers and thresholds to control the level of noise. $l$ and $\theta$ are the length and angle of the motion blur kernel $\mathbf{K} \in \mathbbm{R}^{p \times p}$. We further add the rotated diagonal kernel using Gaussian blur to make the rain thickness $w$. The noise quantity $n$, rain length $l$, rain angle $\theta$, and rain thickness $w$ are obtained by sampling from $[100,300]$, $[20,40]$, $[40^{\circ},120^{\circ}]$, and $[3,7]$, respectively. $*$ represents the spatial convolution operator.
	
	Although in different synthesis methods the rain streaks are simulated and visually similar to humans, our method exhibits more flexible and higher rain diversity, which have great effect on the coverage of real-world rain.
	
	{\flushleft\textbf{Image Blending}}.
	Most existing synthetic rainy images add rain streaks linearly to the rain-free backgrounds, which can easily make the composite image look unnatural, especially in the sky area.
	Our goal is to ensure the visual realism and harmony of synthesized rainy images, thereby reducing the domain gap between the synthetic and real images, which is never explored before.
	Thus, instead of directly copy-and-pasting, we adopt image blending \cite{porter1984compositing} technique to yield a synthesized image.
	Compared to image harmonization task \cite{cong2020dovenet}, due to the similarity of rain streaks, we do not need to accurately depict objects for the blending mask \cite{zhang2020deep}.
	To this end, alpha blending is utilized to process the rain layer and background layer, where the alpha value of a pixel in a given layer indicates how much of the colors from lower levels may be visible through the color at that level.
	Formally, it can be defined as follows:
	\begin{equation}
		\mathbf{R}_{r,g,b}=\alpha \odot \mathbf{S} + (1-\alpha) \odot \mathbf{B},
	\end{equation}
	where $\alpha$ is the blending ratio. Here, we set $\alpha$ to $[0.8, 0.9]$. $\odot$ denotes the element-wise multiplication operator. Noted that $R$, $G$, and $B$ channels are processed separately.
	
	{\flushleft\textbf{Benchmark Statistics}}.
	As a result, we propose a new single image deraining benchmark with high-quality backgrounds, diverse rain streaks, and harmonious layer blending, called HQ-RAIN.
	In total, the training and testing set of the HQ-RAIN contains 4,500 and 500 synthetic images, respectively. The average resolution of all images is $1367 \times 931$.
	See Figure~\ref{fig3} for several image pairs in HQ-RAIN.
	
	\begin{figure}[!t]
		\centering
		\includegraphics[width=1.0\columnwidth]{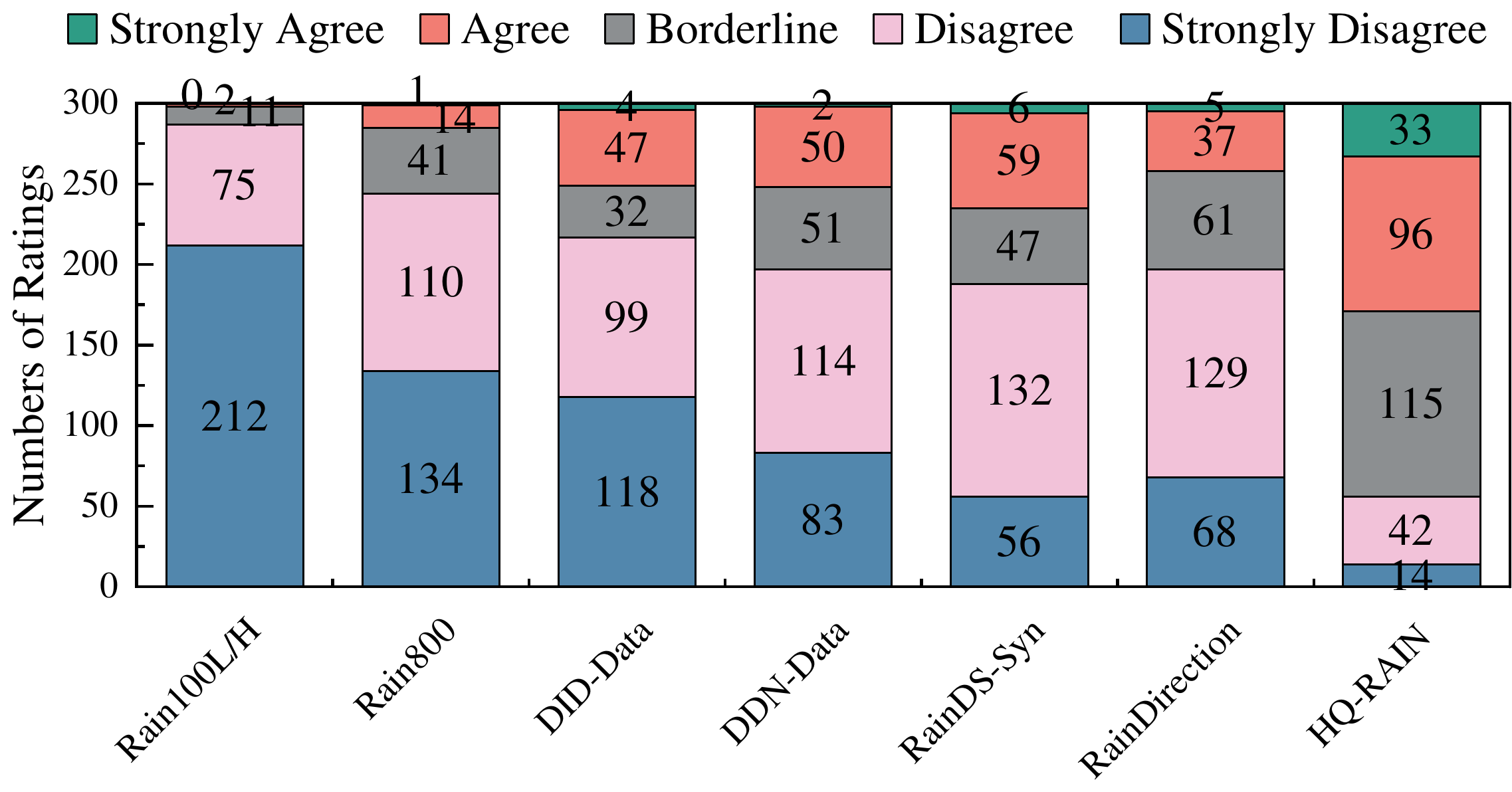}
		\caption{User study results. The ratings given by all participants on different synthesis datasets.}
		\label{fig4}
	\end{figure}
	
	We also propose another new realistic dataset named RE-RAIN, to uniformly evaluate generalization performance of deraining models.
	Although several unlabeled real datasets have been collected \cite{wei2019semi, wei2021semi}, there are some drawbacks that have the negative effects on the evaluation of generalization performance.
	For one thing, these real datasets contain some low-resolution images with watermarks, as well as unusual images from other bad weather conditions, which is beyond the scope of research on removing rain.
	For another, some rainy images have too light rain streaks, making it difficult to recognize their rain regions and thus unable to effectively evaluate deraining performance.
	To this end, we create a high-quality real benchmark RE-RAIN for evaluating real-world image deraining, containing 300 real rainy images without ground truths which are elaborately selected from the Internet and related works \cite{liu2021unpaired}.

	\subsection{Comparisons with Previous Datasets}
	{\flushleft\textbf{Subjective Assessment}}.
	We conduct an online user study to evaluate the quality (\emph{i.e.}, how realistic) of the synthesis rainy images.
	Following the \cite{wang2021rain}, we prepare for 70 rainy images, randomly chosen from 7 datasets (\emph{i.e.}, Rain100L/H \cite{yang2017deep}, Rain800 \cite{li2019single}, DID-Data \cite{zhang2018density}, DDN-Data \cite{fu2017removing}, RainDS-Syn \cite{quan2021removing}, RainDirection \cite{liu2021unpaired} and HQ-RAIN) with 10 samples from each dataset.
	We recruit 30 participants with 15 males and 15 females. For each participant, we randomly show them 70 rainy images. Then, using a 5-point Likert scale (\emph{i.e.}, strongly agree, agree, borderline, disagree, and strongly disagree), all participants are asked to judge how realistic each image looks. Finally, 300 ratings are received for each category.
	Figure \ref{fig4} shows the user study results. Our HQ-RAIN consistently outperforms other benchmarks, which also reveals that our synthetic rain is evaluated to be substantially more realistic than previous datasets.
	
	\begin{table*}[!t]
		\centering
		\caption{Dataset descriptions of mixed training track and independent training track.}
		\resizebox{1.0\textwidth}{!}{
			\begin{tabular}{c|ccccccc|ccccc}
				\hlinew{1.0pt}
				Tracks          & \multicolumn{7}{c|}{\textbf{Mixed Training Track}}                                                                & \multicolumn{5}{c}{\textbf{Independent Training Track}}                                           \\
				Source Datasets & \begin{tabular}[c]{@{}c@{}}Rain800\\ {\cite{li2019single}}\end{tabular}          & \begin{tabular}[c]{@{}c@{}}Rain100H\\ {\cite{yang2017deep}}\end{tabular}          & \begin{tabular}[c]{@{}c@{}}Rain100L\\ {\cite{yang2017deep}}\end{tabular}          & \begin{tabular}[c]{@{}c@{}}Rain14000\\ {\cite{fu2017removing}}\end{tabular}         & \begin{tabular}[c]{@{}c@{}}Rain1200\\ {\cite{fu2017removing}}\end{tabular}          & \begin{tabular}[c]{@{}c@{}}Rain12\\ {\cite{li2016rain}}\end{tabular} & Total & \begin{tabular}[c]{@{}c@{}}Rain200L\\ {\cite{yang2017deep}}\end{tabular}          & \begin{tabular}[c]{@{}c@{}}Rain200H\\ {\cite{yang2017deep}}\end{tabular}          & \begin{tabular}[c]{@{}c@{}}DID-Data\\ {\cite{fu2017removing}}\end{tabular}          & \begin{tabular}[c]{@{}c@{}}DDN-Data\\ {\cite{fu2017removing}}\end{tabular}          & \begin{tabular}[c]{@{}c@{}}SPA-Data\\ {\cite{wang2019spatial}}\end{tabular}          \\ \hline
				Train Samples   & 700              & 1800              & 0                 & 11,200            & 0                 & 12     & 13712 & 1800              & 1800              & 12,000            & 12,600            & 638,492           \\
				Test Samples    & 100              & 100               & 100               & 2800              & 1200              & 0      & 4300  & 200               & 200               & 1,200             & 1,400             & 1,000             \\
				\textbf{Name}   & \textbf{Test100} & \textbf{Rain100H} & \textbf{Rain100L} & \textbf{Test2800} & \textbf{Test1200} & -      & -     & \textbf{Rain200L} & \textbf{Rain200H} & \textbf{DID-Data} & \textbf{DDN-Data} & \textbf{SPA-Data} \\ \hlinew{1.0pt}
			\end{tabular}
		}
		\label{table3}		
	\end{table*}
	
\begin{figure}[!t]
		\centering	\includegraphics[width=1.0\columnwidth]{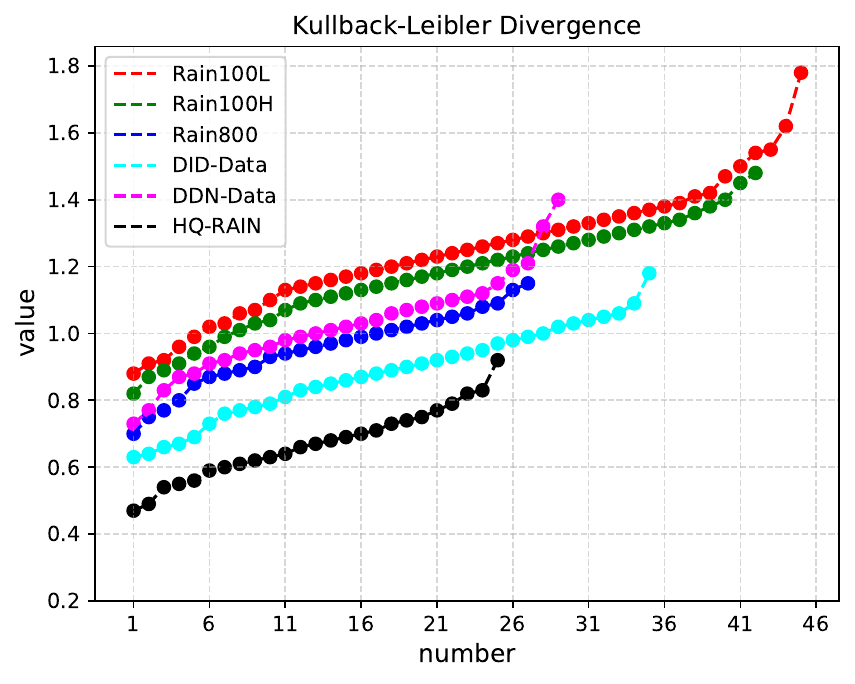}
		\caption{Comparison of Kullback CLeibler divergence (KLD) between different synthetic datasets and one real dataset. The vertical axis represents the value of KLD, and the horizontal axis represents the number of samples. Obviously, our proposed HQ-RAIN obtains a lowest KLD score, indicating that our dataset has a smaller domain gap between the synthetic and real-world images compared to others.}
		\label{fig5}
	\end{figure}

	{\flushleft\textbf{Objective Assessment}}.
	In addition to subjective assessment, we also conduct objective comparisons to verify the high quality of our proposed dataset.
	Here, we adopt the Kullback-Leibler Divergence (KLD) \cite{joyce2011kullback}, also known as relative entropy, to measure the difference between two probability distributions (\emph{i.e.}, the synthetic image and real-world image).
	Figure \ref{fig5} presents the comparison results of the representative synthetic benchmarks \cite{yang2017deep, li2019single, zhang2018density, fu2017removing} and our benchmark, showing that our HQ-RAIN are close to the distribution of real-world rainy images.
	The reason behind this is that HQ-RAIN fully considers the harmony of the synthesized rainy images, thereby narrowing the domain gap between synthetic and real images.
	We hope our benchmark can provide new impetus for future research.
	
	\section{Comprehensive Evaluations and Analysis}
	\label{sec:performance}
	In this section, we provide empirical analysis and conduct extensive evaluations of representative methods on several datasets and our proposed benchmark. In addition, we develop an online platform to
	provide the off-the-shelf toolkit, involving the large-scale performance evaluation.
	
	\subsection{Track Establishment and Evaluated Methods}
	For a comprehensive performance evaluation, we first select previously representative benchmarks to participate in our survey.
	Faced with dozens of existing benchmarks, it is not rare that we feel confused about which dataset to choose for the experiment at hand.
	In fact, our goal is to enable models to learn better generalization from representative datasets.
	To this end, we perform cross-domain generalizability validation on the commonly used datasets using same deep model Restormer \cite{zamir2022restormer}.
	Table~\ref{table4} shows that methods trained on the Rain13K \cite{jiang2020multi} generalizes to an unseen samples well, due to the data diversity.
	In addition, a single dataset can also achieve the best results in certain specific scenarios, such as DDN-Data $\rightarrow$ RainDS-Real-RS.
	In other words, mixed datasets have advantages in comprehensive deraining performance, while single dataset has advantages in image-specific (\emph{e.g.}, heavy rain) deraining performance.

	\begin{table}[!t]
		\centering
		\caption{Cross-domain generalizability analysis. Training Set $\rightarrow$ Testing Set means training on the dataset X and testing on the dataset Y.}
		\resizebox{1.0\columnwidth}{!}{
			\begin{tabular}{lcccc}
				\hline
				Training Set $\rightarrow$ Testing Set & PSNR  & SSIM   & Avg.PSNR               & Avg.SSIM                \\ \hline
				- $\rightarrow$ RainDS-Syn-RS          & 22.47 & 0.6765 & \multirow{2}{*}{23.02} & \multirow{2}{*}{0.6640} \\
				- $\rightarrow$ RainDS-Real-RS         & 23.58 & 0.6515 &                        &                         \\
				Rain100L $\rightarrow$ RainDS-Syn-RS   & 23.41 & 0.7026 & \multirow{2}{*}{23.33} & \multirow{2}{*}{0.6754} \\
				Rain100L $\rightarrow$ RainDS-Real-RS  & 23.25 & 0.6483 &                        &                         \\
				Rain100H $\rightarrow$ RainDS-Syn-RS   & 26.19 & 0.8502 & \multirow{2}{*}{24.90} & \multirow{2}{*}{0.7544} \\
				Rain100H $\rightarrow$ RainDS-Real-RS  & 23.62 & 0.6587 &                        &                         \\
				Rain200L $\rightarrow$ RainDS-Syn-RS   & 23.48 & 0.7085 & \multirow{2}{*}{23.40} & \multirow{2}{*}{0.6796} \\
				Rain200L $\rightarrow$ RainDS-Real-RS  & 23.32 & 0.6507 &                        &                         \\
				Rain200H $\rightarrow$ RainDS-Syn-RS   & 26.33 & 0.8599 & \multirow{2}{*}{25.00} & \multirow{2}{*}{0.7600} \\
				Rain200H $\rightarrow$ RainDS-Real-RS  & 23.68 & 0.6601 &                        &                         \\
				DID-Data $\rightarrow$ RainDS-Syn-RS   & 26.93 & 0.7699 & \multirow{2}{*}{25.56} & \multirow{2}{*}{0.7175} \\
				DID-Data $\rightarrow$ RainDS-Real-RS  & 24.19 & 0.6652 &                        &                         \\
				DDN-Data $\rightarrow$ RainDS-Syn-RS   & 29.02 & 0.8193 & \multirow{2}{*}{26.88} & \multirow{2}{*}{0.7543} \\
				DDN-Data $\rightarrow$ RainDS-Real-RS  & \textbf{24.74} & \textbf{0.6893} &                        &                         \\
				SPA-Data $\rightarrow$ RainDS-Syn-RS   & 22.65 & 0.7226 & \multirow{2}{*}{23.29} & \multirow{2}{*}{0.6924} \\
				SPA-Data $\rightarrow$ RainDS-Real-RS  & 23.93 & 0.6623 &                        &                         \\
				GT-RAIN $\rightarrow$ RainDS-Syn-RS    & 22.33 & 0.7088 & \multirow{2}{*}{22.80} & \multirow{2}{*}{0.6919} \\
				GT-RAIN $\rightarrow$ RainDS-Real-RS   & 23.28 & 0.6750 &                        &                         \\
				Rain13K $\rightarrow$ RainDS-Syn-RS    & \textbf{30.83} & \textbf{0.8890} & \multirow{2}{*}{\textbf{27.78}} & \multirow{2}{*}{\textbf{0.7889}} \\
				Rain13K $\rightarrow$ RainDS-Real-RS   & 24.73 & 0.6889 &                        &                         \\ \hline
			\end{tabular}
		}
		\label{table4}
	\end{table}
	
	To intuitively compare the deraining results under these two different training modes, we create two main tracks, \emph{i.e.}, mixed training track and independent training track.
	For mixed training track, we adopt Rain13K benchmark \cite{jiang2020multi} as the track participant.
	For independent training track, we find that Rain200L/H can help method generalize well than Rain100L/H, because Rain200L/H avoids the problem of duplicate image backgrounds in the training and testing sets in the old version.
	According to the generalization gain in Table~\ref{table4}, we finally adopt Rain200L/H \cite{yang2017deep}, DID-Data \cite{fu2017removing}, DDN-Data \cite{fu2017removing} and SPA-Data \cite{wang2019spatial} as this track participants.
	Note that currently these two tracks only consider general rain removal, and do not include the datasets created for new tasks, such as RainKITTI2012/2015 \cite{zhang2020beyond} in stereo image deraining.
	In each track, the usage setting of training and testing datasets, as well as the measurement criteria, are the same. The detailed usage descriptions are tabulated in Table~\ref{table3}.
	In what follows, we will report the benchmarking results of representative methods on these two tracks.
	
	\begin{table*}[!t]
		\centering
		\caption{Quantitative comparisons on the Rain13K mixed benchmark dataset. Top $1_{s t}$ and $2_{nd}$ results are marked in \textcolor{red}{red} and \textcolor{blue}{blue} respectively.}
		\resizebox{1.0\textwidth}{!}{
			\begin{tabular}{c|cccccccccc||cc}
				\hlinew{1.0pt}
				\multirow{2}{*}{\textbf{Methods}} & \multicolumn{2}{c}{\textbf{Test100} \cite{li2019single}} & \multicolumn{2}{c}{\textbf{Rain100H} \cite{yang2017deep}} & \multicolumn{2}{c}{\textbf{Rain100L} \cite{yang2017deep}} & \multicolumn{2}{c}{\textbf{Test2800} \cite{fu2017removing}} & \multicolumn{2}{c}{\textbf{Test1200} \cite{fu2017removing}} & \multicolumn{2}{c}{\textbf{Average}} \\
				& PSNR $\uparrow$             & SSIM $\uparrow$            & PSNR $\uparrow$             & SSIM $\uparrow$             & PSNR $\uparrow$             & SSIM $\uparrow$             & PSNR $\uparrow$             & SSIM $\uparrow$             & PSNR $\uparrow$              & SSIM $\uparrow$             & PSNR $\uparrow$             & SSIM $\uparrow$            \\ \hline
				Inputs                        & 22.54             & 0.686            & 13.55             & 0.378             & 26.90             & 0.838             & 24.35             & 0.782             & 23.63              & 0.732             & 22.19             & 0.683            \\
				DerainNet \cite{fu2017clearing}                        & 22.77             & 0.810            & 14.92             & 0.592             & 27.03             & 0.884             & 24.31             & 0.861             & 23.38              & 0.835             & 22.48             & 0.796            \\
				SIRR \cite{wei2019semi}                             & 22.35             & 0.788            & 16.56             & 0.486             & 25.03             & 0.842             & 24.43             & 0.782             & 26.05              & 0.822             & 22.88             & 0.744            \\
				DIDMDN \cite{zhang2018density}                           & 22.56             & 0.818            & 17.35             & 0.524             & 25.23             & 0.741             & 28.13             & 0.867             & 29.95              & 0.901             & 24.58             & 0.770            \\
				UMRL \cite{yasarla2019uncertainty}                             & 24.41             & 0.829            & 26.01             & 0.832             & 29.18             & 0.923             & 29.97             & 0.905             & 30.55              & 0.910             & 28.02             & 0.880            \\
				RESCAN \cite{li2018recurrent}                           & 25.00             & 0.835            & 26.36             & 0.786             & 29.80             & 0.881             & 31.29             & 0.904             & 30.51              & 0.882             & 28.59             & 0.857            \\
				PReNet \cite{ren2019progressive}                           & 24.81             & 0.851            & 26.77             & 0.858             & 32.44             & 0.950             & 31.75             & 0.916             & 31.36              & 0.911             & 29.42             & 0.897            \\
				MSPFN \cite{jiang2020multi}                            & 27.50             & 0.876            & 28.66             & 0.860             & 32.40             & 0.933             & 32.82             & 0.930             & 32.39              & 0.916             & 30.75             & 0.903            \\
				MPRNet \cite{zamir2021multi}                           & 30.27             & 0.897            & 30.41             & 0.890             & 36.40             & 0.965             & 33.64             & 0.938             & 32.91              & 0.916             & 32.73             & 0.921            \\		
				HINet \cite{chen2021hinet}                           & 30.26             & 0.905            & 30.63             & 0.893             & 37.20             & 0.969             & 33.87             & 0.940             & 33.01              & 0.918             & 33.00             & 0.925            \\			
				SPAIR \cite{purohit2021spatially}                            & 30.35             & 0.909            & 30.95             & 0.892             & 36.93             & 0.969             & 33.34             & 0.936             & 33.04              & \textcolor{blue}{0.922}             & 32.91             & 0.926            \\
				KiT \cite{lee2022knn}                              & 30.26             & 0.904            & 30.47             & 0.897             & 36.65             & 0.969             & 33.85             & 0.941             & 32.81              & 0.918             & 32.81             & 0.929            \\
				DGUNet \cite{mou2022deep}                           & 30.32             & 0.899            & 30.66             & 0.891             & 37.42             & 0.969             & 33.68             & 0.938             & \textcolor{red}{33.23}              & 0.920             & 33.06             & 0.923            \\
				DGUNet+ \cite{mou2022deep}                          & 30.86             & 0.907            & \textcolor{blue}{31.06}             & 0.897             & \textcolor{blue}{38.25}             & 0.974             & \textcolor{blue}{34.01}             & 0.942             & 33.08              & 0.916             & \textcolor{blue}{33.46}             & 0.927            \\
				MAXIM-2S \cite{tu2022maxim}                         & \textcolor{blue}{31.17}             & \textcolor{blue}{0.922}            & 30.81             & \textcolor{blue}{0.903}             & 38.06             & \textcolor{blue}{0.977}             & 33.80             & \textcolor{blue}{0.943}             & 32.37              & \textcolor{blue}{0.922}             & 33.24             & \textcolor{blue}{0.933}            \\
				Restormer \cite{zamir2022restormer}                        & \textcolor{red}{32.00}             & \textcolor{red}{0.923}            & \textcolor{red}{31.46}             & \textcolor{red}{0.904}             & \textcolor{red}{38.99}             & \textcolor{red}{0.978}             & \textcolor{red}{34.18}             & \textcolor{red}{0.944}             & \textcolor{blue}{33.19}              & \textcolor{red}{0.926}             & \textcolor{red}{33.96}             & \textcolor{red}{0.935}            \\ \hlinew{1.0pt}
			\end{tabular}
		}
		\label{table5}		
	\end{table*}
	
	\begin{figure*}[!t]
		\centering 	
		\begin{subfigure}[t]{0.19\textwidth}
			\centering
			\includegraphics[width=\textwidth]{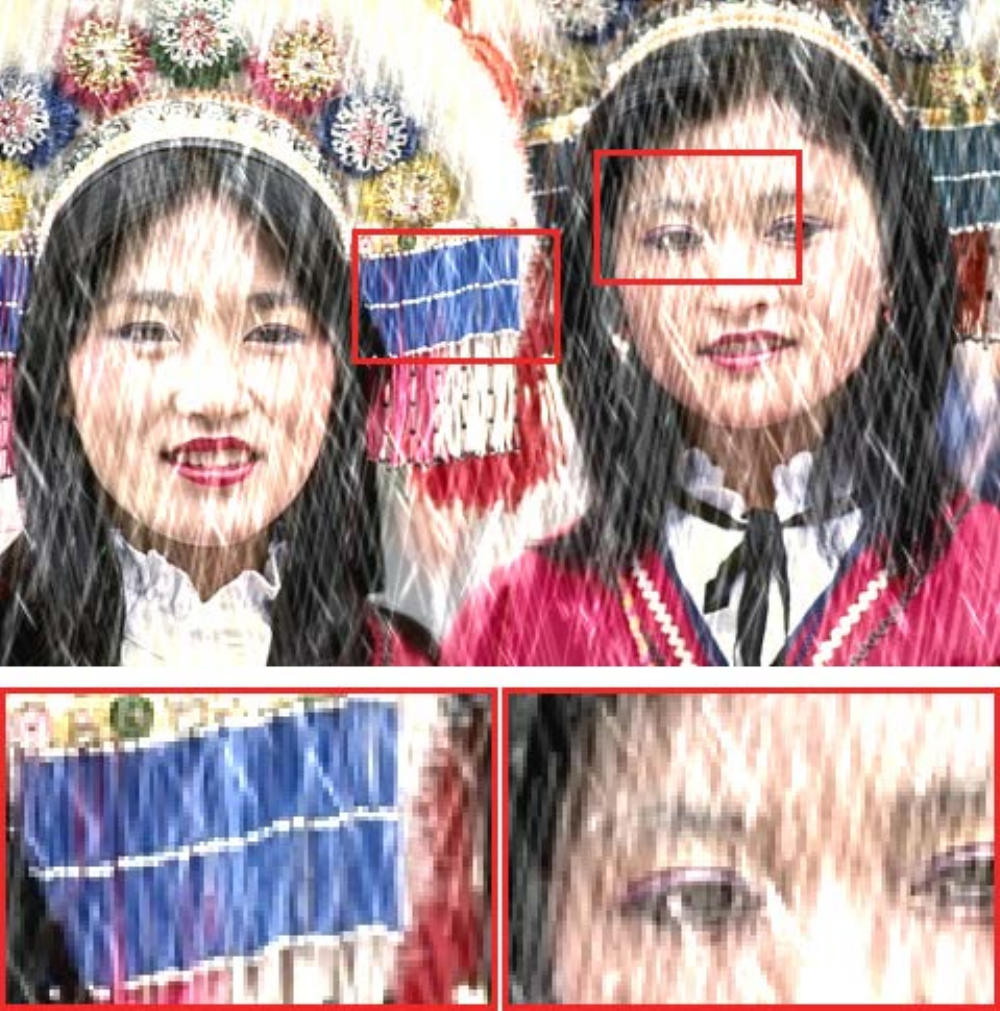}
			\caption{Rainy Input}
		\end{subfigure}
		\begin{subfigure}[t]{0.19\textwidth}
			\centering
			\includegraphics[width=\textwidth]{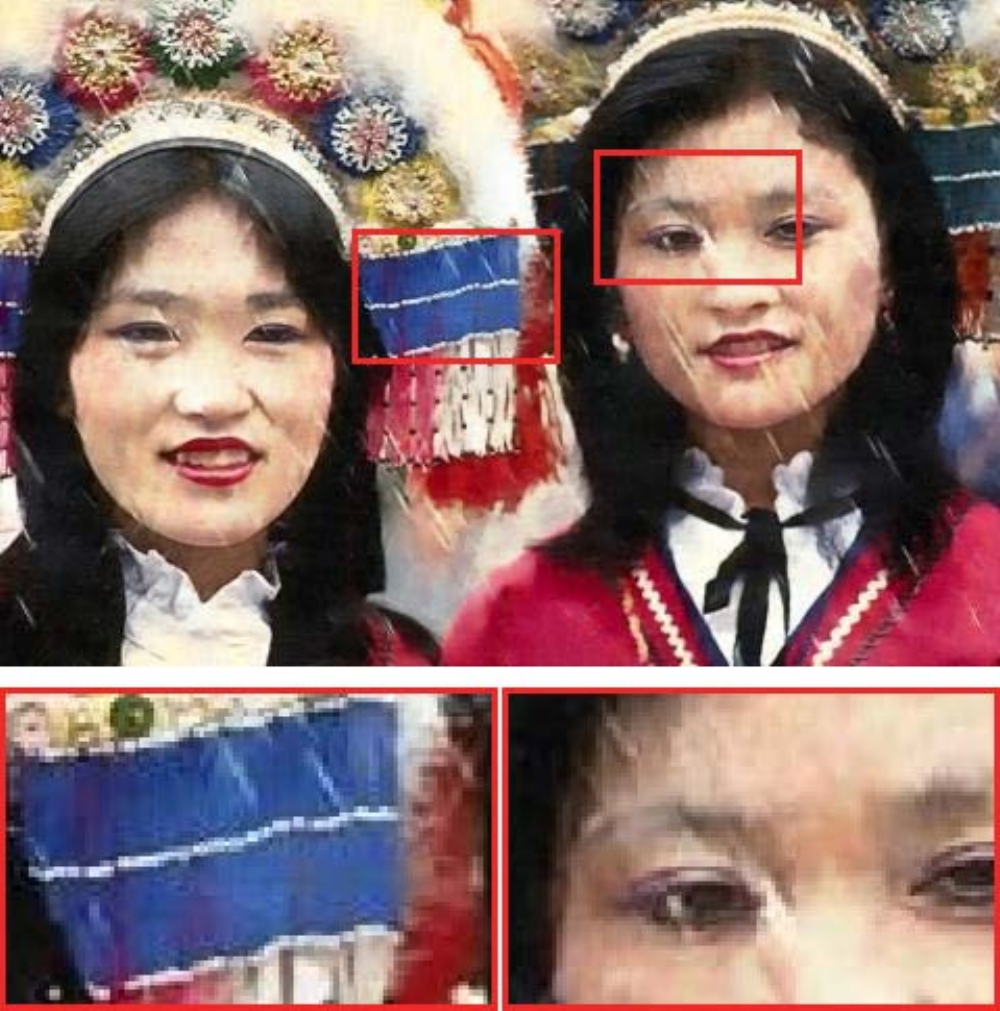}
			\caption{RESCAN}
		\end{subfigure}
		\begin{subfigure}[t]{0.19\textwidth}
			\centering
			\includegraphics[width=\textwidth]{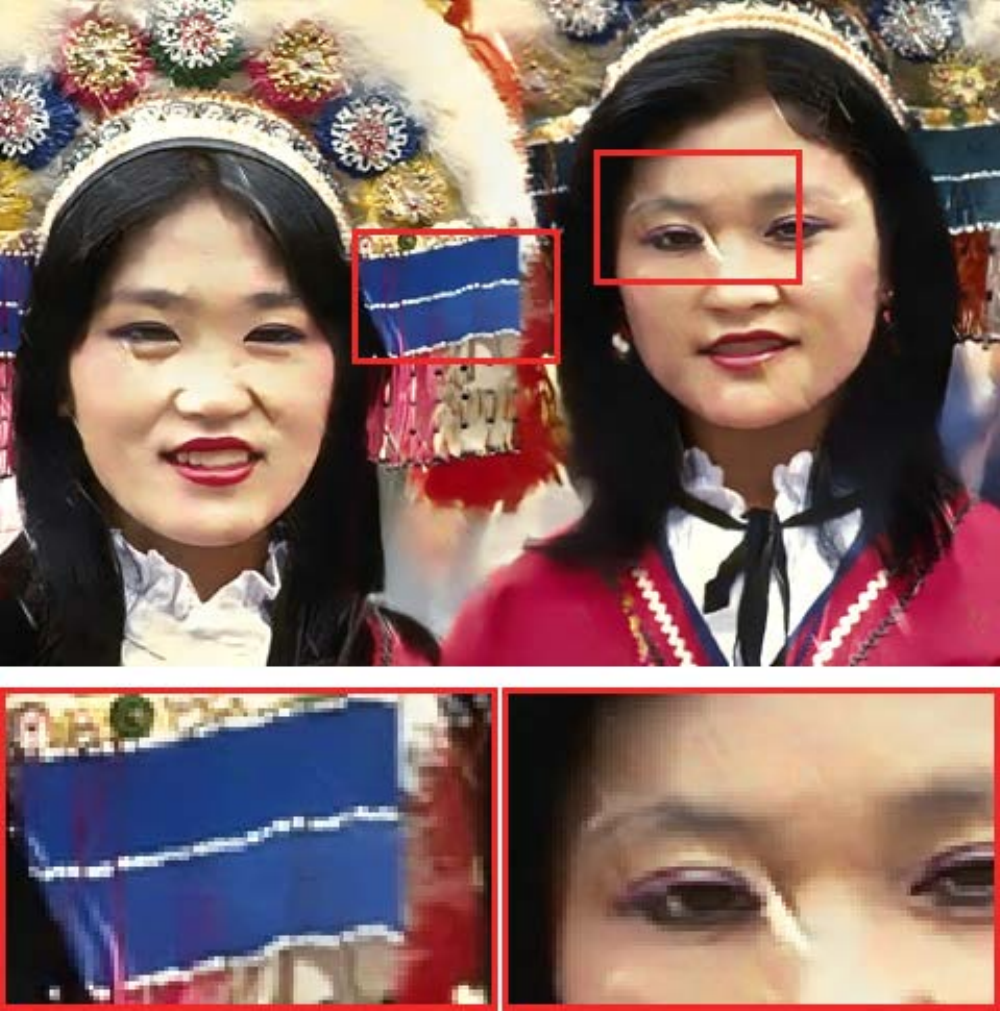}
			\caption{PReNet}
		\end{subfigure}	
		\begin{subfigure}[t]{0.19\textwidth}
			\centering
			\includegraphics[width=\textwidth]{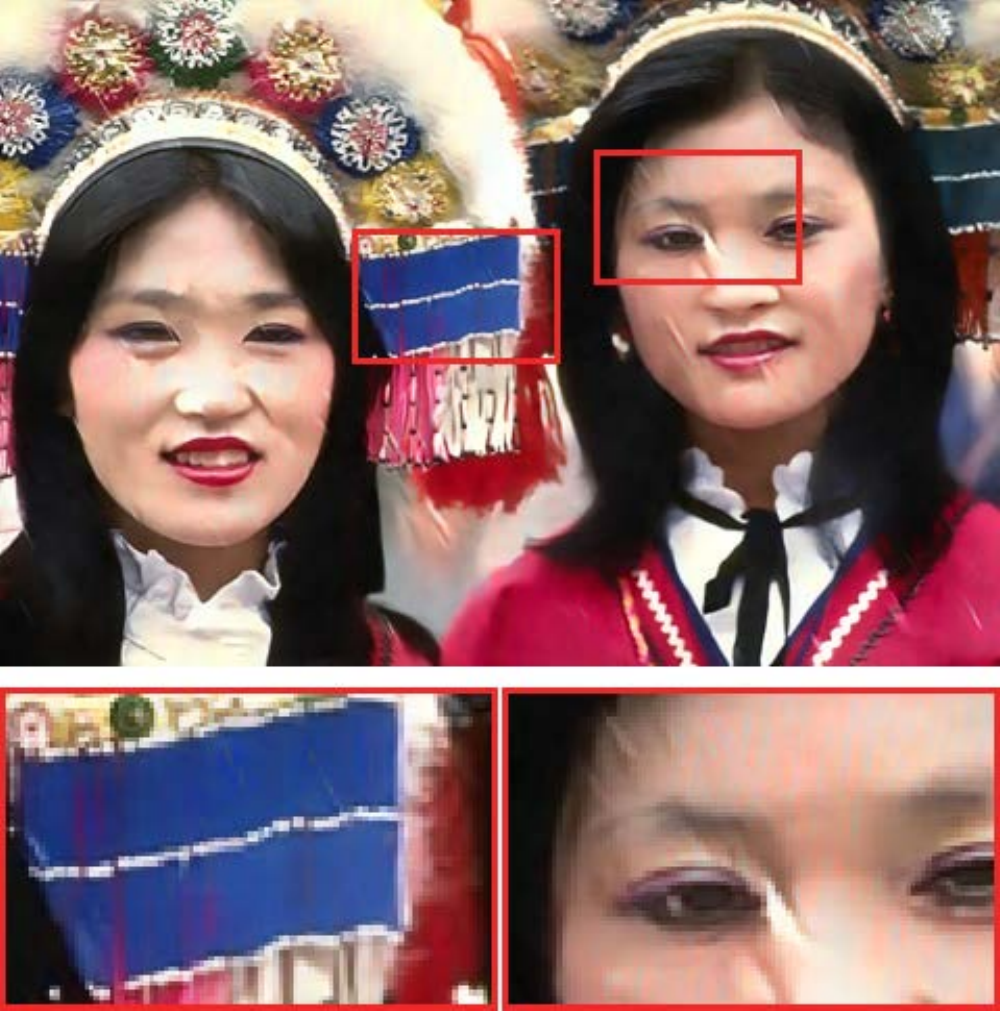}
			\caption{MSPFN}
		\end{subfigure}	
		\begin{subfigure}[t]{0.19\textwidth}
			\centering
			\includegraphics[width=\textwidth]{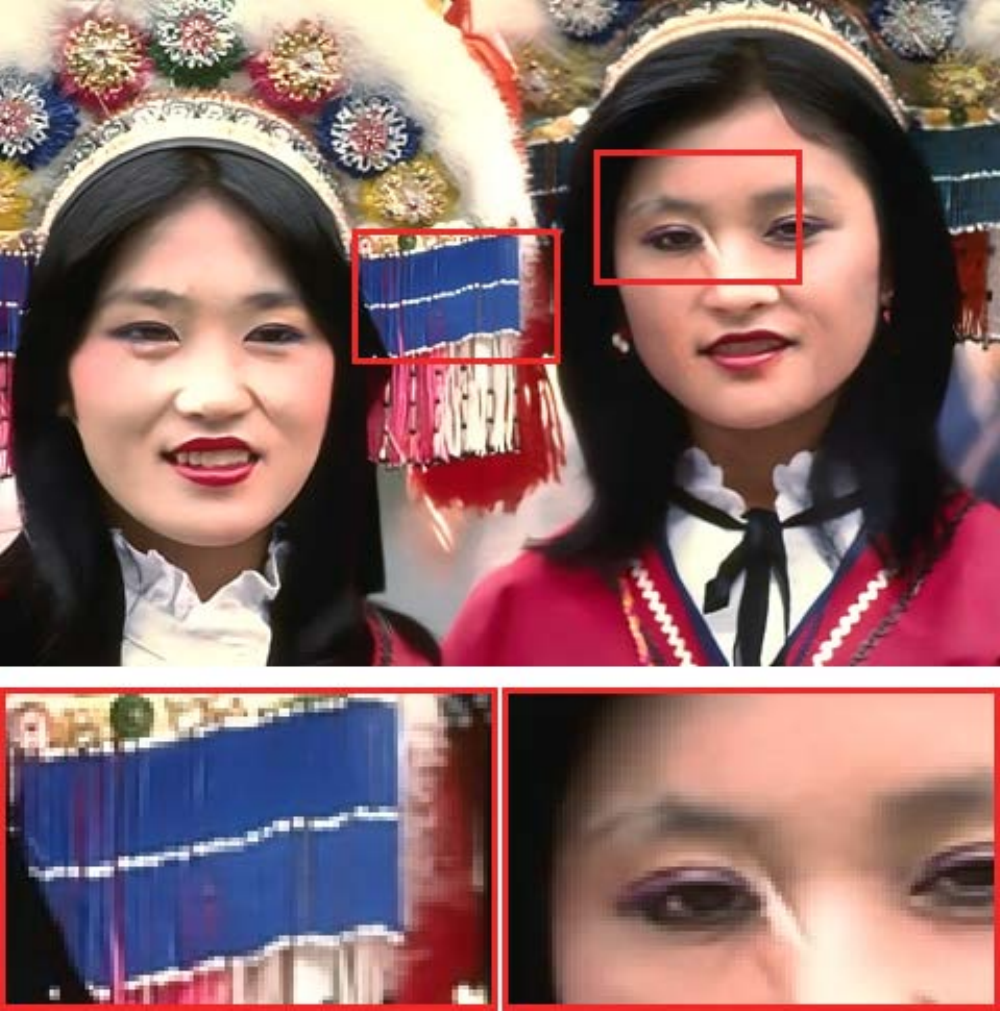}
			\caption{MPRNet}
		\end{subfigure}
		\\
		\begin{subfigure}[t]{0.19\textwidth}
			\centering
			\includegraphics[width=\textwidth]{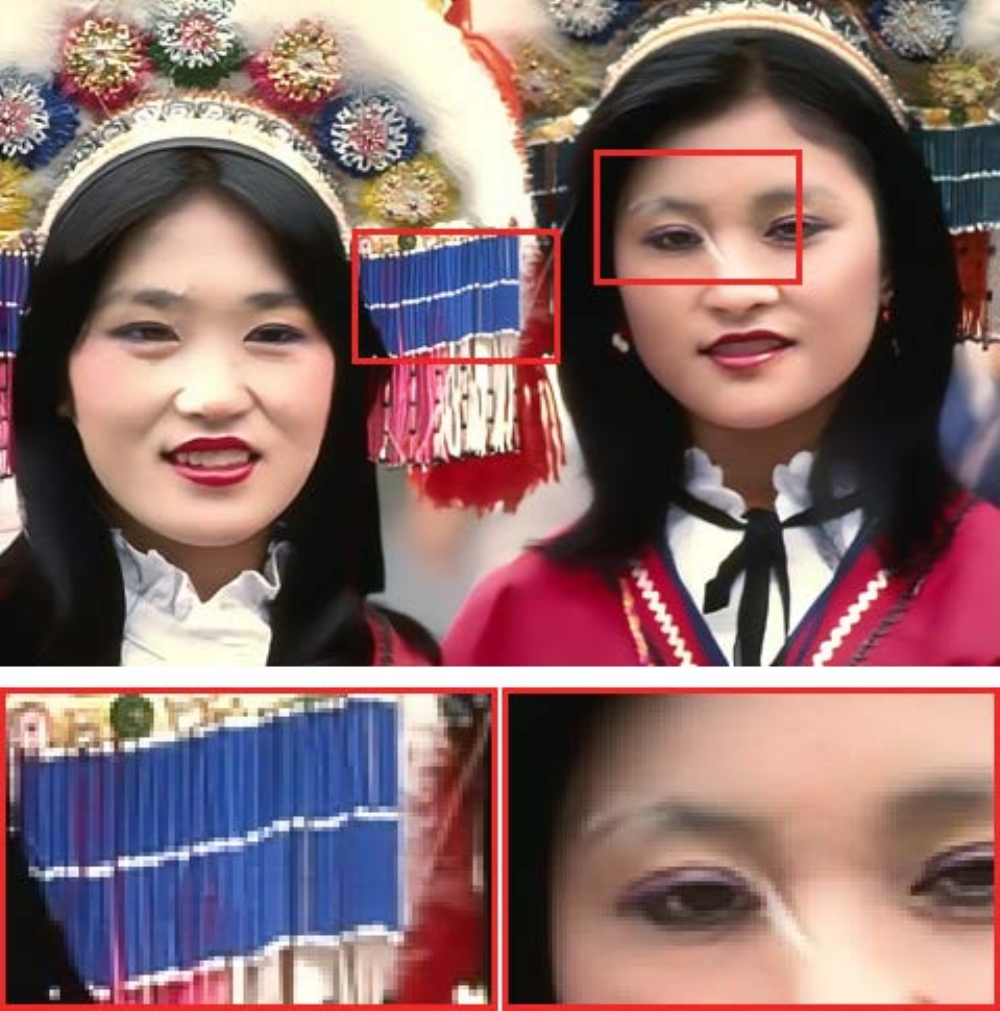}
			\caption{DGUNet}
		\end{subfigure}	
		\begin{subfigure}[t]{0.19\textwidth}
			\centering
			\includegraphics[width=\textwidth]{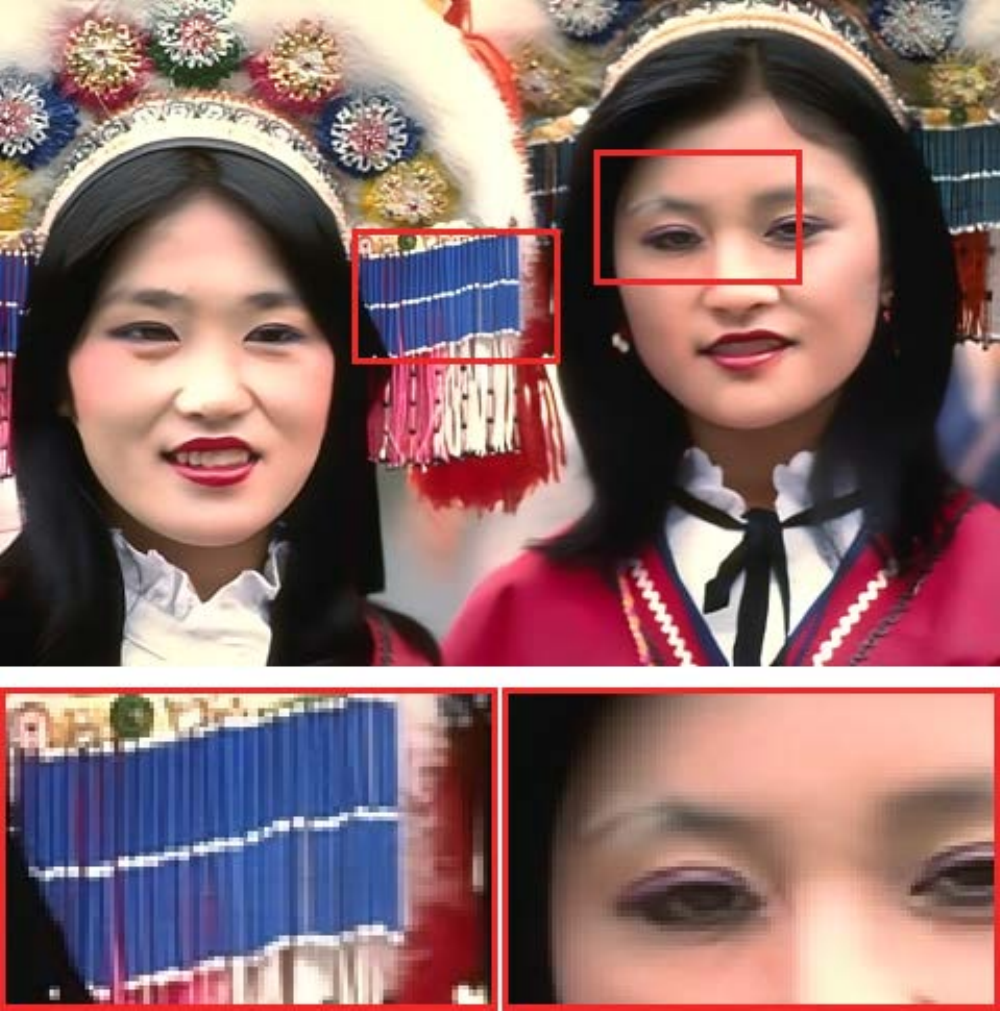}
			\caption{DGUNet+}
		\end{subfigure}
		\begin{subfigure}[t]{0.19\textwidth}
			\centering
			\includegraphics[width=\textwidth]{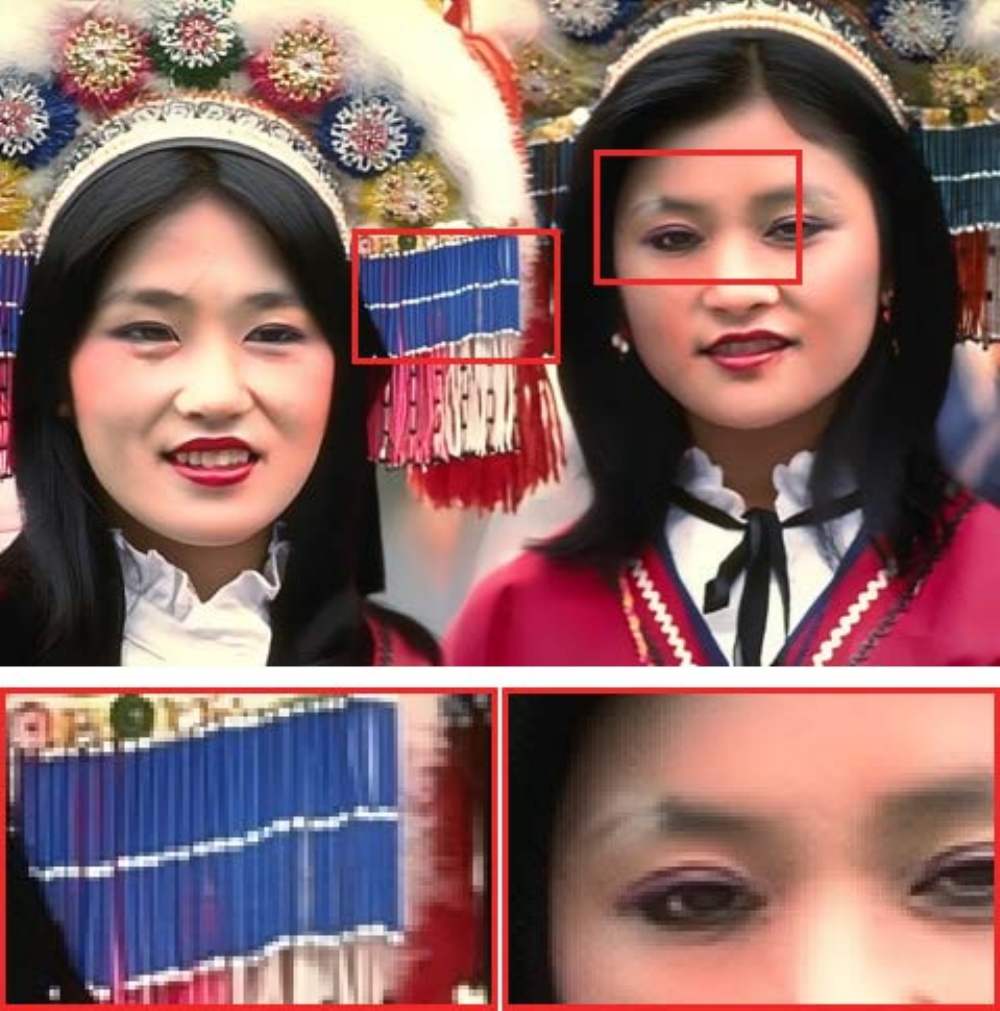}
			\caption{MAXIM-2S}
		\end{subfigure}
		\begin{subfigure}[t]{0.19\textwidth}
			\centering
			\includegraphics[width=\textwidth]{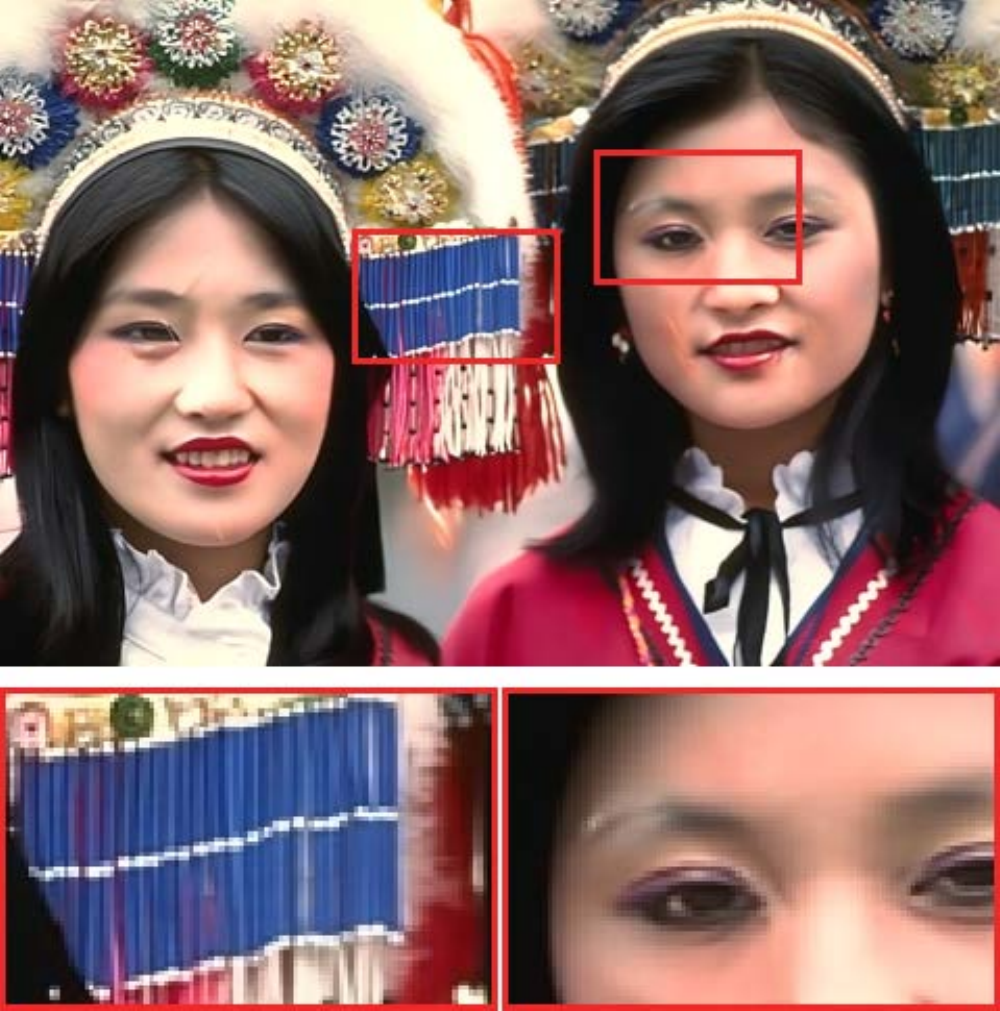}
			\caption{Restormer}
		\end{subfigure}	
		\begin{subfigure}[t]{0.19\textwidth}
			\centering
			\includegraphics[width=\textwidth]{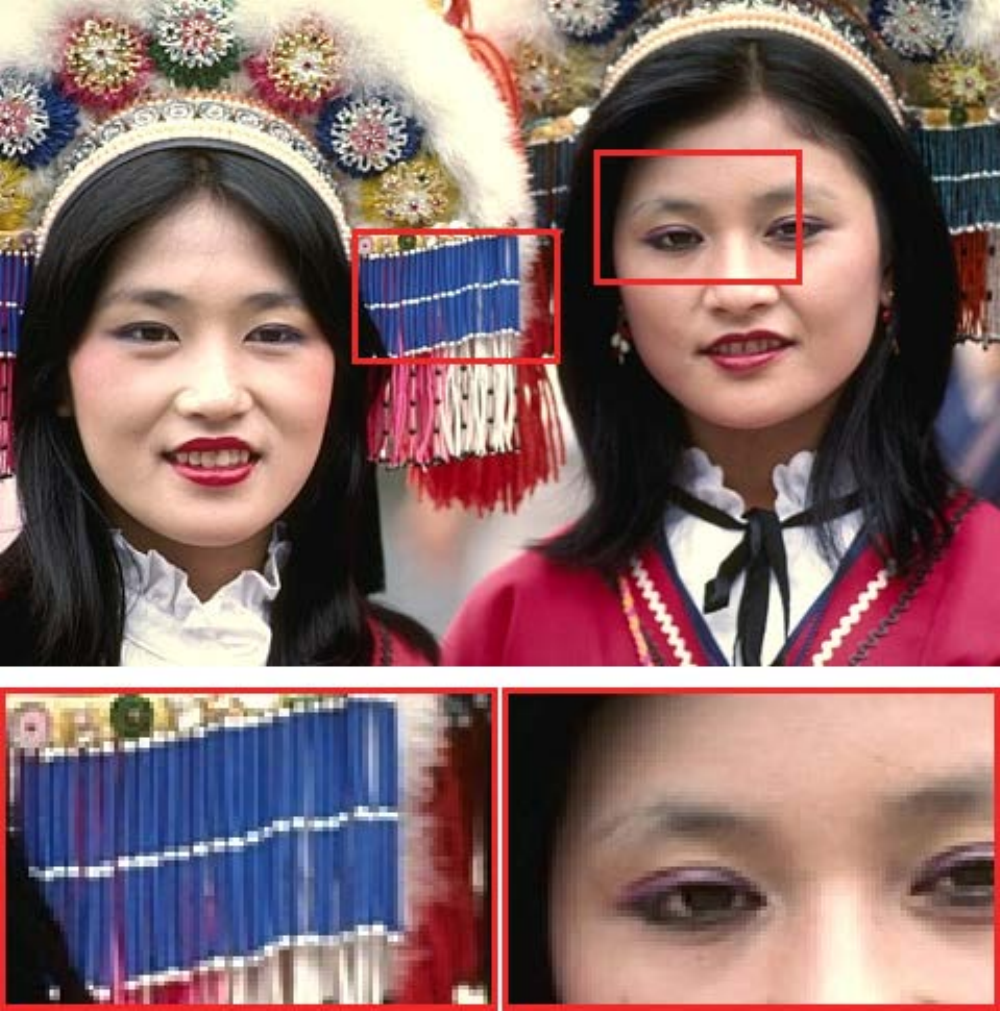}
			\caption{Ground Truth}
		\end{subfigure}	
		\caption{Visual quality comparison of mixed training track on the Rain100H dataset. Please zoom in the figures for better view of the rain removal and detail recovery.}
		\label{fig6}
	\end{figure*}
	
	\subsection{Evaluation on Mixed Training Track}
	In this track, a pretrained model is obtained to evaluate multiple testing sets, including Test100 \cite{li2019single}, Rain100H \cite{yang2017deep}, Rain100L \cite{yang2017deep}, Test2800 \cite{fu2017removing}, and Test1200 \cite{fu2017removing}.
	From statistics in Table~\ref{table1}, we find that this track is highly favored by image restoration task, thanks to the convenience of training on this track, which facilitates rapid evaluation of deraining performance.
	Here, we report 15 representative methods in Table \ref{table5}, \emph{i.e.}, DerainNet \cite{fu2017clearing}, SIRR \cite{wei2019semi}, DIDMDN \cite{zhang2018density}, UMRL \cite{yasarla2019uncertainty}, RESCAN \cite{li2018recurrent}, PReNet \cite{ren2019progressive}, MSPFN \cite{jiang2020multi}, MPRNet \cite{zamir2021multi}, HINet \cite{chen2021hinet}, SPAIR \cite{purohit2021spatially}, KiT \cite{lee2022knn}, DGUNet \cite{mou2022deep}, DGUNet+ \cite{mou2022deep}, MAXIM-2S \cite{tu2022maxim}, and Restormer \cite{zamir2022restormer}.
	Each method is evaluated under the same implementation using two well-known used indicators: PSNR \cite{huynh2008scope} and SSIM \cite{wang2004image}.
	
	So far, Restormer \cite{zamir2022restormer} has achieved the state-of-the-art quantitative average results in the mixed training track. In addition, DGUNet+ \cite{mou2022deep} obtained the second best average PSNR value, while MAXIM-2S \cite{tu2022maxim} achieved the second best average SSIM score.
	Furthermore, we also present the visual comparison and the examples are shown in Figure~\ref{fig6}.
	It can be seen that Transformers-based methods (\emph{e.g.}, MAXIM-2S \cite{tu2022maxim} and Restormer \cite{zamir2022restormer}) can achieve significant improvement compared to the previous CNN models, as they can model the non-local information which is vital for high-quality image reconstruction. However, all approaches still need to break through in terms of image detail recovery.
	
	\begin{table*}[t]
		\centering
		\caption{Results of independent training track. Quantitative comparisons on five previous commonly used benchmark datasets. Top $1_{s t}$ and $2_{nd}$ results are marked in \textcolor{red}{red} and \textcolor{blue}{blue} respectively.}
		\resizebox{1.0\textwidth}{!}{
			\begin{tabular}{c|cccccccccc||cc}
				\hlinew{1.0pt}
				\multirow{2}{*}{\textbf{Methods}} & \multicolumn{2}{c}{\textbf{Rain200L} \cite{yang2017deep}} & \multicolumn{2}{c}{\textbf{Rain200H} \cite{yang2017deep}} & \multicolumn{2}{c}{\textbf{DID-Data} \cite{zhang2018density}} & \multicolumn{2}{c}{\textbf{DDN-Data} \cite{fu2017removing}} & \multicolumn{2}{c}{\textbf{SPA-Data} \cite{wang2019spatial}} & \multicolumn{2}{c}{\textbf{Average}} \\
				& PSNR $\uparrow$             & SSIM $\uparrow$             & PSNR $\uparrow$              & SSIM $\uparrow$             & PSNR $\uparrow$             & SSIM $\uparrow$              & PSNR $\uparrow$              & SSIM $\uparrow$              & PSNR $\uparrow$              & SSIM $\uparrow$               & PSNR $\uparrow$             & SSIM $\uparrow$              \\ \hline
				Inputs                            & 26.71             & 0.8438            & 13.08             & 0.3734            & 23.64             & 0.7794            & 25.24             & 0.8098            & 34.14             & 0.9415             & 24.56            & 0.7496            \\
				DDN \cite{fu2017removing}                              & 34.68             & 0.9671            & 26.05             & 0.8056            & 30.97             & 0.9116            & 30.00             & 0.9041            & 36.16             & 0.9457             & 31.57            & 0.9068            \\
				DID-MDN \cite{zhang2018density}                          & 35.40             & 0.9618            & 26.61             & 0.8242            & 31.30             & 0.9207            & 31.49             & 0.9146            & 38.16             & 0.9763             & 32.59            & 0.9195            \\
				RESCAN \cite{li2018recurrent}                           & 36.09             & 0.9697            & 26.75             & 0.8353            & 33.38             & 0.9417            & 31.94             & 0.9345            & 38.11             & 0.9707             & 33.25            & 0.9304            \\
				NLEDN \cite{li2018non}                            & 39.13             & 0.9821            & 29.79             & 0.9005            & 34.68             & 0.9583            & 32.15             & 0.9398            & 42.97             & 0.9835             & 35.74            & 0.9528            \\
				JORDER-E \cite{yang2019joint}                         & 37.25             & 0.9752            & 29.35             & 0.8905            & 33.98             & 0.9502            & 32.01             & 0.9321            & 40.78             & 0.9801             & 34.67            & 0.9456            \\
				ID-CGAN \cite{zhang2019image}                          & 35.19             & 0.9694            & 25.02             & 0.8430            & 30.25             & 0.9217            & 29.06             & 0.9162            & 38.47             & 0.9624             & 31.59            & 0.9225            \\
				SIRR \cite{wei2019semi}                             & 34.75             & 0.9690            & 26.55             & 0.8190            & 30.57             & 0.9104            & 30.01             & 0.9078            & 35.31             & 0.9411             & 31.43            & 0.9095            \\
				PReNet \cite{ren2019progressive}                           & 37.80             & 0.9814            & 29.04             & 0.8991            & 33.17             & 0.9481            & 32.60             & 0.9459            & 40.16             & 0.9816             & 34.55            & 0.9512            \\
				SPANet \cite{wang2019spatial}                           & 35.79             & 0.9653            & 26.27             & 0.8666            & 33.04             & 0.9489            & 29.85             & 0.9117            & 40.24             & 0.9811             & 33.03            & 0.9347            \\
				FBL \cite{yang2020towards}                              & 39.02             & 0.9827            & 30.07             & 0.9021            & 34.26             & 0.9320            & 33.05             & 0.9334            & 42.80             & 0.9824             & 35.84            & 0.9465            \\
				MSPFN \cite{jiang2020multi}                            & 38.58             & 0.9827            & 29.36             & 0.9034            & 33.72             & 0.9550            & 32.99             & 0.9333            & 43.43             & 0.9843             & 35.61            & 0.9517            \\
				RCDNet \cite{wang2020model}                           & 39.17             & 0.9885            & 30.24             & 0.9048            & 34.08             & 0.9532            & 33.04             & 0.9472            & 43.36             & 0.9831             & 35.97            & 0.9554            \\
				Syn2Real \cite{yasarla2020syn2real}                         & 35.92             & 0.9709            & 27.90             & 0.8605            & 32.39             & 0.9390            & 29.29             & 0.9057            & 35.28             & 0.9745             & 32.15            & 0.9301            \\
				SGCN \cite{fu2021successive}                             & 37.95             & 0.9822            & 29.35             & 0.9055            & 33.58             & 0.9477            & 32.14             & 0.9364            & 43.84             & 0.9862             & 35.37            & 0.9516            \\
				MPRNet \cite{zamir2021multi}                           & 39.47             & 0.9825            & 30.67             & 0.9110            & 33.99             & 0.9590            & 33.10             & 0.9347            & 43.64             & 0.9844             & 36.17            & 0.9543            \\
				DualGCN \cite{fu2021rain}                          & 40.73             & 0.9886            & 31.15             & 0.9125            & 34.37             & 0.9620            & 33.01             & 0.9489            & 44.18             & 0.9902             & 36.68            & 0.9604            \\
				SPDNet \cite{yi2021structure}                           & 40.50             & 0.9875            & 31.28             & 0.9207            & 34.57             & 0.9560            & 33.15             & 0.9457            & 43.20             & 0.9871             & 36.54            & 0.9594            \\
				Uformer \cite{wang2022uformer}                          & 40.20             & 0.9860            & 30.80             & 0.9105            & 35.02             & 0.9621            & 33.95             & 0.9545            & 46.13             & 0.9913             & 37.22            & 0.9609            \\
				Restormer \cite{zamir2022restormer}                        & 40.99             & 0.9890            & 32.00             & \textcolor{blue}{0.9329}            & \textcolor{blue}{35.29}             & \textcolor{blue}{0.9641}            & \textcolor{blue}{34.20}             & \textcolor{blue}{0.9571}            & \textcolor{blue}{47.98}             & 0.9921             & \textcolor{blue}{38.09}            & \textcolor{blue}{0.9670}            \\
				IDT \cite{xiao2022image}                              & 40.74             & 0.9884            & \textcolor{blue}{32.10}             & \textcolor{red}{0.9344}            & 34.89             & 0.9623            & 33.84             & 0.9549            & 47.35             & \textcolor{red}{0.9930}             & 37.78            & 0.9666            \\
				HCN \cite{fu2023continual}                              & \textcolor{red}{41.31}             & \textcolor{blue}{0.9892}            & 31.34             & 0.9248            & 34.70             & 0.9613            & 33.42             & 0.9512            & 45.03             & 0.9907             & 37.16            & 0.9634            \\
				DRSformer \cite{chen2023learning}                        & \textcolor{blue}{41.23}             & \textcolor{red}{0.9894}            & \textcolor{red}{32.17}             & 0.9326            & \textcolor{red}{35.35}             & \textcolor{red}{0.9646}            & \textcolor{red}{34.35}             & \textcolor{red}{0.9588}            & \textcolor{red}{48.54}             & \textcolor{blue}{0.9924}             & \textcolor{red}{38.32}            & \textcolor{red}{0.9676}            \\ \hlinew{1.0pt}
			\end{tabular}
		}
		\label{table6}		
	\end{table*}
	
	\begin{figure*}[!t]
		\centering 	
		\begin{subfigure}[t]{0.19\textwidth}
			\centering
			\includegraphics[width=\textwidth]{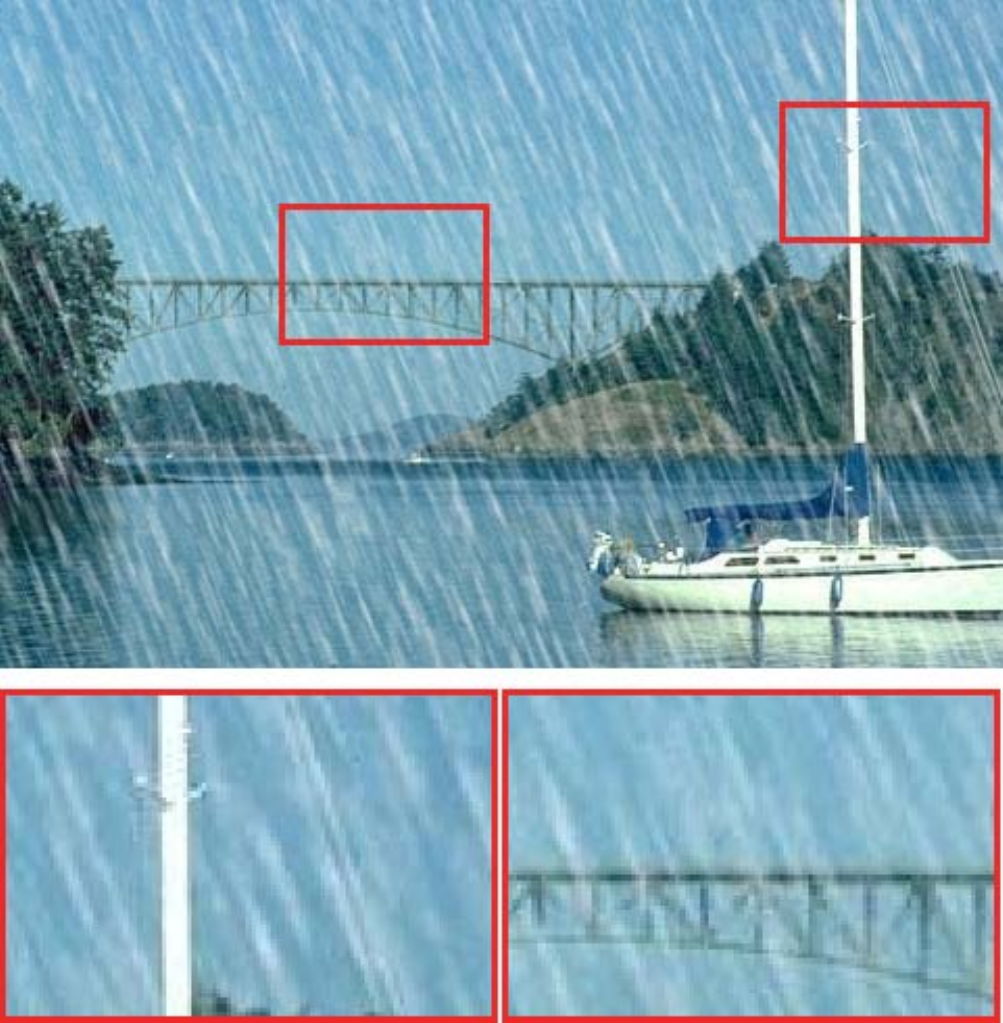}
			\caption{Rainy Input}
		\end{subfigure}
		\begin{subfigure}[t]{0.19\textwidth}
			\centering
			\includegraphics[width=\textwidth]{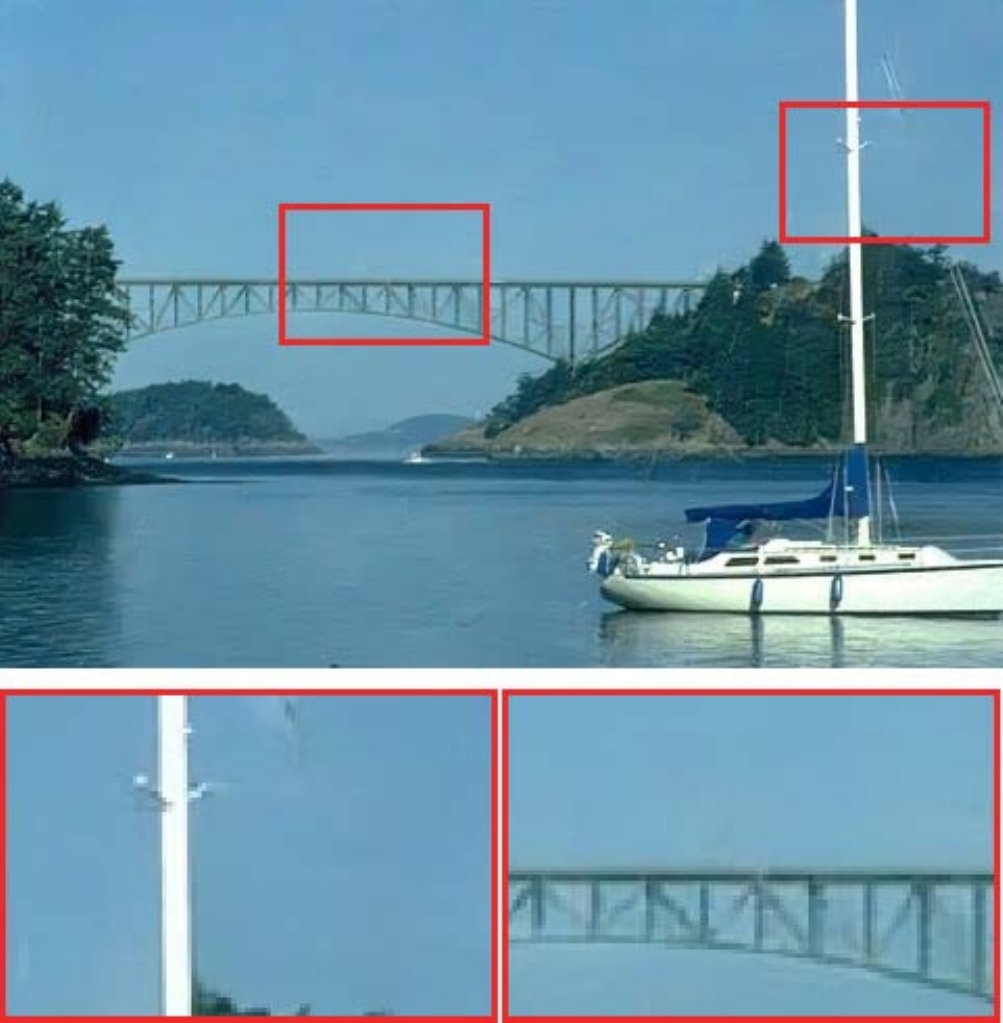}
			\caption{PReNet}
		\end{subfigure}
		\begin{subfigure}[t]{0.19\textwidth}
			\centering
			\includegraphics[width=\textwidth]{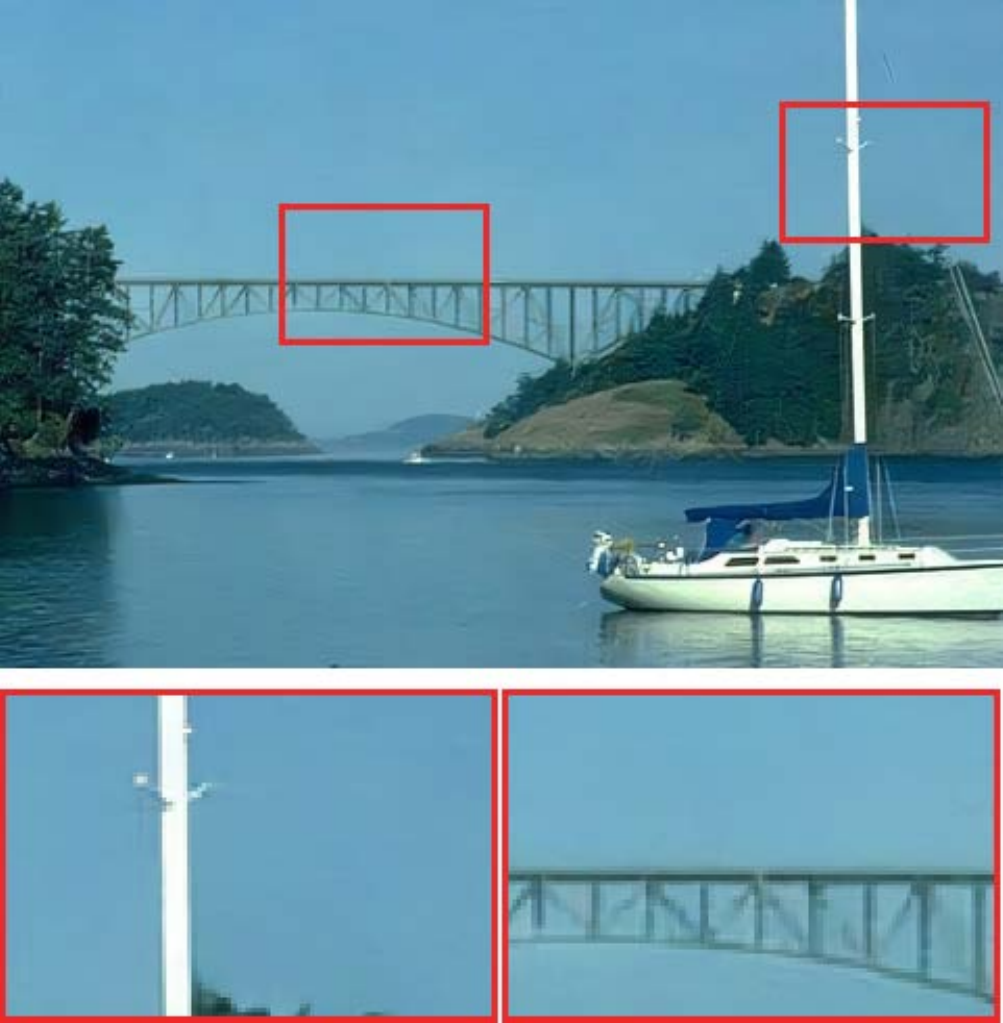}
			\caption{RCDNet}
		\end{subfigure}	
		\begin{subfigure}[t]{0.19\textwidth}
			\centering
			\includegraphics[width=\textwidth]{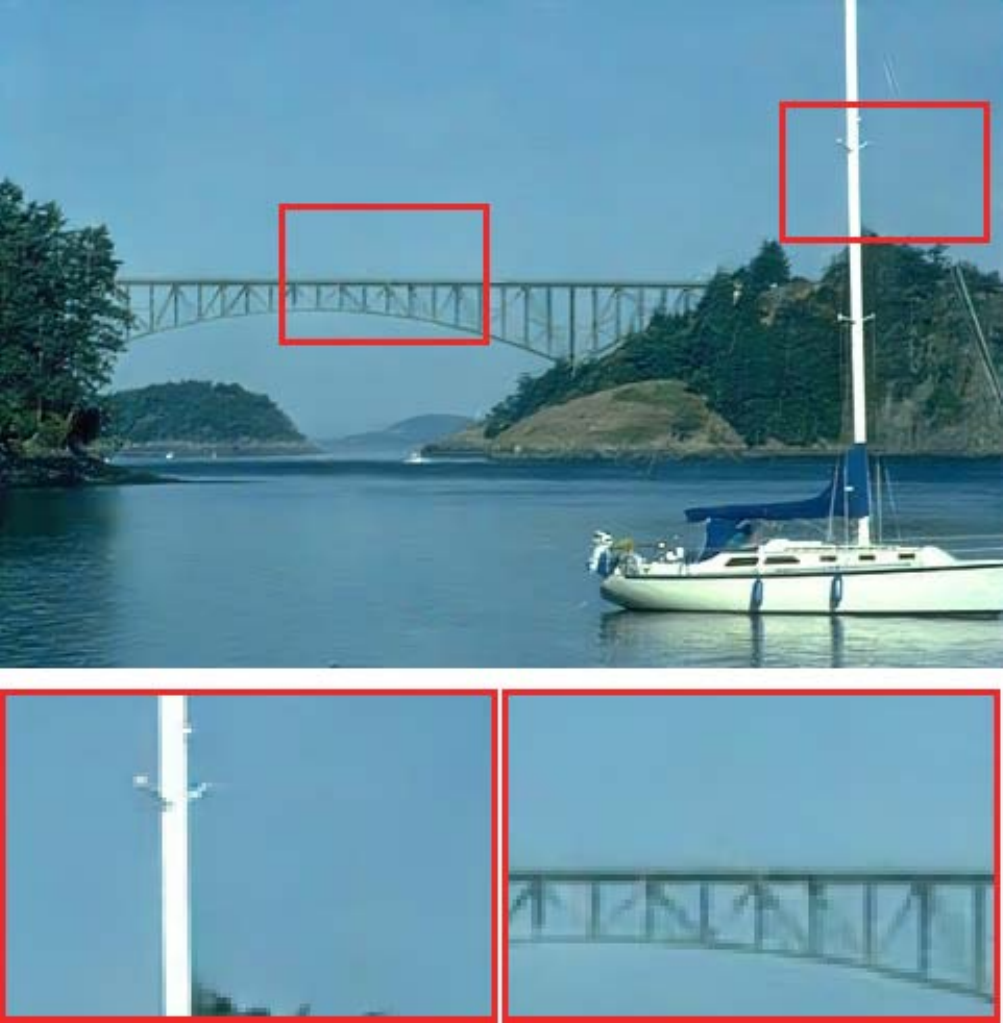}
			\caption{DualGCN}
		\end{subfigure}	
		\begin{subfigure}[t]{0.19\textwidth}
			\centering
			\includegraphics[width=\textwidth]{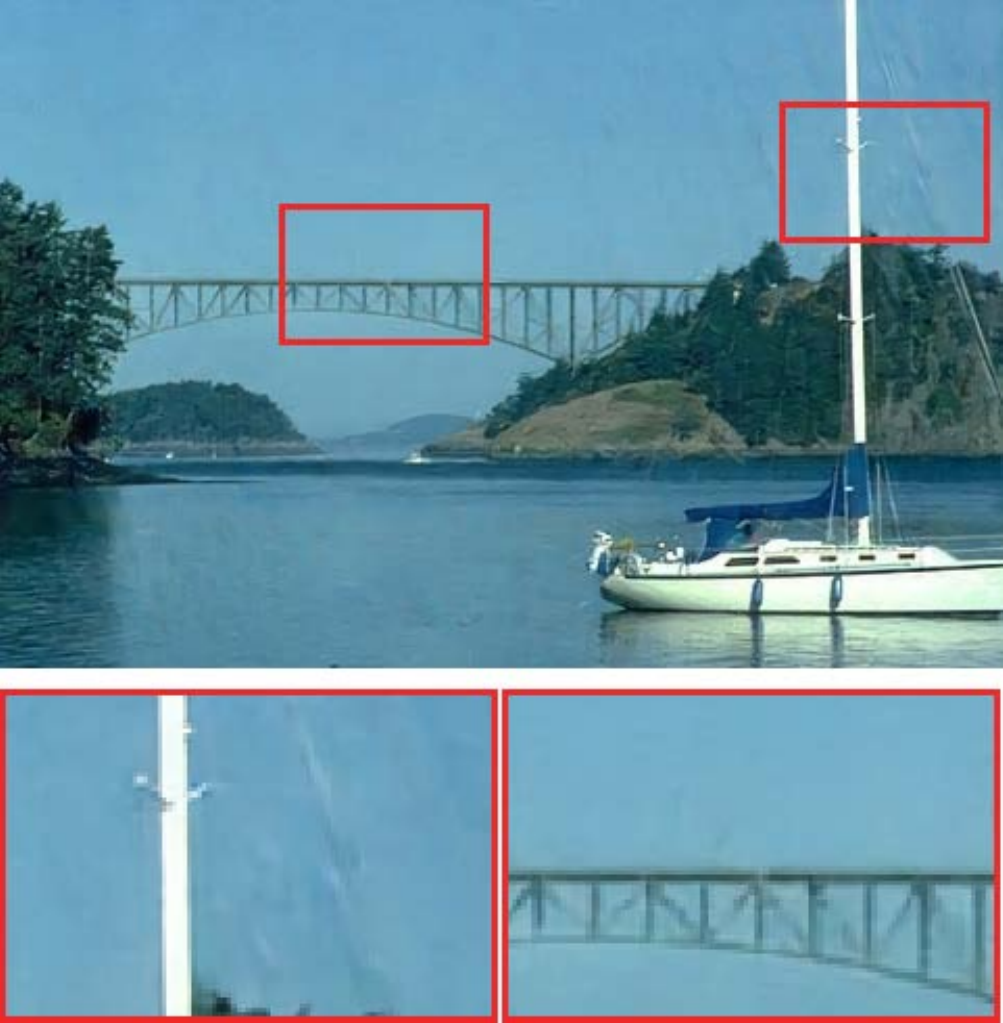}
			\caption{SPDNet}
		\end{subfigure}	
		\\
		\begin{subfigure}[t]{0.19\textwidth}
			\centering
			\includegraphics[width=\textwidth]{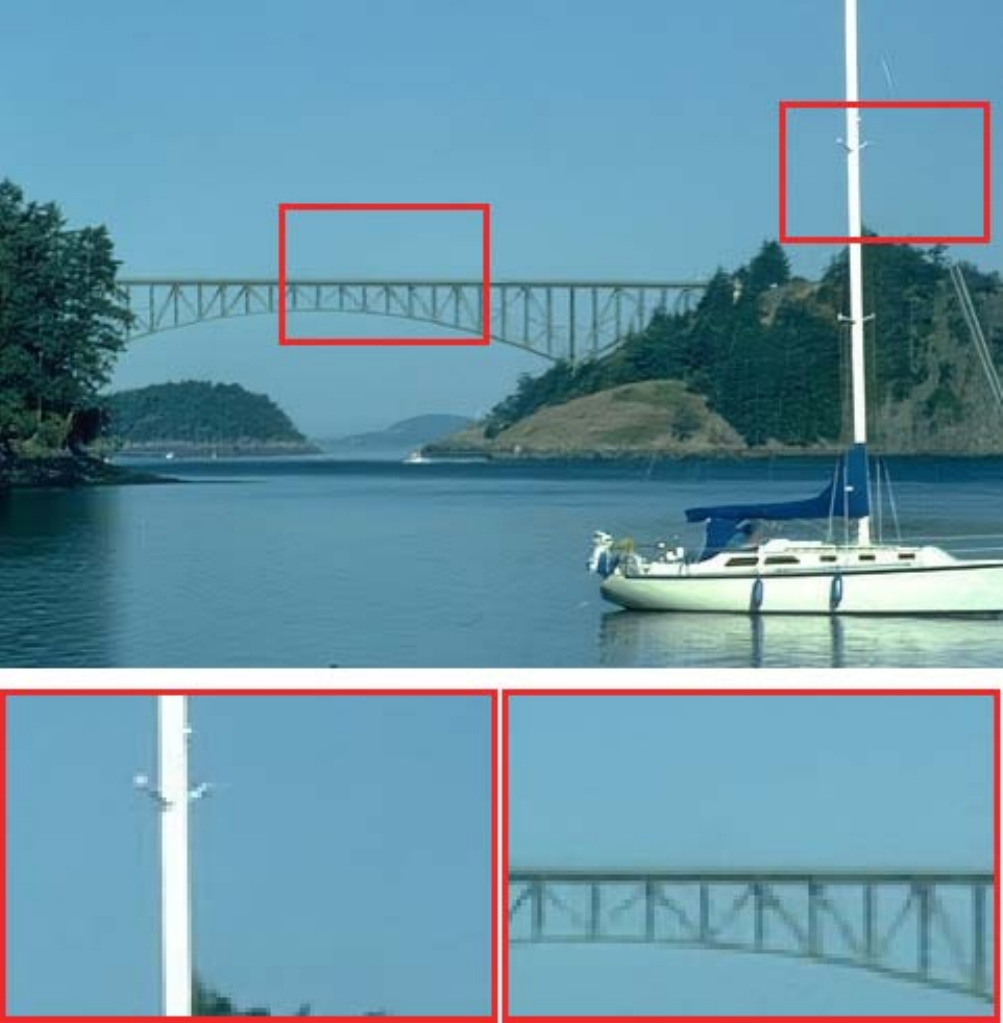}
			\caption{Uformer}
		\end{subfigure}
		\begin{subfigure}[t]{0.19\textwidth}
			\centering
			\includegraphics[width=\textwidth]{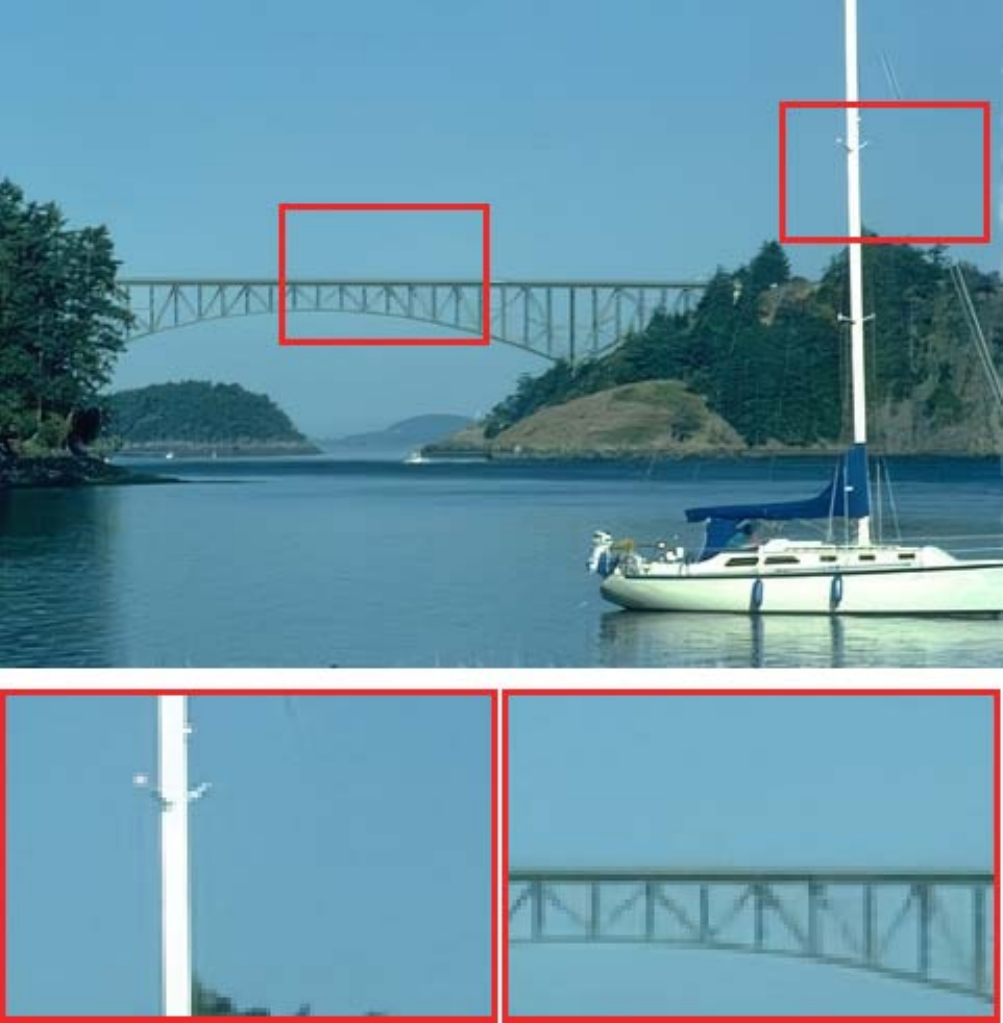}
			\caption{Restormer}
		\end{subfigure}
		\begin{subfigure}[t]{0.19\textwidth}
			\centering
			\includegraphics[width=\textwidth]{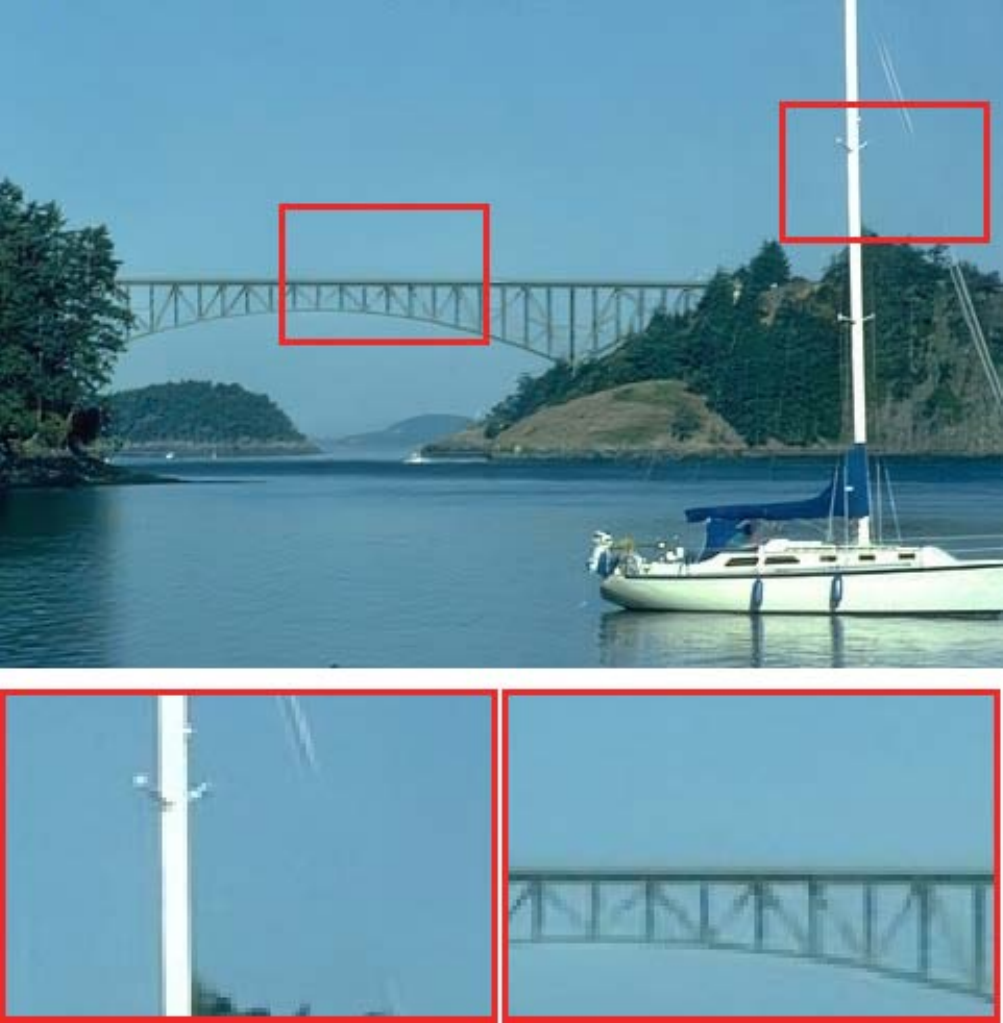}
			\caption{IDT}
		\end{subfigure}
		\begin{subfigure}[t]{0.19\textwidth}
			\centering
			\includegraphics[width=\textwidth]{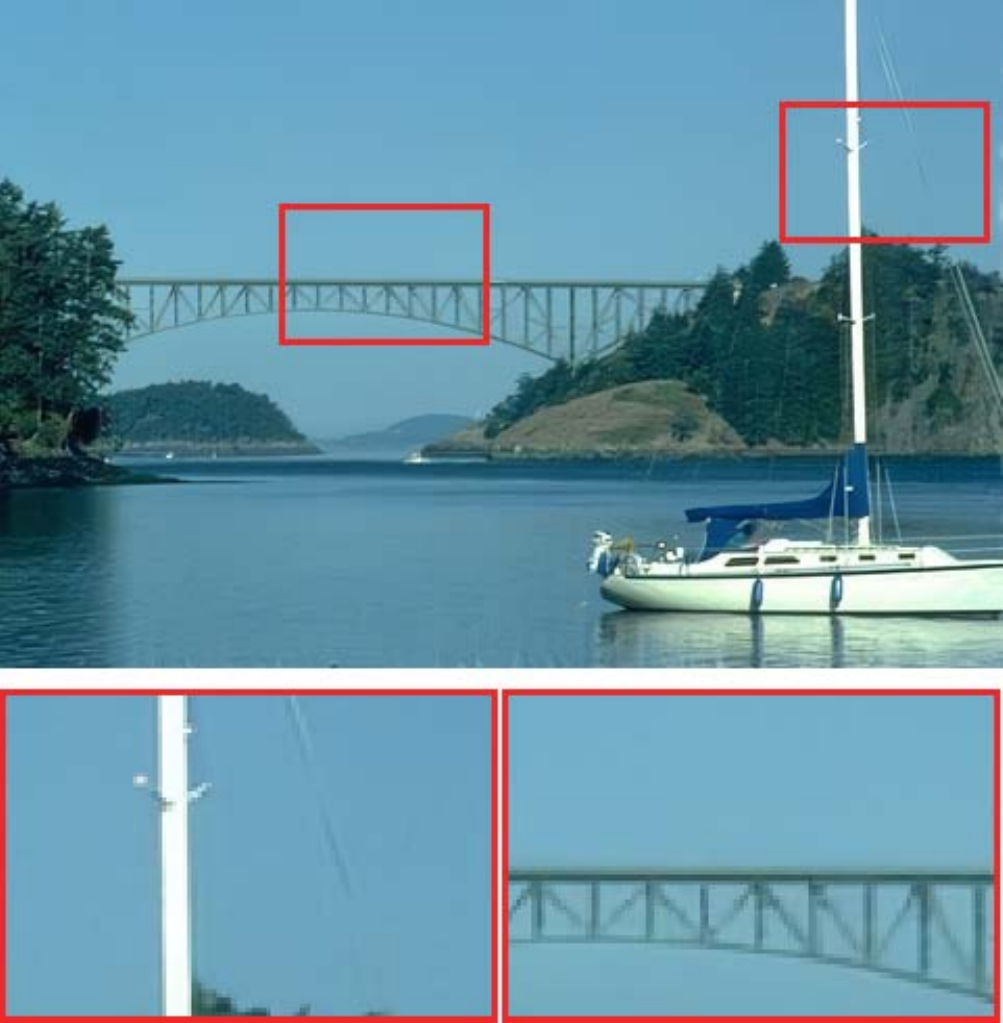}
			\caption{DRSformer}
		\end{subfigure}	
		\begin{subfigure}[t]{0.19\textwidth}
			\centering
			\includegraphics[width=\textwidth]{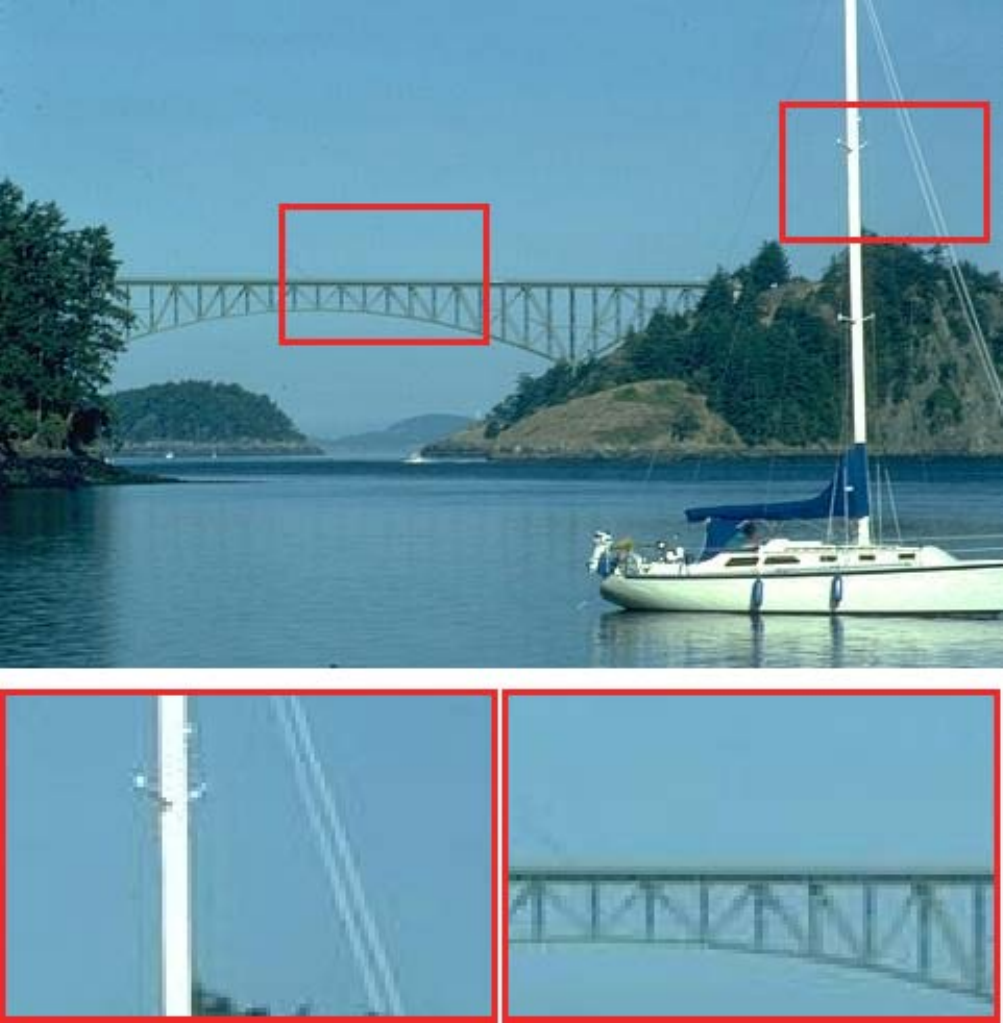}
			\caption{Ground Truth}
		\end{subfigure}	
		\caption{Visual quality comparison of independent training track on the DDN-Data dataset. Please zoom in the figures for better view of the rain removal and detail recovery.}
		\label{fig7}
	\end{figure*}
	
	\subsection{Evaluation on Independent Training Track}
	In this track, multiple pretrained models are obtained to evaluate the corresponding testing sets, including Rain200L \cite{yang2017deep}, Rain200H \cite{yang2017deep}, DID-Data \cite{zhang2018density}, DDN-Data \cite{fu2017removing}, and SPA-Data \cite{wang2019spatial}.
	Here, we provide 22 representative methods in Table \ref{table6}, \emph{i.e.}, DDN \cite{fu2017removing}, DID-MDN \cite{zhang2018density}, RESCAN \cite{li2018recurrent}, NLEDN \cite{li2018non}, JORDER-E \cite{yang2019joint}, ID-CGAN \cite{zhang2019image}, SIRR \cite{wei2019semi}, PReNet \cite{ren2019progressive}, SPANet \cite{wang2019spatial}, FBL \cite{yang2020towards}, MSPFN \cite{jiang2020multi}, RCDNet \cite{wang2020model}, Syn2Real \cite{yasarla2020syn2real}, SGCN \cite{fu2021successive}, MPRNet \cite{zamir2021multi}, DualGCN \cite{fu2021rain}, SPDNet \cite{yi2021structure}, Uformer \cite{wang2022uformer}, Restormer \cite{zamir2022restormer}, IDT \cite{xiao2022image}, HCN \cite{fu2023continual}, and DRSformer \cite{chen2023learning}.
	The quantitative results are quoted from previous works \cite{fu2023continual, chen2023learning}.
	Each method is evaluated under the same implementation using PSNR \cite{huynh2008scope} and SSIM \cite{wang2004image}.
	
	So far, DRSformer \cite{chen2023learning} achieves the state-of-the-art quantitative average results in the independent training track.
	We can note that the best performance of different methods gradually converges. The performance gaps between DRSformer \cite{chen2023learning}, Restormer \cite{zamir2022restormer} and IDT \cite{xiao2022image} are considerably close.
	We further present qualitative comparisons of deraining results in Figure~\ref{fig7}, where the ``spar'' shares similar directions and distributions to rain streaks.
	Notably, none of the methods can preserve the original details of the image content area while removing the rain streaks.
	Overall, there is still a certain gap between these results and ground truths.
	
	\begin{table*}[t]
		\centering
		\caption{Quantitative comparisons on the HQ-RAIN and RE-RAIN benchmark datasets. Top $1_{s t}$ and $2_{nd}$ results are marked in \textcolor{red}{red} and \textcolor{blue}{blue} respectively. ``\#FLOPs'', ``\#Params'' and ``\#Runtime'' represent FLOPs (in G), the number of trainable parameters (in M) and inference time (in second), respectively.}
		\resizebox{1.0\textwidth}{!}{
			\begin{tabular}{c|cccccccccc}
				\hlinew{1.0pt}
				\textbf{Methods} & \begin{tabular}[c]{@{}c@{}}LPNet\\ {\cite{fu2019lightweight}}\end{tabular} & \begin{tabular}[c]{@{}c@{}}PReNet\\ {\cite{ren2019progressive}}\end{tabular} & \begin{tabular}[c]{@{}c@{}}JORDER-E\\ {\cite{yang2019joint}}\end{tabular} & \begin{tabular}[c]{@{}c@{}}RCDNet\\ {\cite{wang2020model} }\end{tabular} & \begin{tabular}[c]{@{}c@{}}HINet\\ {\cite{chen2021hinet} }\end{tabular} & \begin{tabular}[c]{@{}c@{}}SPDNet\\ {\cite{yi2021structure}  }\end{tabular} & \begin{tabular}[c]{@{}c@{}}Uformer\\ {\cite{wang2022uformer}  }\end{tabular} & \begin{tabular}[c]{@{}c@{}}Restormer\\ {\cite{zamir2022restormer}  }\end{tabular} & \begin{tabular}[c]{@{}c@{}}IDT\\ {\cite{xiao2022image}  }\end{tabular} & \begin{tabular}[c]{@{}c@{}}DRSformer\\ {\cite{chen2023learning}   }\end{tabular} \\ \hline
				Benchmark        & \multicolumn{10}{c}{\textbf{HQ-RAIN}}                                                    \\
				PSNR $\uparrow$            & 31.39  & 36.73  & 36.46    & 39.53  & 42.30  & 34.02  & 37.57   & \textcolor{blue}{44.59}     & 38.38  & \textcolor{red}{44.67}     \\
				SSIM $\uparrow$            & 0.9203 & 0.9657 & 0.9599   & 0.9771 & 0.9841 & 0.9359 & 0.9673  & \textcolor{red}{0.9905}    & 0.9737 & \textcolor{blue}{0.9898}    \\
				LPIPS $\downarrow$           & 0.1024 & 0.0350 & 0.0435   & 0.0242 & 0.0147 & 0.0813 & 0.0314  & \textcolor{red}{0.0076}    & 0.0261 & \textcolor{blue}{0.0081}    \\ \hline
				Benchmark        & \multicolumn{10}{c}{\textbf{RE-RAIN}}                                                    \\
				NIQE $\downarrow$            & 5.550  & 5.619  & \textcolor{blue}{5.401}    & 5.658  & 5.784  & 5.405  & \textcolor{red}{5.370}   & 5.864     & 5.683  & 5.857     \\
				PIQE $\downarrow$            & 13.787 & \textcolor{red}{13.324} & 13.716   & 13.813 & \textcolor{blue}{13.418} & 14.249 & 13.473  & 13.623    & 13.746 & 13.678    \\
				BRISQUE $\downarrow$         & \textcolor{red}{24.863} & 28.180 & \textcolor{blue}{25.436}   & 28.006 & 29.003 & 26.134 & 26.491  & 29.725    & 28.330 & 29.747    \\ \hline
				\#FLOPs (G) $\downarrow$            & \textcolor{red}{3.5}  & 66.4  & 27.2    & \textcolor{blue}{21.2}  & 170.7  & 89.3  & 45.9   & 174.7     & 61.9  & 242.9     \\
				\#Params (M) $\downarrow$            & \textcolor{red}{0.03} & \textcolor{blue}{0.17} & 4.21   & 3.17 & 88.67 & 3.04 & 50.88  & 26.12    & 16.41 & 33.65   \\			
				\#Runtime (s) $\downarrow$            & \textcolor{red}{0.07} & \textcolor{blue}{0.09} & 0.25   & 0.26 & 0.38 & 0.24 & 0.21  & 0.28    & 0.28 & 0.30   \\					
				\hlinew{1.0pt}
			\end{tabular}
		}
		\label{table7}		
	\end{table*}
	
	\begin{figure*}[!t]
		\centering 	
		\begin{subfigure}[t]{0.162\textwidth}
			\centering
			\includegraphics[width=\textwidth]{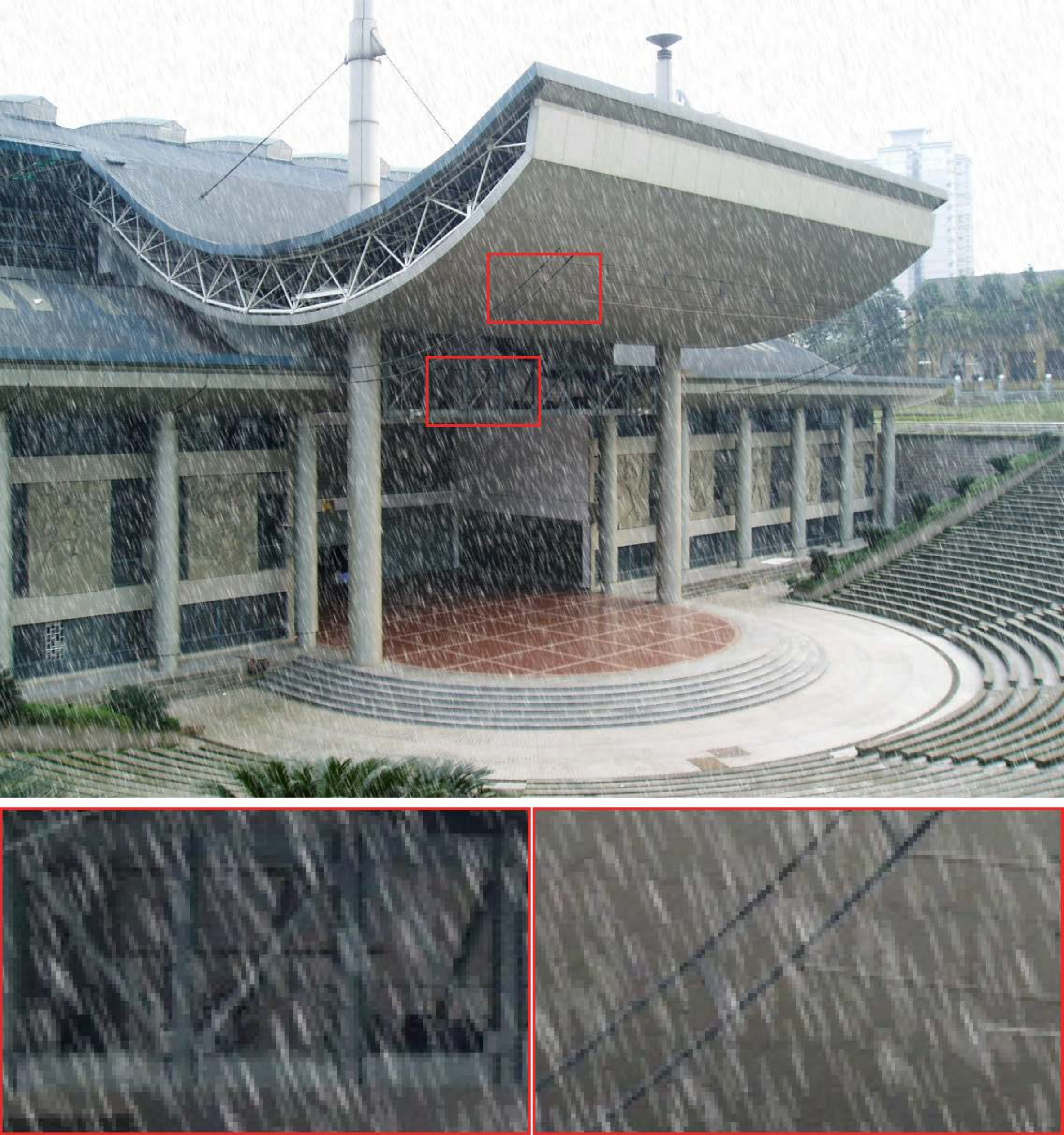}
			\caption{Rainy Input}
		\end{subfigure}
		\begin{subfigure}[t]{0.162\textwidth}
			\centering
			\includegraphics[width=\textwidth]{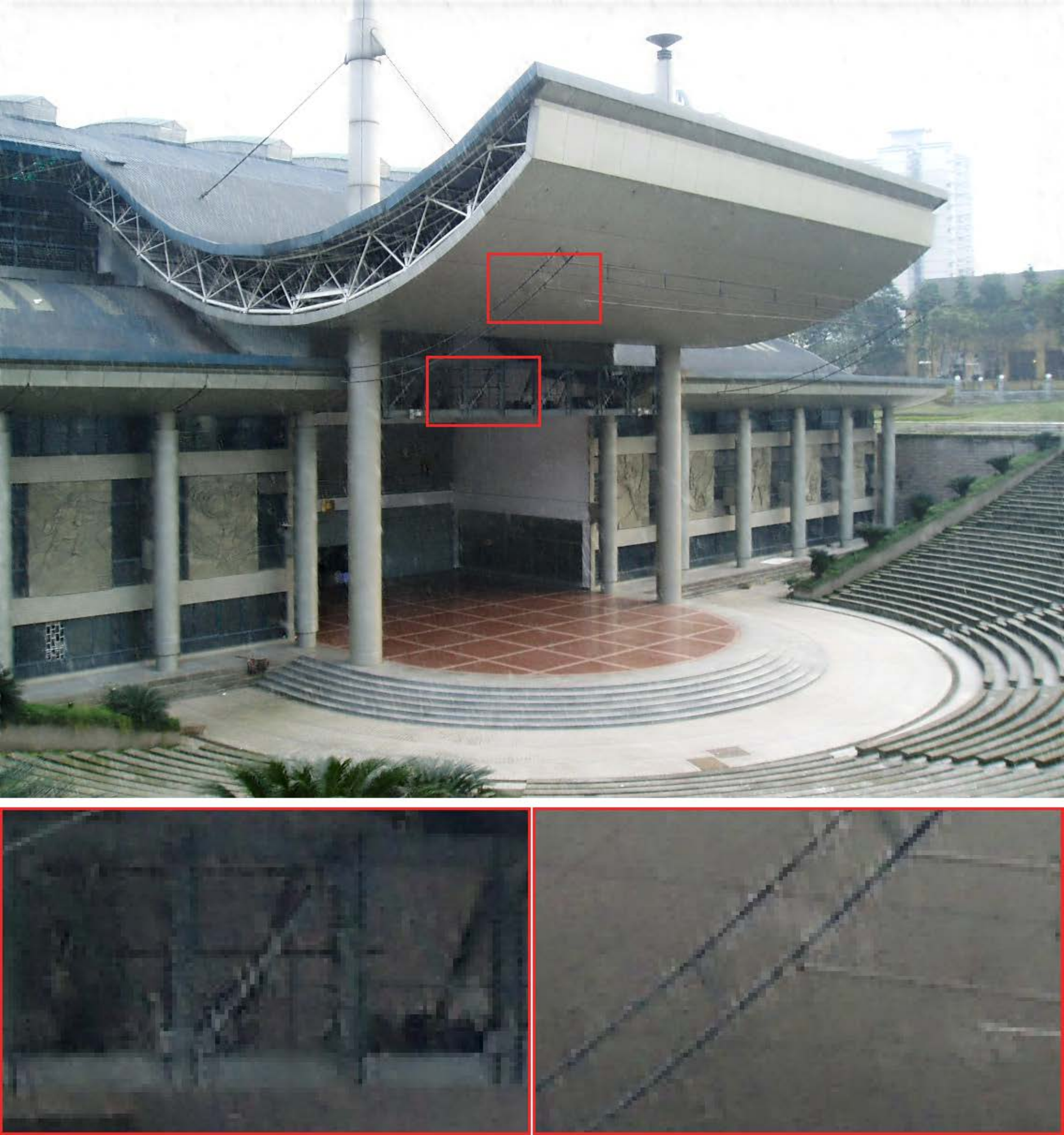}
			\caption{LPNet}
		\end{subfigure}
		\begin{subfigure}[t]{0.162\textwidth}
			\centering
			\includegraphics[width=\textwidth]{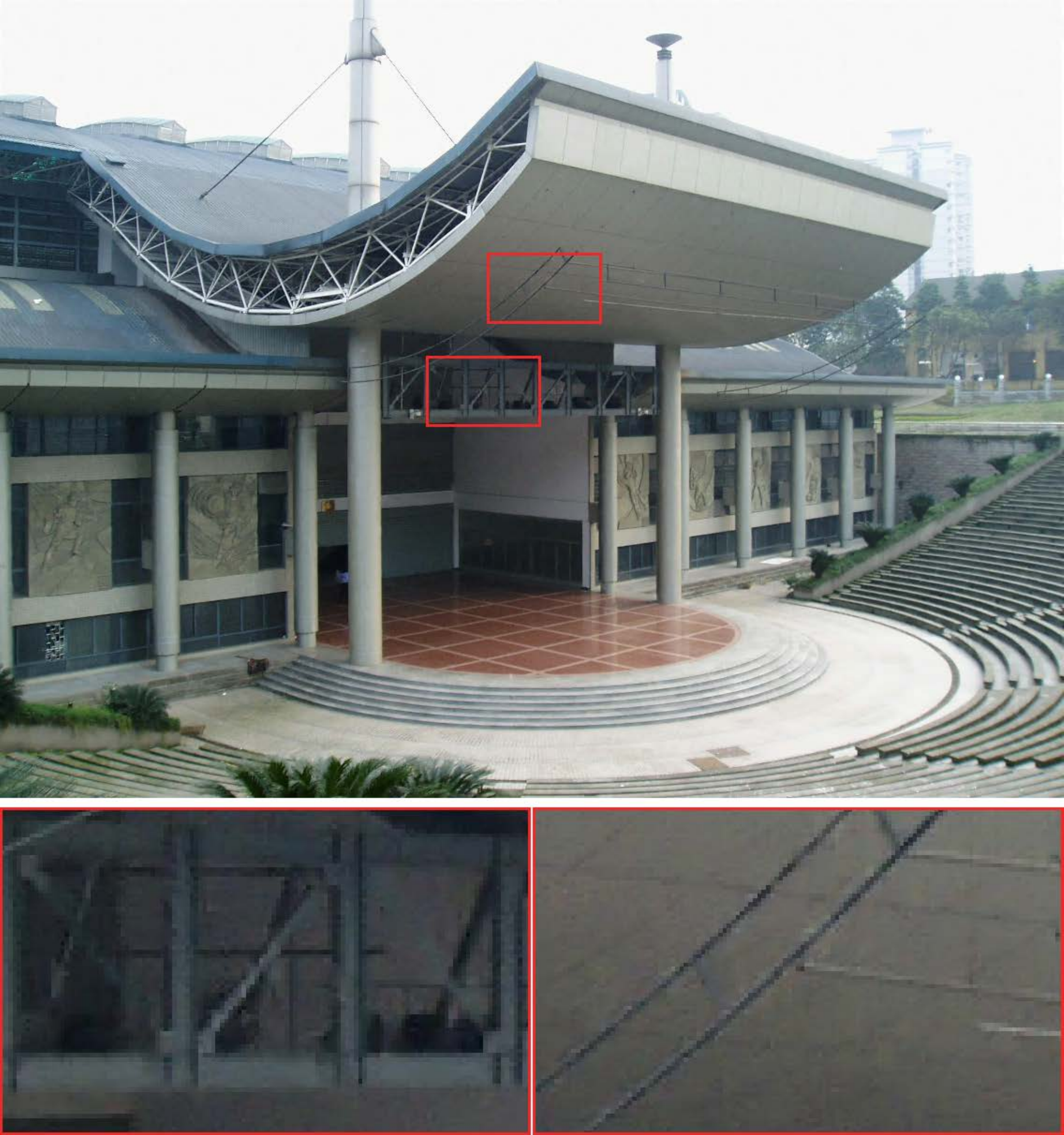}
			\caption{PReNet}
		\end{subfigure}	
		\begin{subfigure}[t]{0.162\textwidth}
			\centering
			\includegraphics[width=\textwidth]{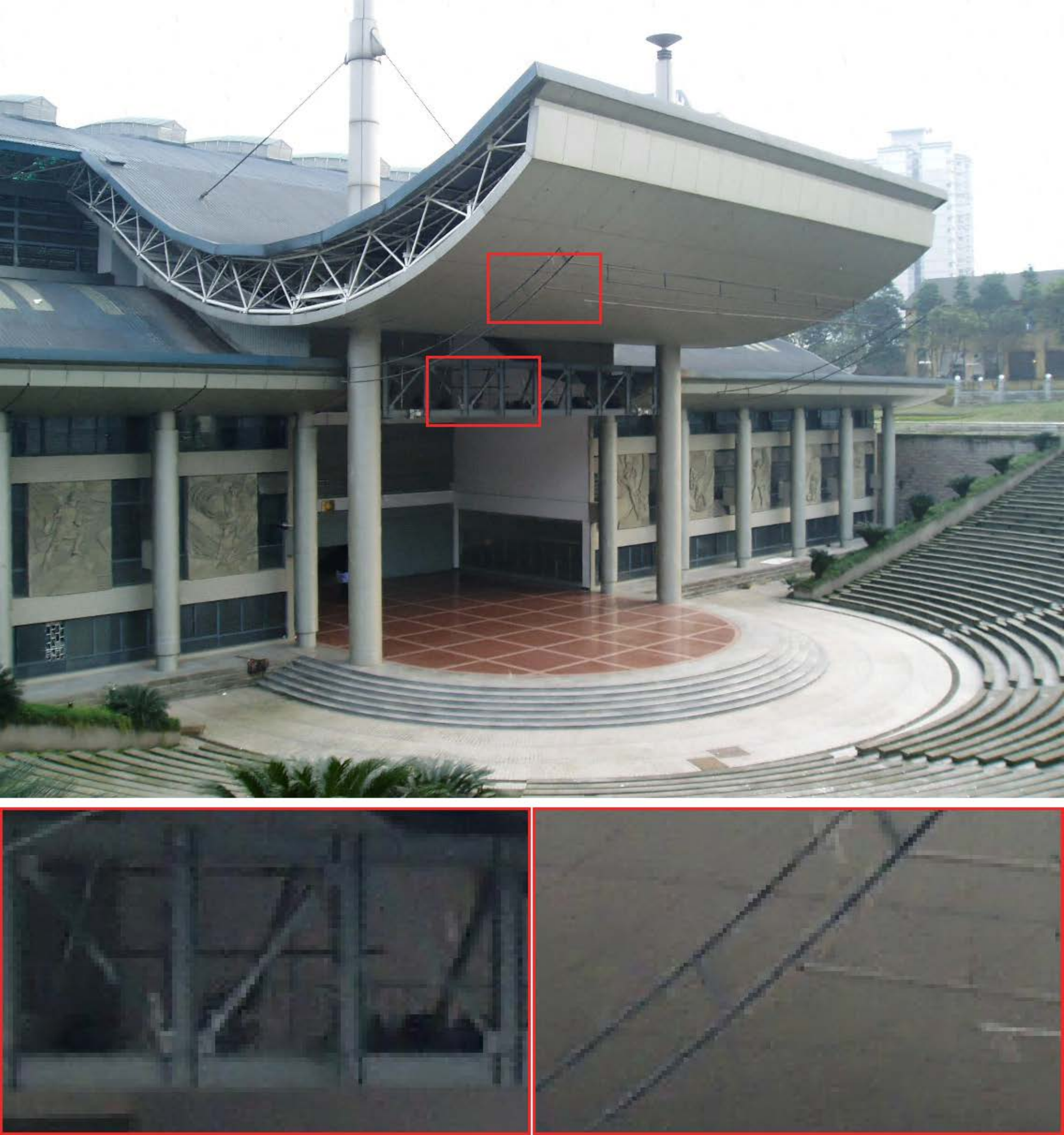}
			\caption{JORDER-E}
		\end{subfigure}	
		\begin{subfigure}[t]{0.162\textwidth}
			\centering
			\includegraphics[width=\textwidth]{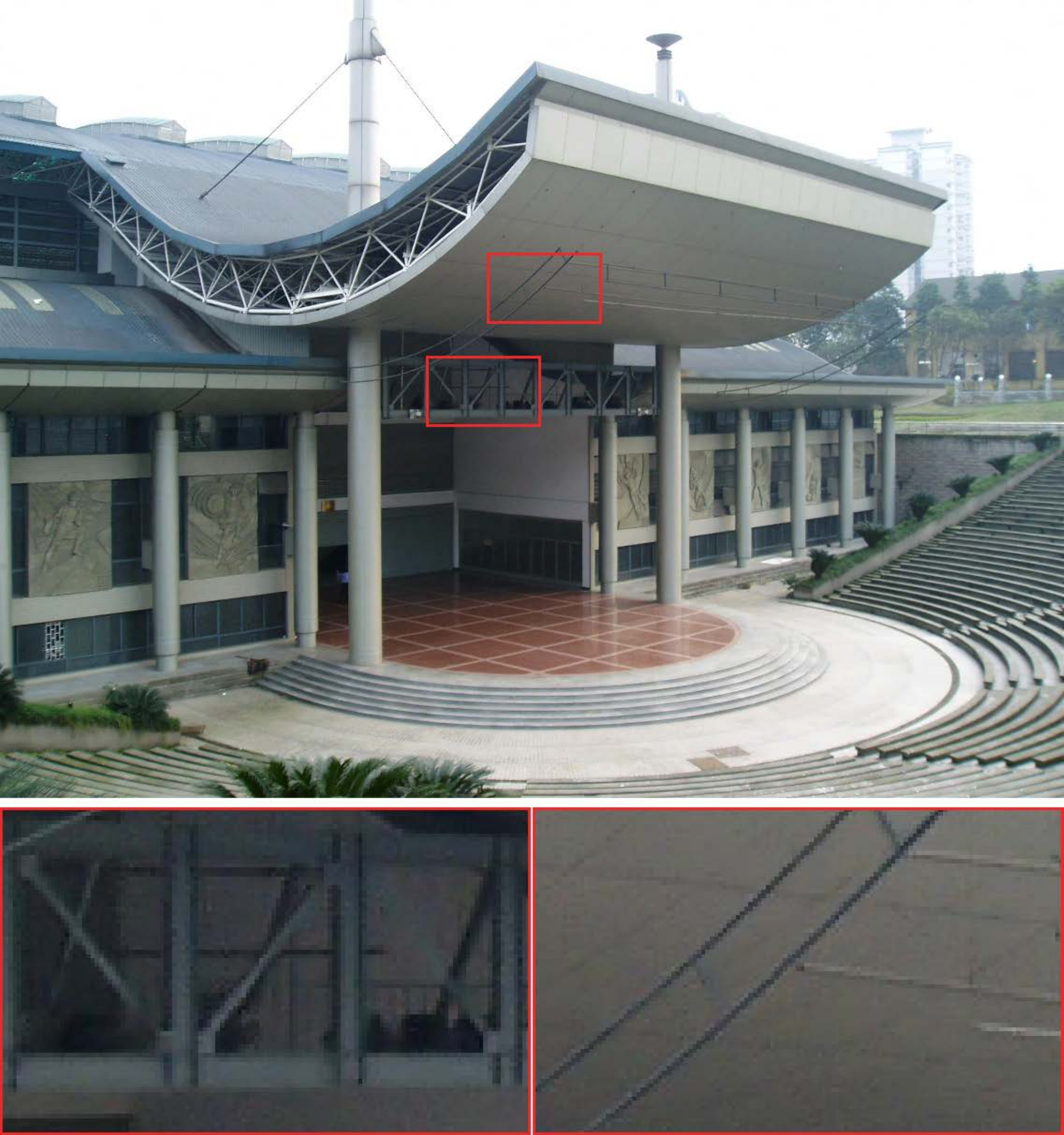}
			\caption{RCDNet}
		\end{subfigure}	
		\begin{subfigure}[t]{0.162\textwidth}
			\centering
			\includegraphics[width=\textwidth]{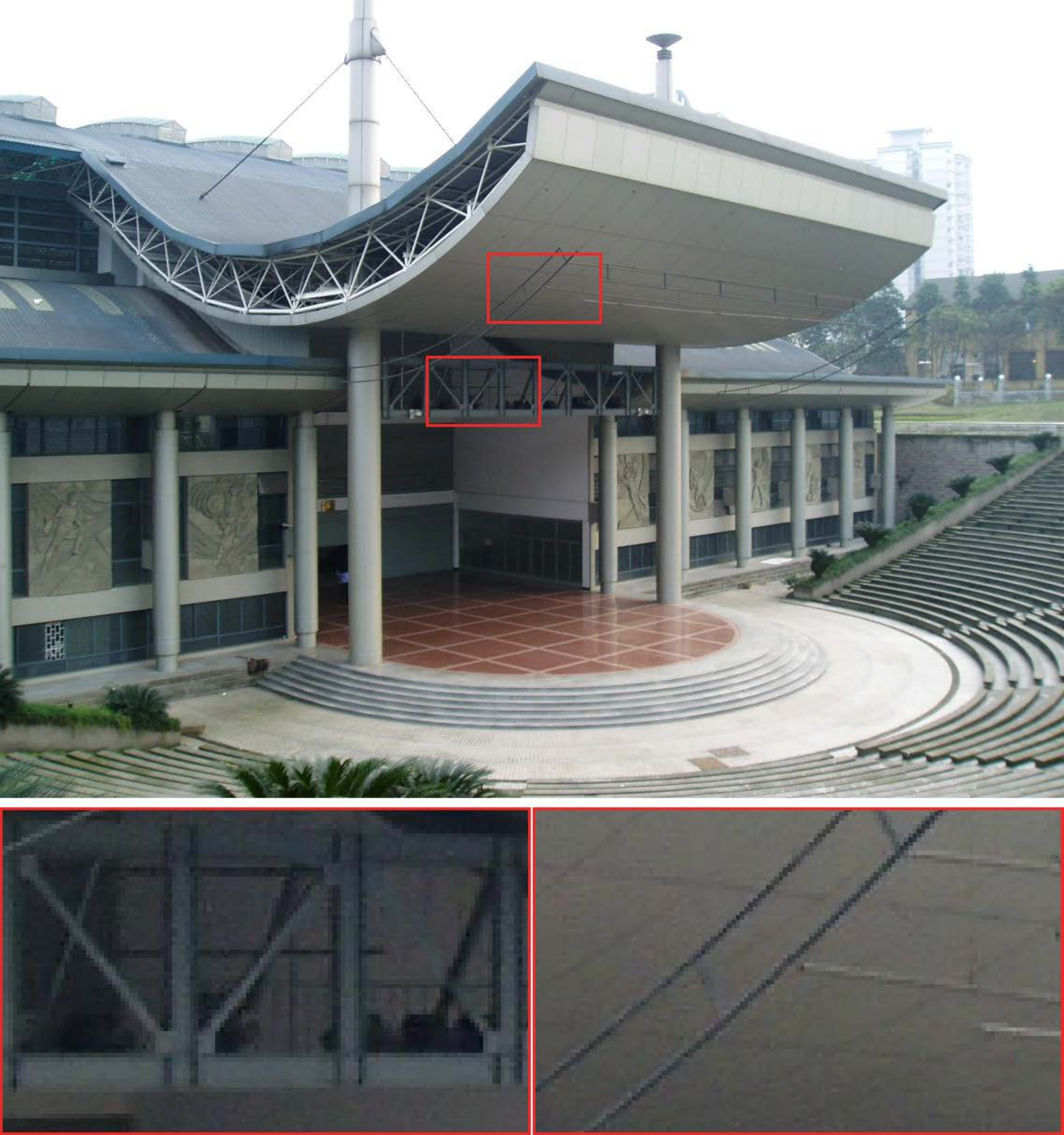}
			\caption{HINet}
		\end{subfigure}	
		\\
		\begin{subfigure}[t]{0.162\textwidth}
			\centering
			\includegraphics[width=\textwidth]{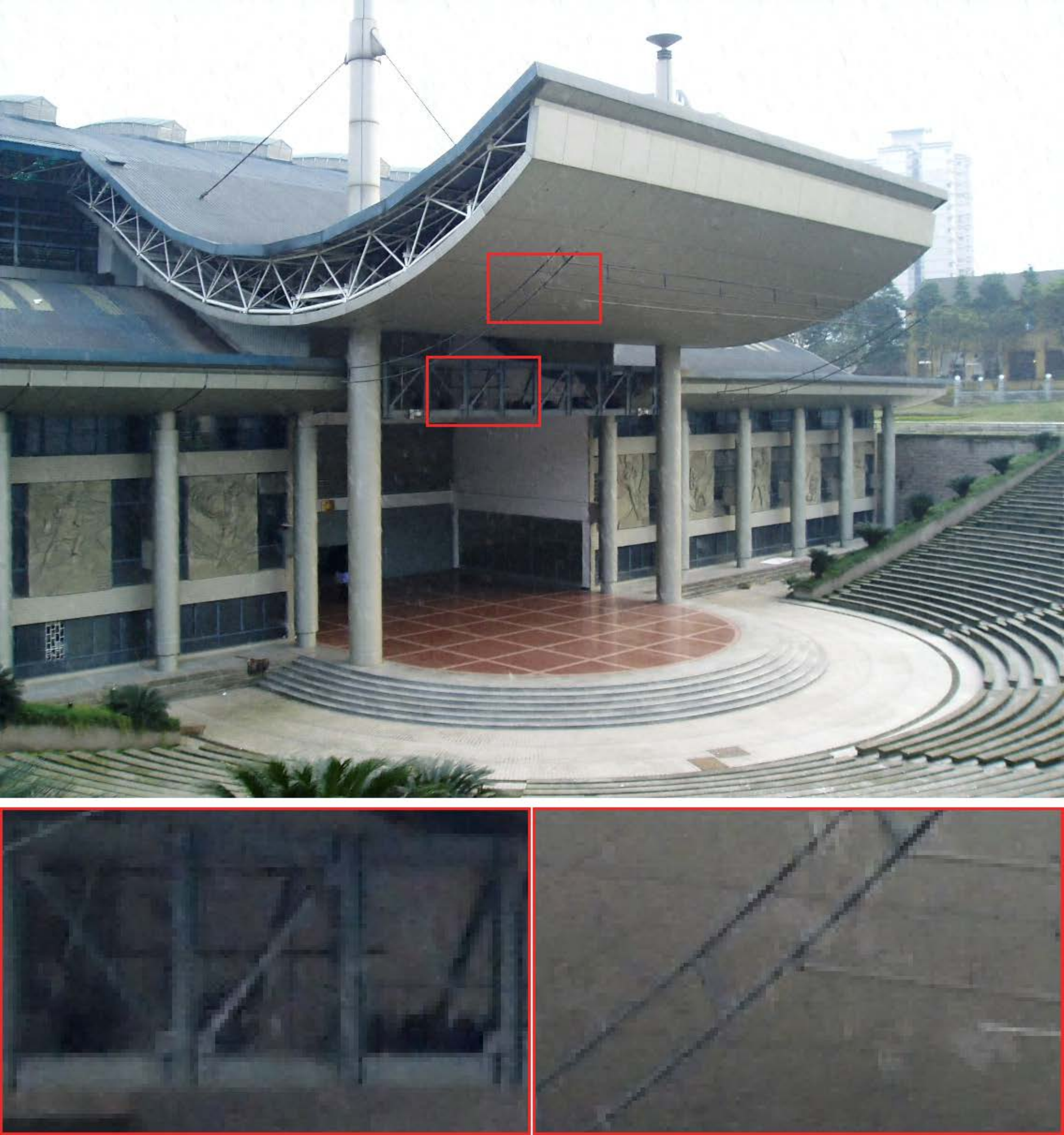}
			\caption{SPDNet}
		\end{subfigure}
		\begin{subfigure}[t]{0.162\textwidth}
			\centering
			\includegraphics[width=\textwidth]{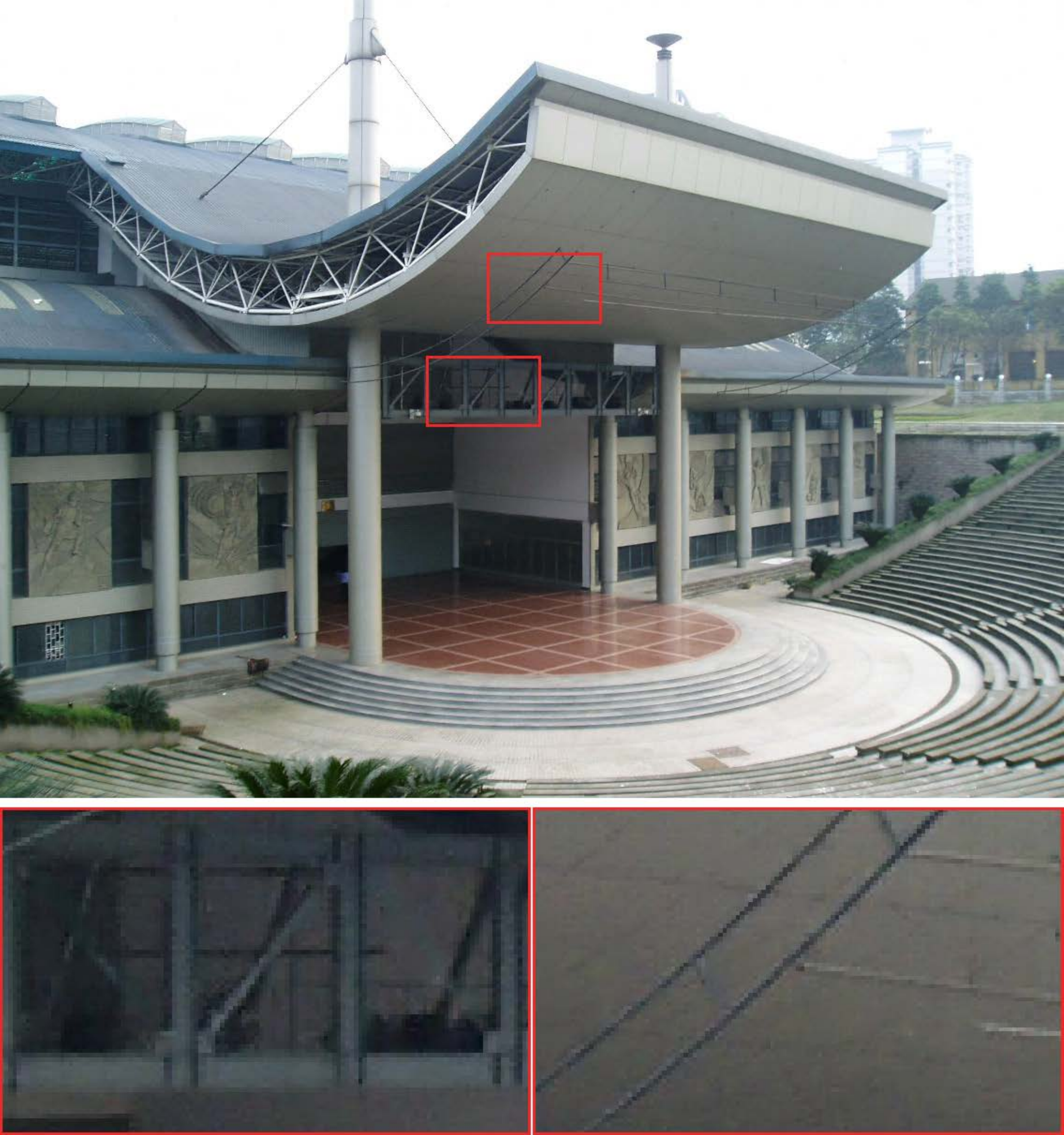}
			\caption{Uformer}
		\end{subfigure}
		\begin{subfigure}[t]{0.162\textwidth}
			\centering
			\includegraphics[width=\textwidth]{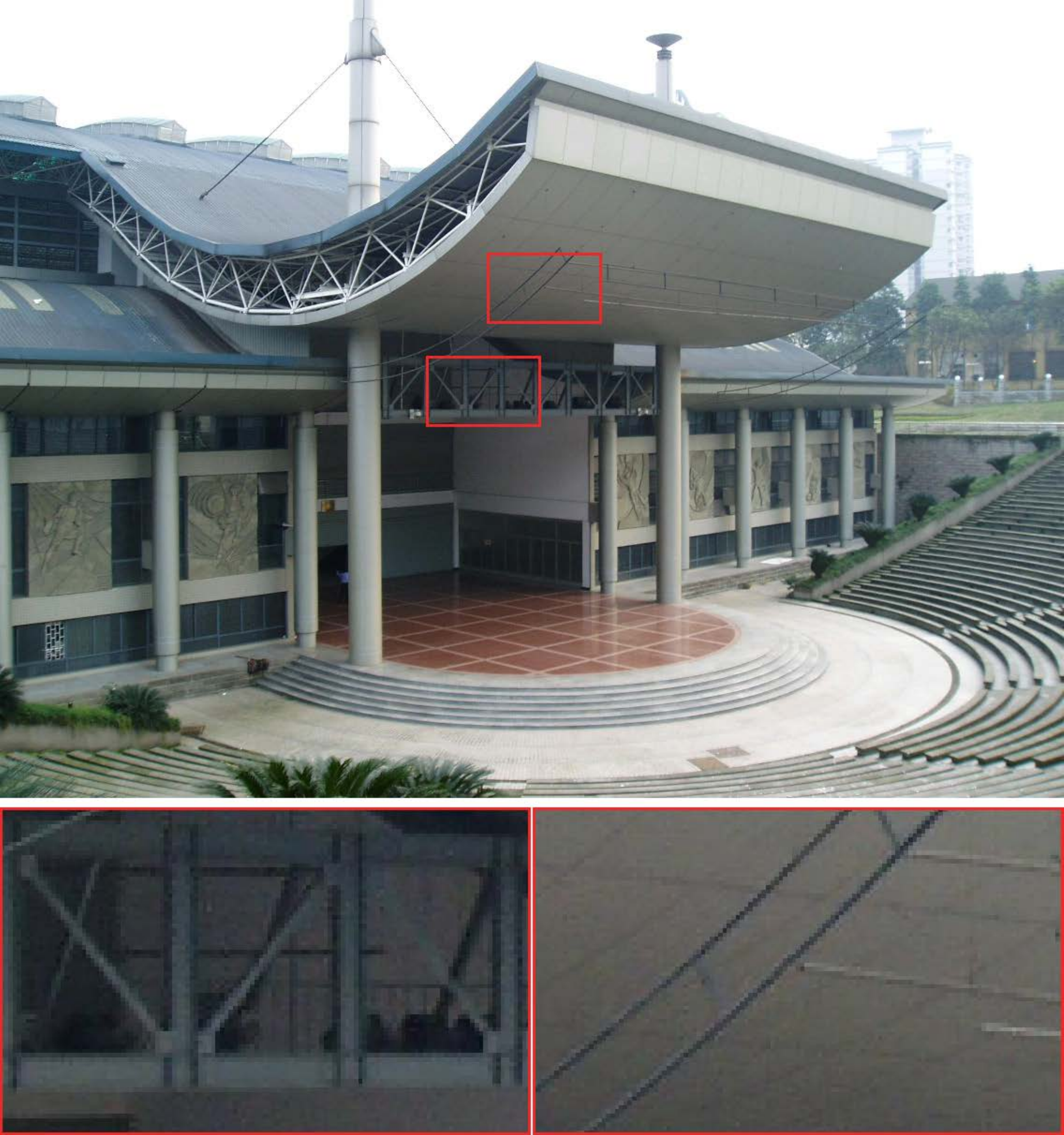}
			\caption{Restormer}
		\end{subfigure}
		\begin{subfigure}[t]{0.162\textwidth}
			\centering
			\includegraphics[width=\textwidth]{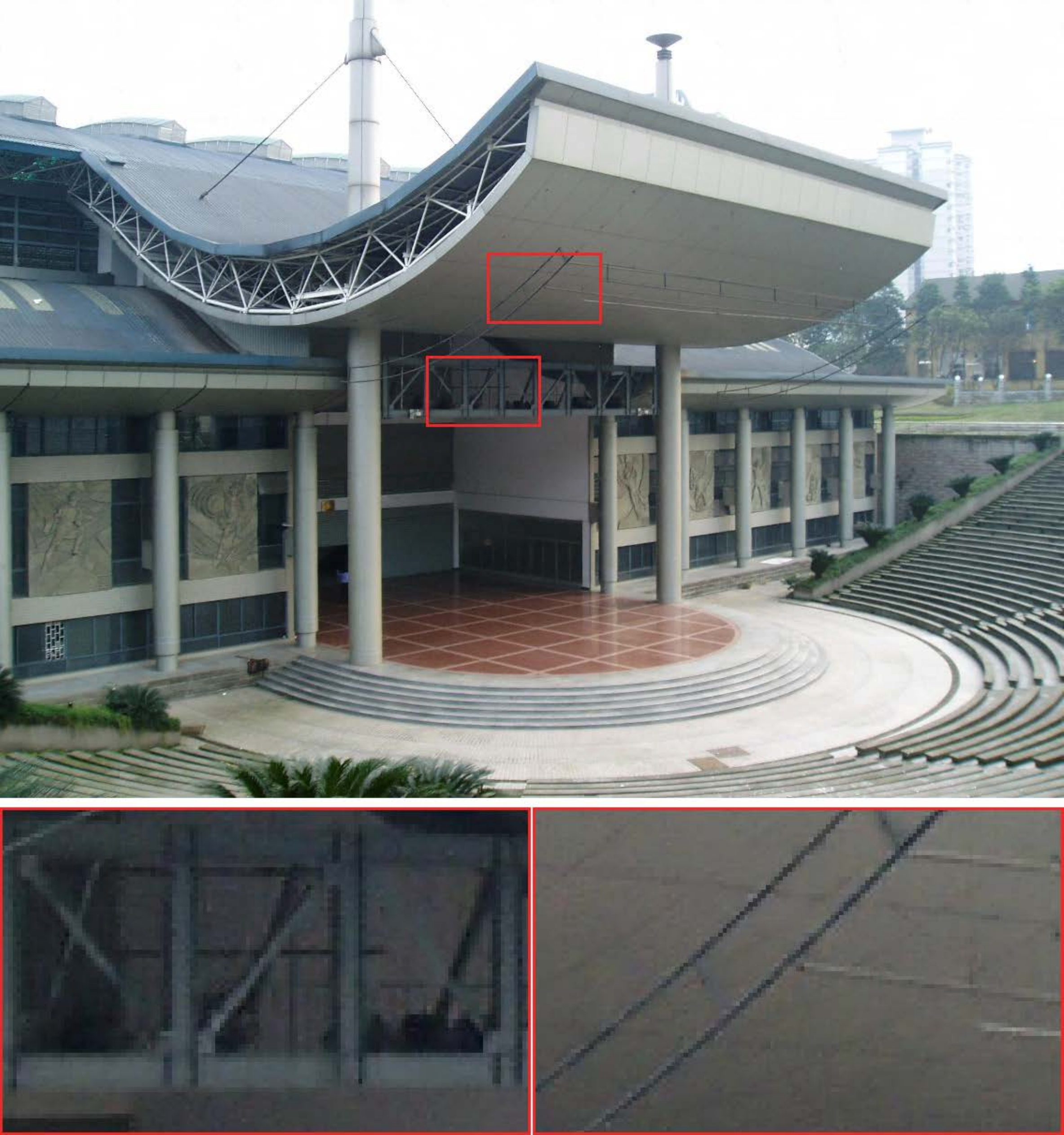}
			\caption{IDT}
		\end{subfigure}	
		\begin{subfigure}[t]{0.162\textwidth}
			\centering
			\includegraphics[width=\textwidth]{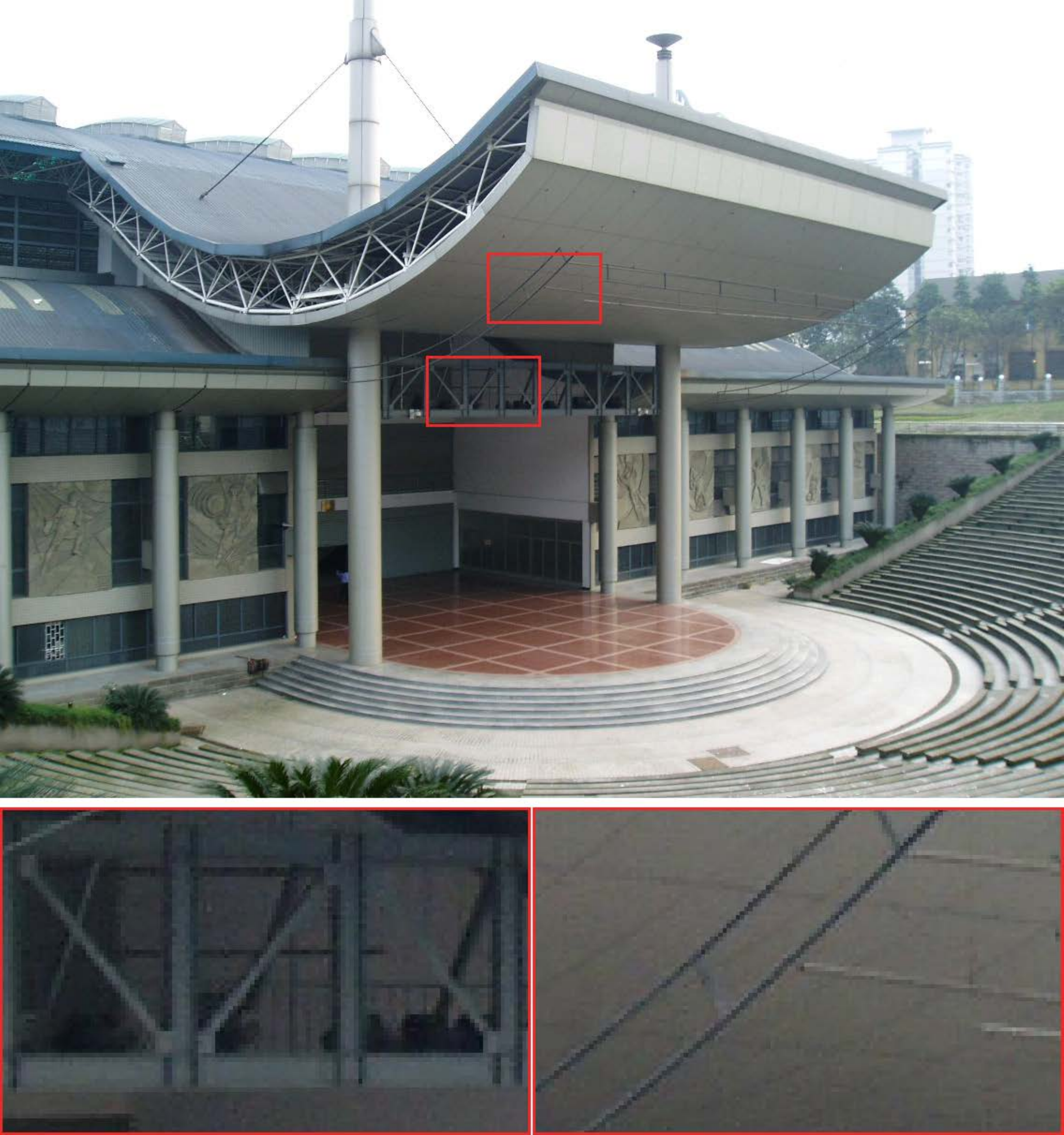}
			\caption{DRSformer}
		\end{subfigure}	
		\begin{subfigure}[t]{0.162\textwidth}
			\centering
			\includegraphics[width=\textwidth]{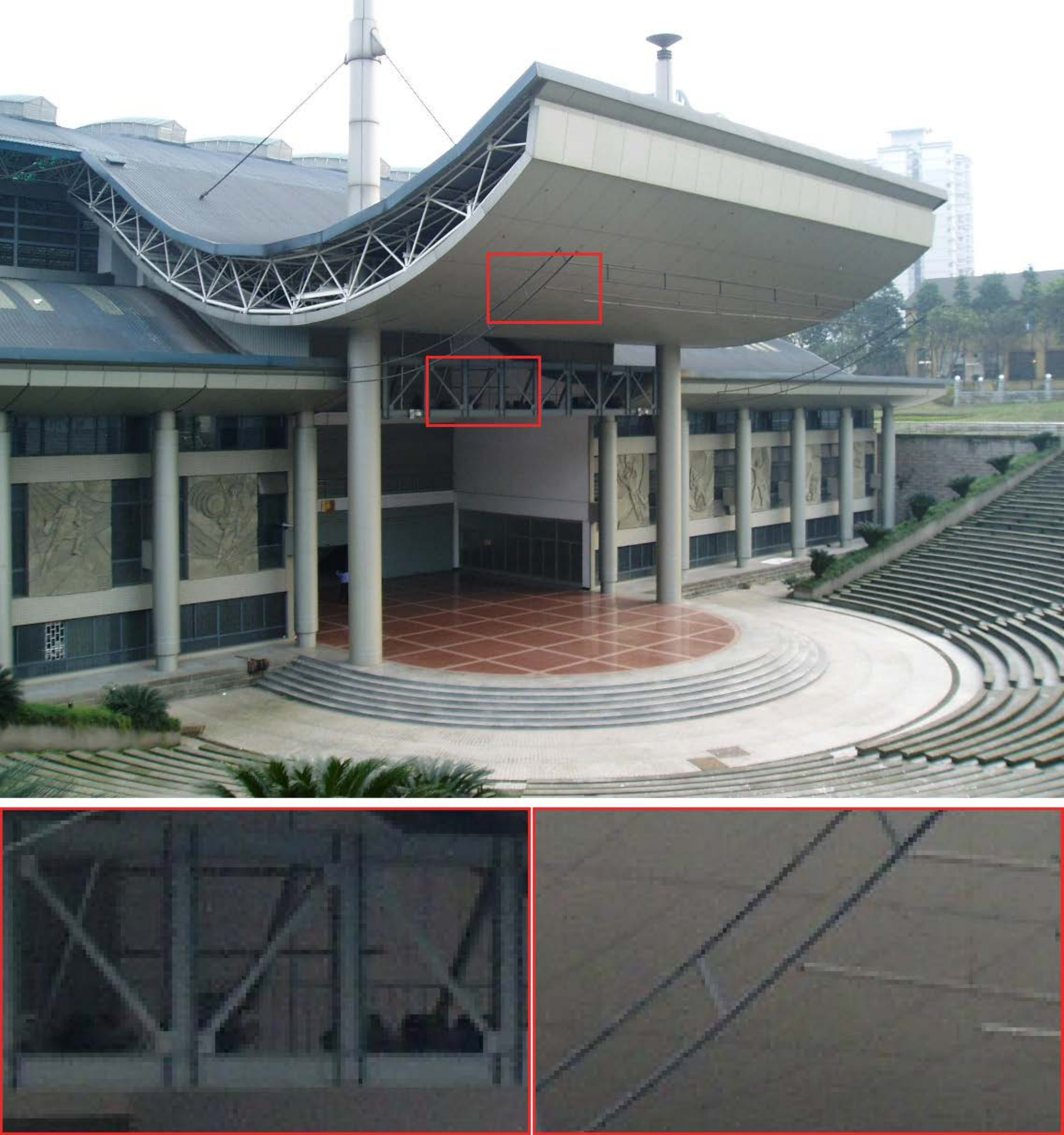}
			\caption{Ground Truth}
		\end{subfigure}	
		\caption{Visual quality comparison of new benchmark track on the HQ-RAIN dataset. Please zoom in the figures for better view of the rain removal and detail recovery.}
		\label{fig8}
	\end{figure*}
	
	\subsection{Evaluation on the HQ-RAIN Dataset}
	We further conduct experiments on the proposed new benchmark to evaluate the performance of existing methods. Specifically, we select 10 representative methods including 6 CNN-based approaches (\emph{i.e.}, LPNet \cite{fu2019lightweight}, PReNet \cite{ren2019progressive}, JORDER-E \cite{yang2019joint}, RCDNet \cite{wang2020model}, HINet \cite{chen2021hinet} and SPDNet \cite{yi2021structure}) and 4 Transformer-based methods (\emph{i.e.}, Uformer \cite{wang2022uformer}, Restormer \cite{zamir2022restormer}, IDT \cite{xiao2022image} and DRSformer \cite{chen2023learning}) for evaluation.
	For fair comparisons, we use the official released codes of these approaches.
	All methods are retrained on the HQ-RAIN benchmark. We choose their best pre-trained models, which achieve the highest PSNR value on the HQ-RAIN testing set. Meanwhile, these pre-trained models also are used to validate the RE-RAIN benchmark.
	All experiments are implemented on the servers equipped with NVIDIA GeForce RTX 3090 GPUs.
	For the HQ-RAIN benchmark with ground truth images, we use PSNR \cite{huynh2008scope}, SSIM \cite{wang2004image}, and LPIPS \cite{zhang2018unreasonable} to evaluate the quality of each restored images. For the RE-RAIN benchmark without ground truth images, we use the non-reference metrics including NIQE \cite{mittal2012no}, PIQE \cite{venkatanath2015blind}, and BRISQUE \cite{mittal2012making}.
	Higher PSNR and SSIM scores indicate higher restoration quality, while lower LPIPS, NIQE, PIQE and BRISQUE values indicate better perceptual quality.
	The above-mentioned metrics are calculated based on the Y channel in YCbCr space of derained images for fair comparisons.

\begin{figure*}[!t]
		\centering 	
		\begin{subfigure}[t]{0.196\textwidth}
			\centering
			\includegraphics[width=\textwidth]{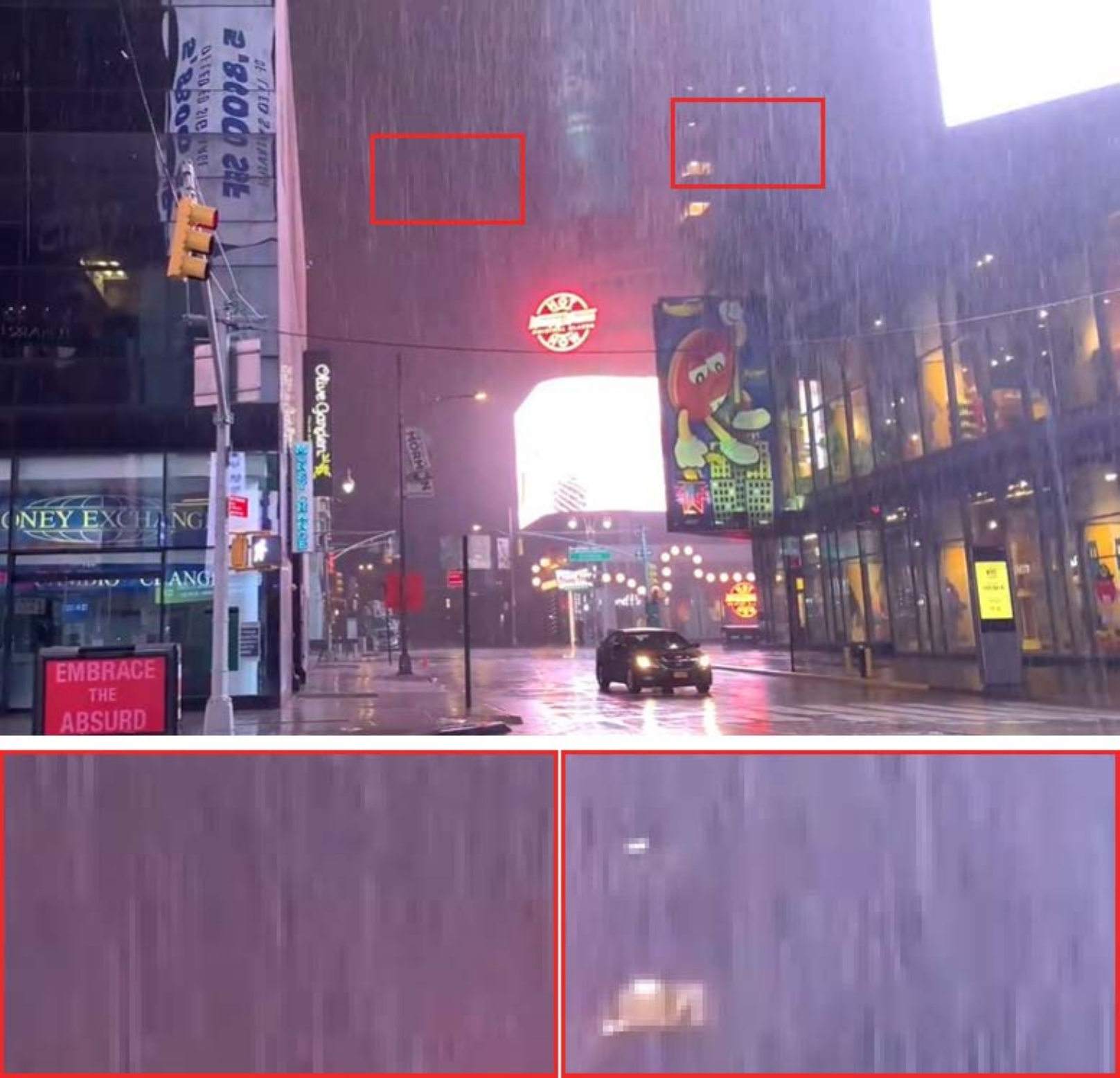}
			\caption{Rainy Input}
		\end{subfigure}
		\begin{subfigure}[t]{0.196\textwidth}
			\centering
			\includegraphics[width=\textwidth]{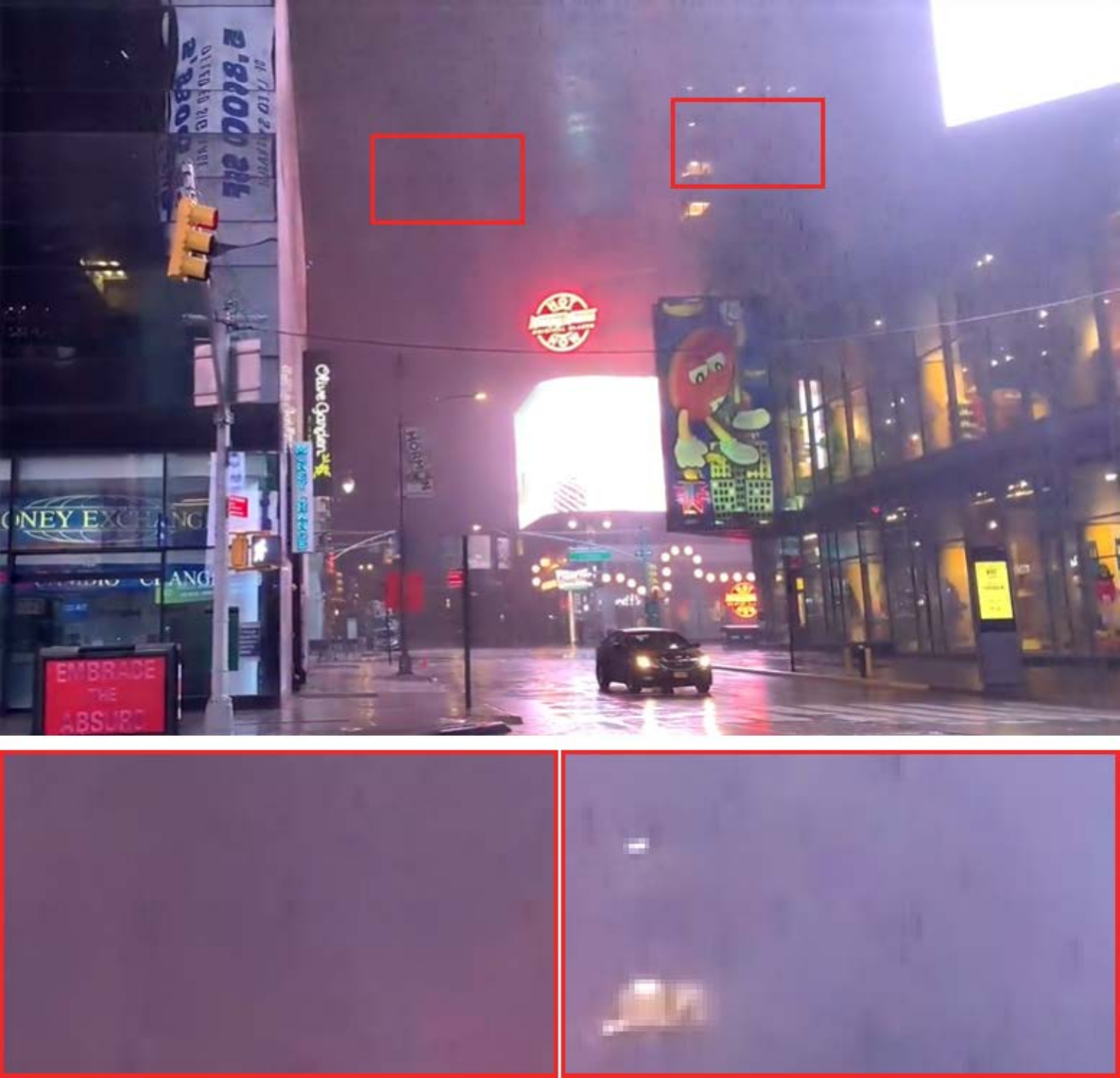}
			\caption{LPNet}
		\end{subfigure}
		\begin{subfigure}[t]{0.196\textwidth}
			\centering
			\includegraphics[width=\textwidth]{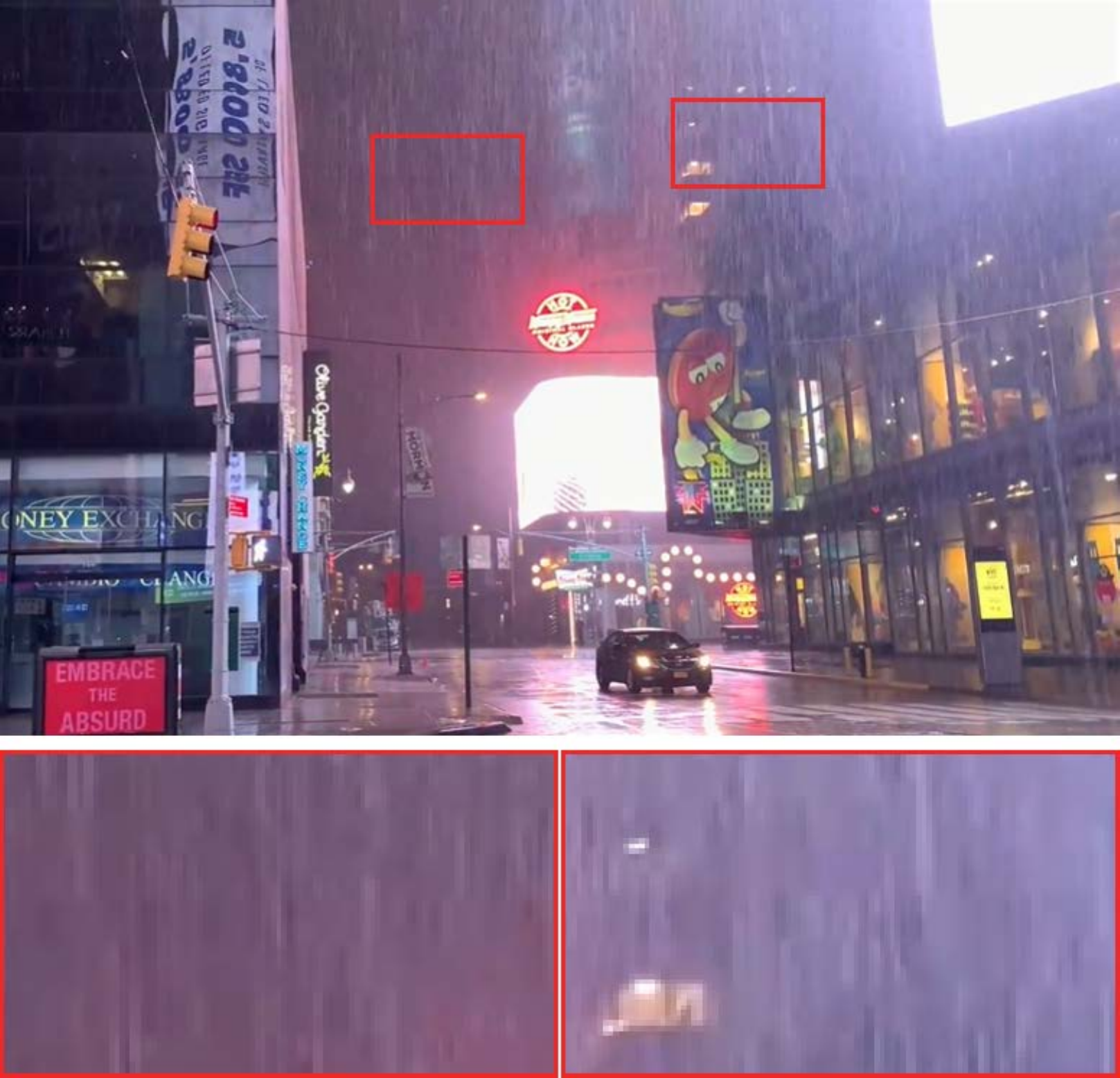}
			\caption{PReNet}
		\end{subfigure}	
		\begin{subfigure}[t]{0.196\textwidth}
			\centering
			\includegraphics[width=\textwidth]{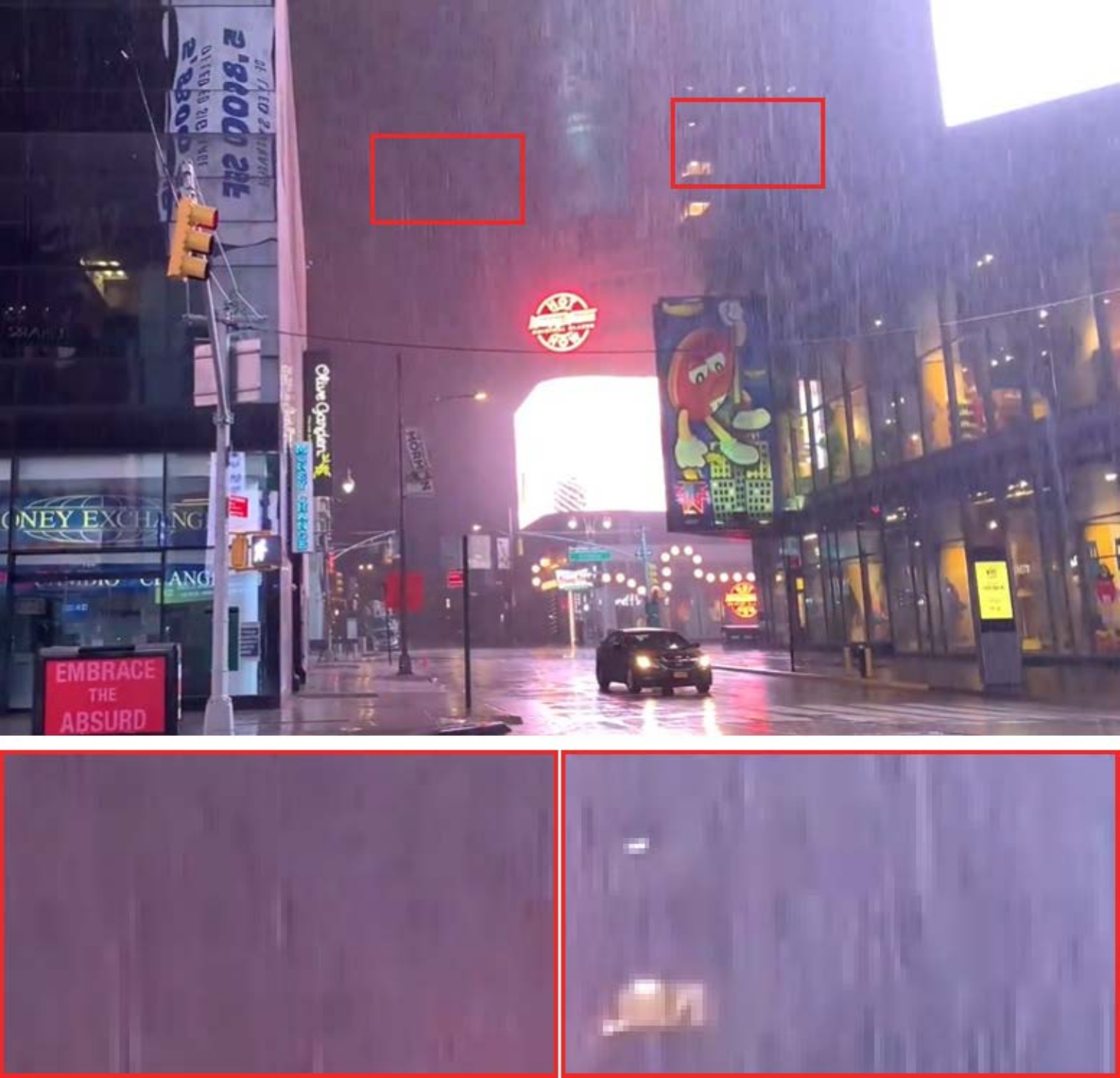}
			\caption{JORDER-E}
		\end{subfigure}	
		\begin{subfigure}[t]{0.196\textwidth}
			\centering
			\includegraphics[width=\textwidth]{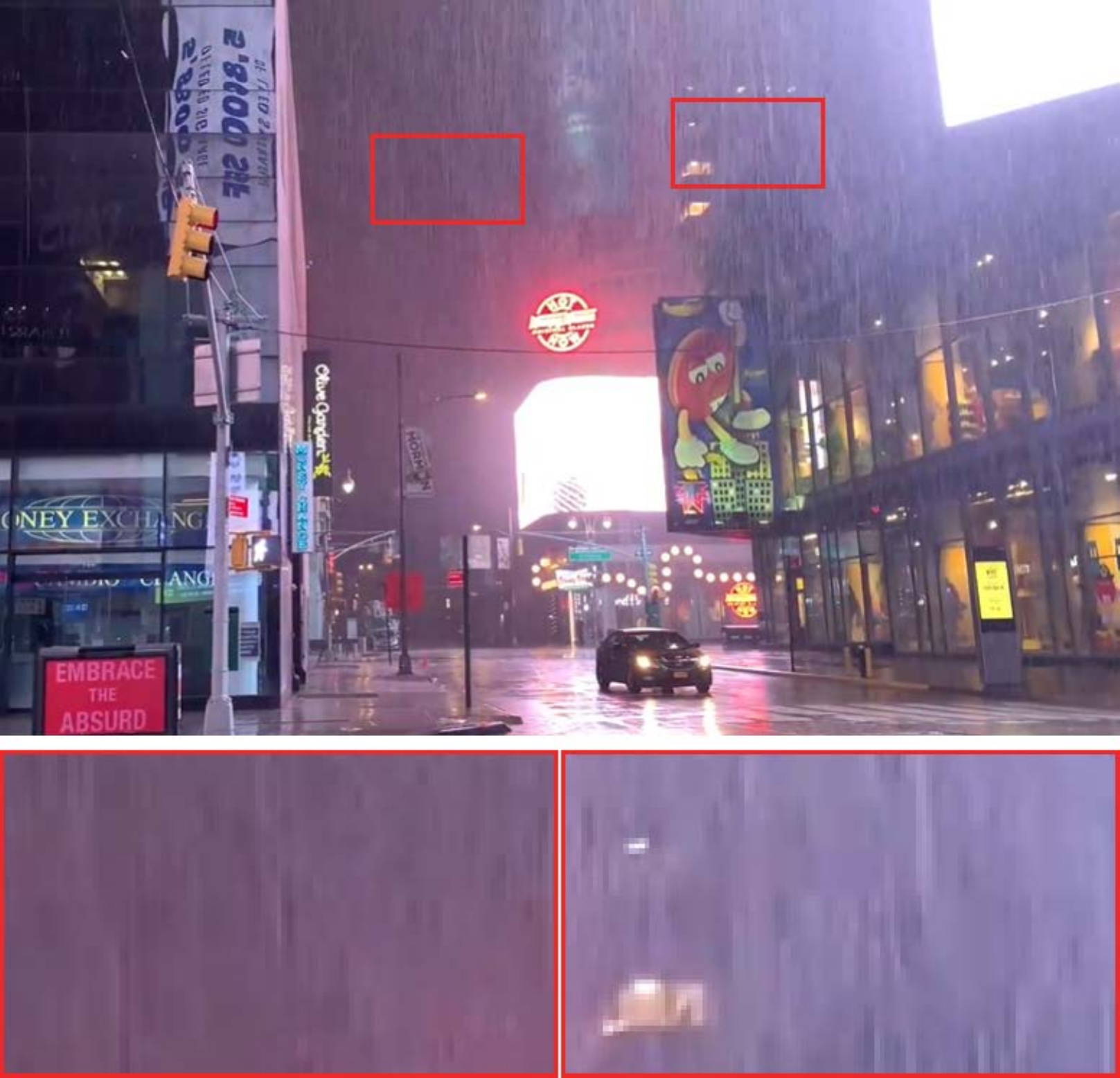}
			\caption{RCDNet}
		\end{subfigure}	
		\\
		\begin{subfigure}[t]{0.196\textwidth}
			\centering
			\includegraphics[width=\textwidth]{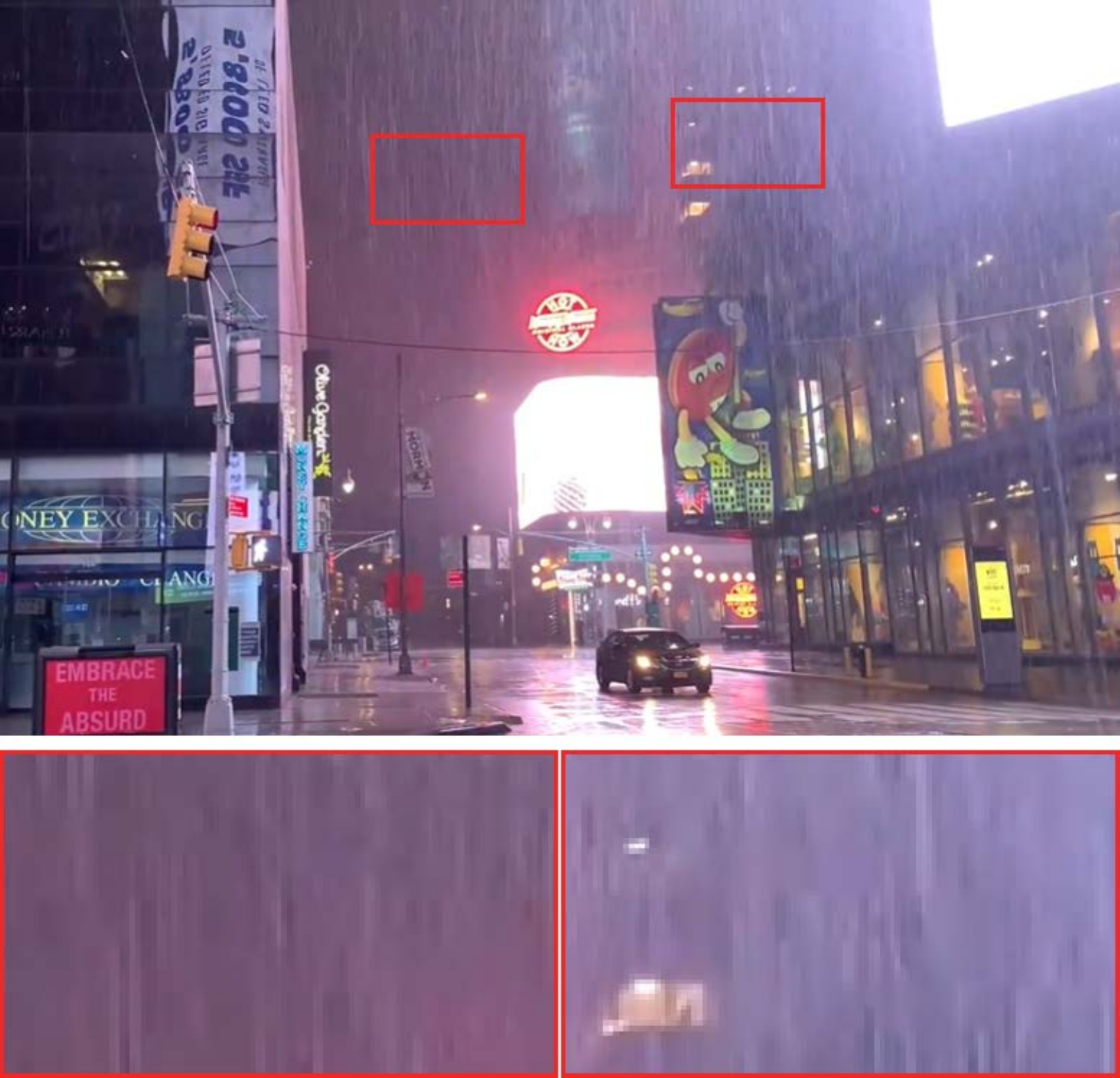}
			\caption{HINet}
		\end{subfigure}	
		\begin{subfigure}[t]{0.196\textwidth}
			\centering
			\includegraphics[width=\textwidth]{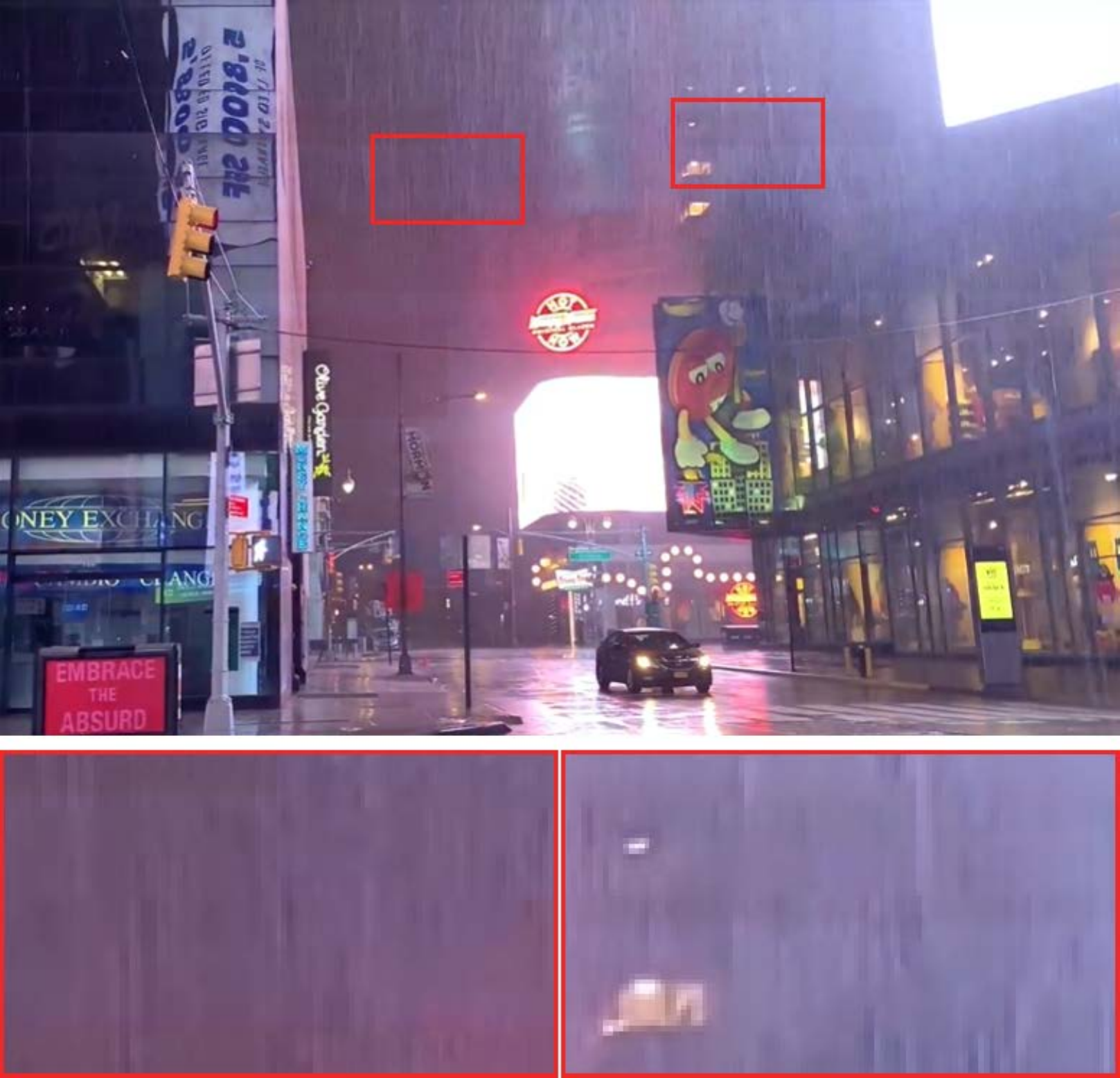}
			\caption{SPDNet}
		\end{subfigure}
		\begin{subfigure}[t]{0.196\textwidth}
			\centering
			\includegraphics[width=\textwidth]{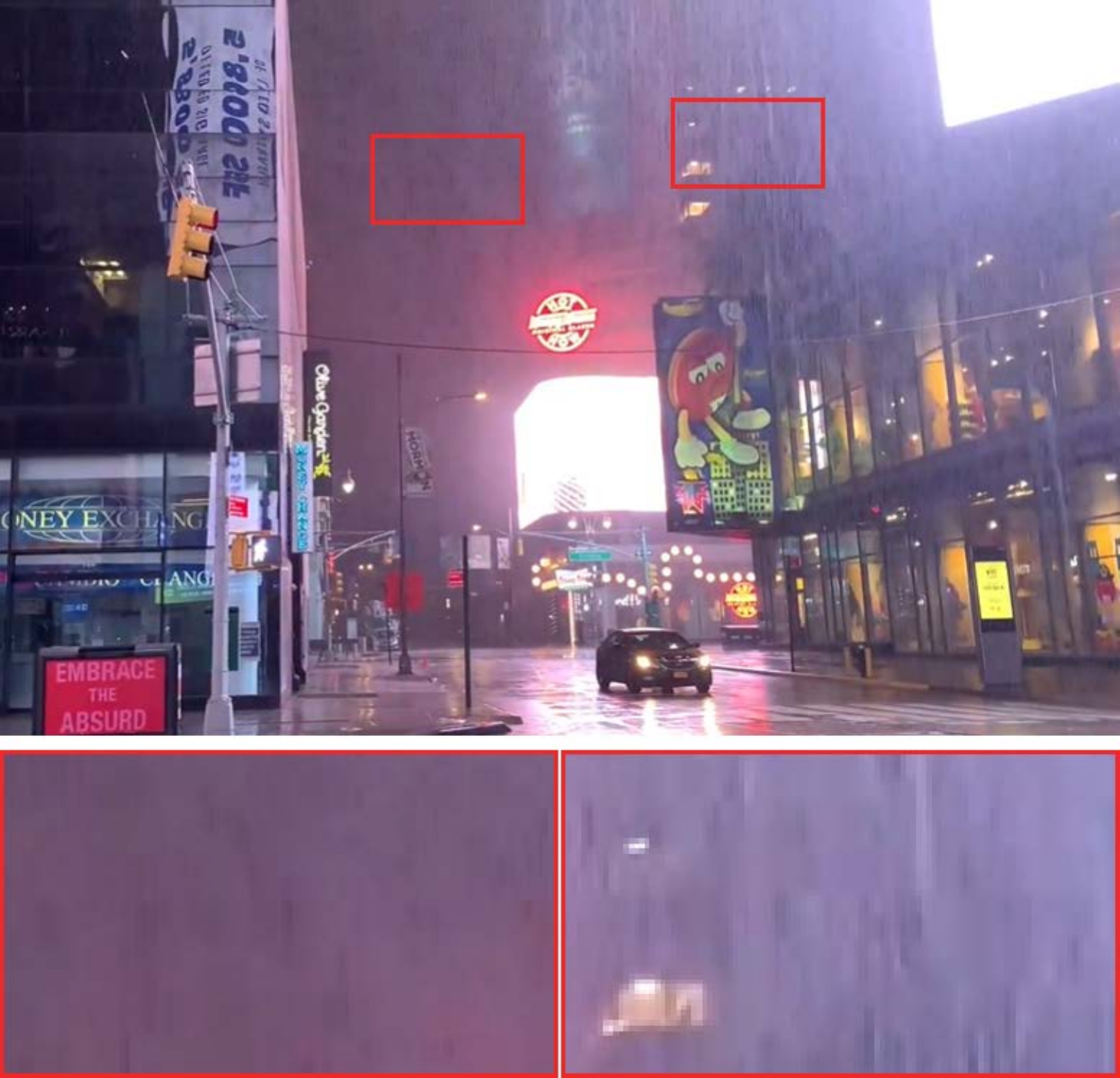}
			\caption{Uformer}
		\end{subfigure}
		\begin{subfigure}[t]{0.196\textwidth}
			\centering
			\includegraphics[width=\textwidth]{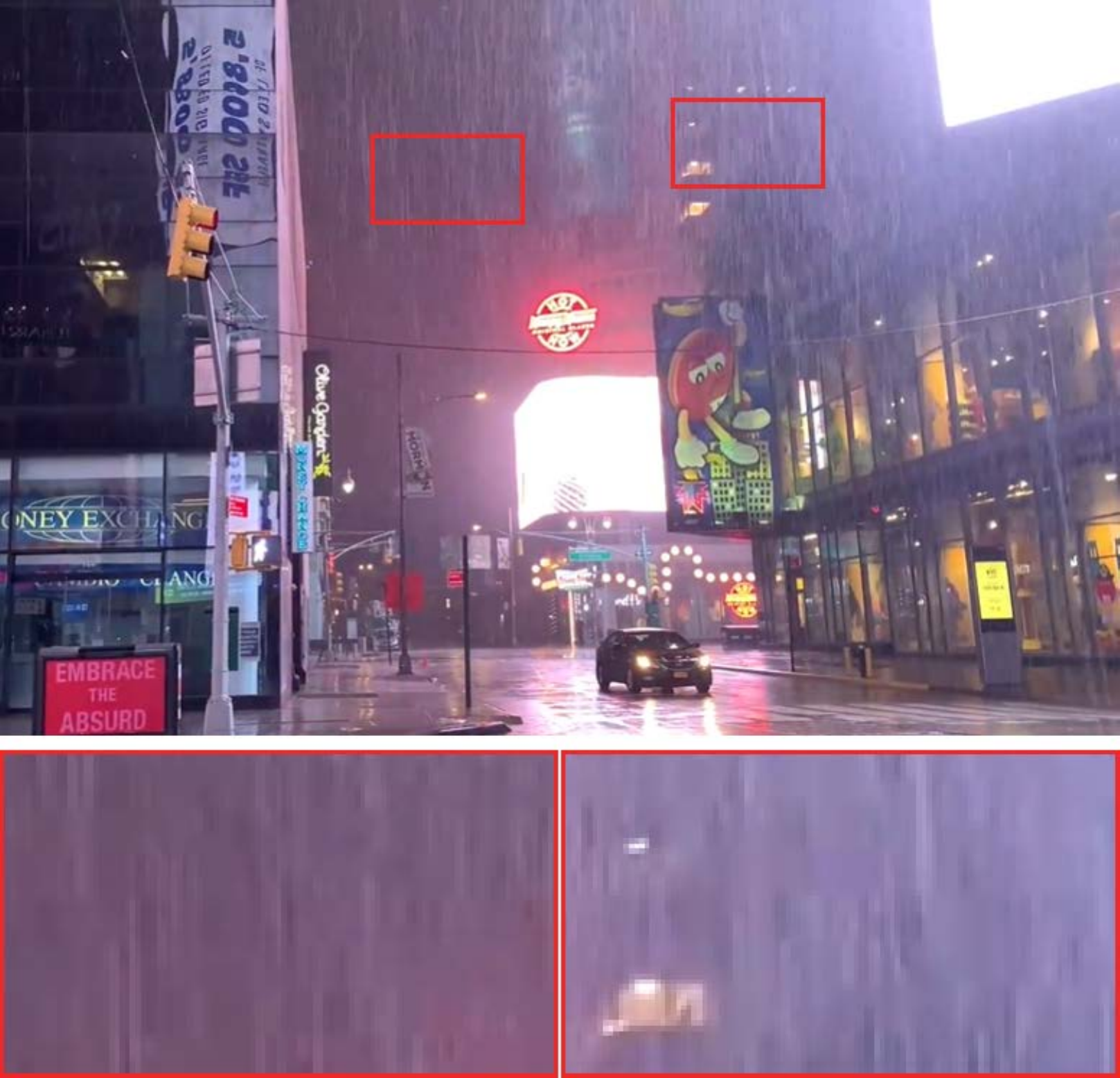}
			\caption{IDT}
		\end{subfigure}	
		\begin{subfigure}[t]{0.196\textwidth}
			\centering
			\includegraphics[width=\textwidth]{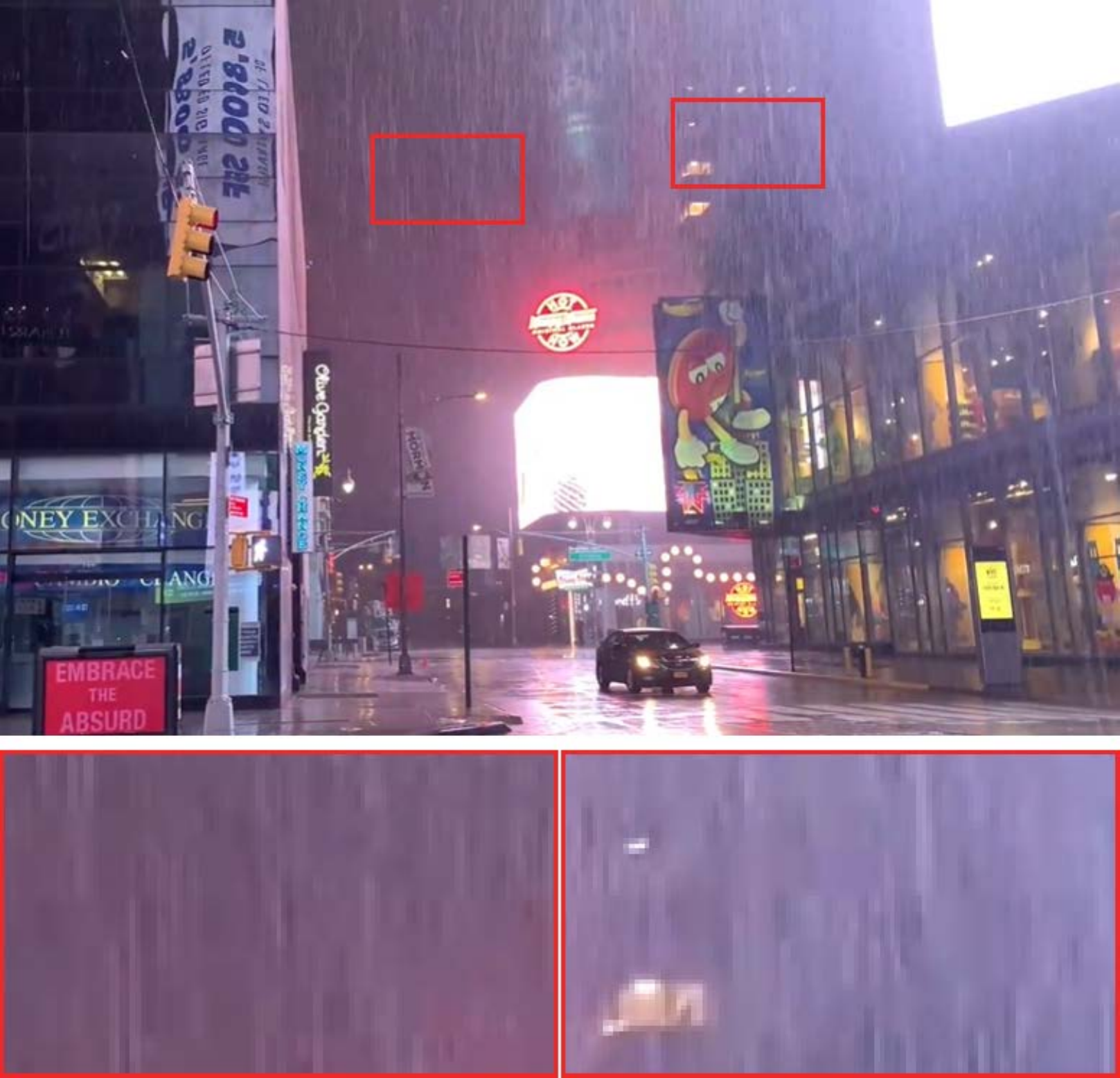}
			\caption{DRSformer}
		\end{subfigure}	
		\caption{Visual quality comparison of new benchmark track on the RE-RAIN dataset. Please zoom in the figures for better view of the rain removal and detail recovery.}
		\label{fig9}
	\end{figure*}

	Table \ref{table7} demonstrates quantitative results of all methods on the HQ-RAIN and RE-RAIN benchmarks.
	We note that DRSformer \cite{chen2023learning} and Restormer \cite{zamir2022restormer} obtain the top two performance on the synthetic HQ-RAIN benchmark. However, they do not perform well on the real-world RE-RAIN benchmark, which means that these two methods do generalize well on real-world applications.
	In addition, among the CNN-based methods, HINet \cite{chen2021hinet} achieves the competitive results on the HQ-RAIN.
	As some specific domain knowledge is beneficial for real-world rain removal \cite{wang2020model}, intergrading some specific domain knowledge into the model designs improves the generalization ability of deep learning.
	The results on the RE-RAIN demonstrate that most CNN-based methods have better generalization performance than Transformer-based methods as most CNN-based approaches integrate domain knowledge (\emph{e.g.}, rain structure and kernel) to better guide the model designs.
	Figures \ref{fig8} and \ref{fig9} show several visual comparisons.
	The DRSformer \cite{chen2023learning} performs better than the evaluated methods, which keeps consistent with the above quantitative results.
	Figure \ref{fig9} show that LPNet \cite{fu2019lightweight} generates better derained images.
	Based on above evaluations, deraining real-world images is challenging and worthy further studied.
	
	\begin{table*}[t]
		\centering
		\caption{Quantitative comparisons of mAP and mAR under different IoU thresholds of downstream object detection task. Top $1_{s t}$ and $2_{nd}$ results are marked in \textcolor{red}{red} and \textcolor{blue}{blue} respectively.}
		\resizebox{1.0\textwidth}{!}{
			\begin{tabular}{c|cccccccccc}
				\hlinew{1.0pt}
				\textbf{Methods} & \begin{tabular}[c]{@{}c@{}}LPNet\\ {\cite{fu2019lightweight}}\end{tabular} & \begin{tabular}[c]{@{}c@{}}PReNet\\ {\cite{ren2019progressive}}\end{tabular} & \begin{tabular}[c]{@{}c@{}}JORDER-E\\ {\cite{yang2019joint}}\end{tabular} & \begin{tabular}[c]{@{}c@{}}RCDNet\\ {\cite{wang2020model} }\end{tabular} & \begin{tabular}[c]{@{}c@{}}HINet\\ {\cite{chen2021hinet} }\end{tabular} & \begin{tabular}[c]{@{}c@{}}SPDNet\\ {\cite{yi2021structure}  }\end{tabular} & \begin{tabular}[c]{@{}c@{}}Uformer\\ {\cite{wang2022uformer}  }\end{tabular} & \begin{tabular}[c]{@{}c@{}}Restormer\\ {\cite{zamir2022restormer}  }\end{tabular} & \begin{tabular}[c]{@{}c@{}}IDT\\ {\cite{xiao2022image}  }\end{tabular} & \begin{tabular}[c]{@{}c@{}}DRSformer\\ {\cite{chen2023learning}   }\end{tabular} \\ \hline
				
				& \multicolumn{10}{c}{\textbf{IoU threshold=0.5}}                                                    \\
				mAP $\uparrow$            & 0.480 & 0.507 & 0.542   & 0.556 & \textcolor{blue}{0.566} & 0.469 & \textcolor{red}{0.573}  & 0.499    & 0.541 & 0.556    \\
				mAR $\uparrow$           & 0.704 & 0.699 & 0.698   & \textcolor{red}{0.739} & 0.732 & 0.657 & \textcolor{blue}{0.736}  & 0.687    & \textcolor{blue}{0.736} & 0.735    \\	\hline
				
				& \multicolumn{10}{c}{\textbf{IoU threshold=0.6}}                                                    \\
				mAP $\uparrow$            & 0.333 & 0.375 & \textcolor{blue}{0.429}   & \textcolor{red}{0.430} & 0.379 & 0.381 & 0.425  & 0.426    & 0.388 & 0.397    \\
				mAR $\uparrow$           & 0.560 & 0.469 & 0.575   & \textcolor{blue}{0.617} & 0.507 & 0.471 & 0.606  & \textcolor{red}{0.641}    & 0.474 & 0.482    \\	\hline
				
				& \multicolumn{10}{c}{\textbf{IoU threshold=0.7}}                                                    \\
				mAP $\uparrow$            & 0.249 & 0.291 & 0.285   & \textcolor{red}{0.306} & 0.299 & 0.291 & 0.300  & \textcolor{blue}{0.305}    & 0.301 & 0.301    \\
				mAR $\uparrow$           & 0.363 & 0.364 & 0.357   & \textcolor{blue}{0.374} & 0.368 & 0.359 & 0.362  & \textcolor{red}{0.375}    & 0.367 & 0.372    \\	
				
				\hlinew{1.0pt}
			\end{tabular}
		}
		\label{table8}		
	\end{table*}
	
	\subsection{Computational Complexity}
	The computational complexity is also one of the important factor for image deaining methods. Table \ref{table7} shows the computational complexity of different methods, including trainable model parameters, FLOPs and running time on a $256 \times 256$ image.
	LPNet \cite{fu2019lightweight} requires lower computational complexity as it develops a lightweight pyramid network using domain-specific knowledge to simplify the learning process.
	In contrast, the model size of HINet \cite{chen2021hinet} is extremely large, reaching 88.67M, which limits its usage in practical applications.
	The FLOPs of DRSformer \cite{chen2023learning} is relatively higher as it involves the computation of the top-\emph{k} self-attention.
	For inference time, most existing methods, especially Transformer-based approaches, are not efficient.
	Thus, how to develop efficient yet effective method is still worthy investigation.
	
	\begin{figure*}[!t]
		\centering 	
		\begin{subfigure}[t]{0.196\textwidth}
			\centering
			\includegraphics[width=\textwidth]{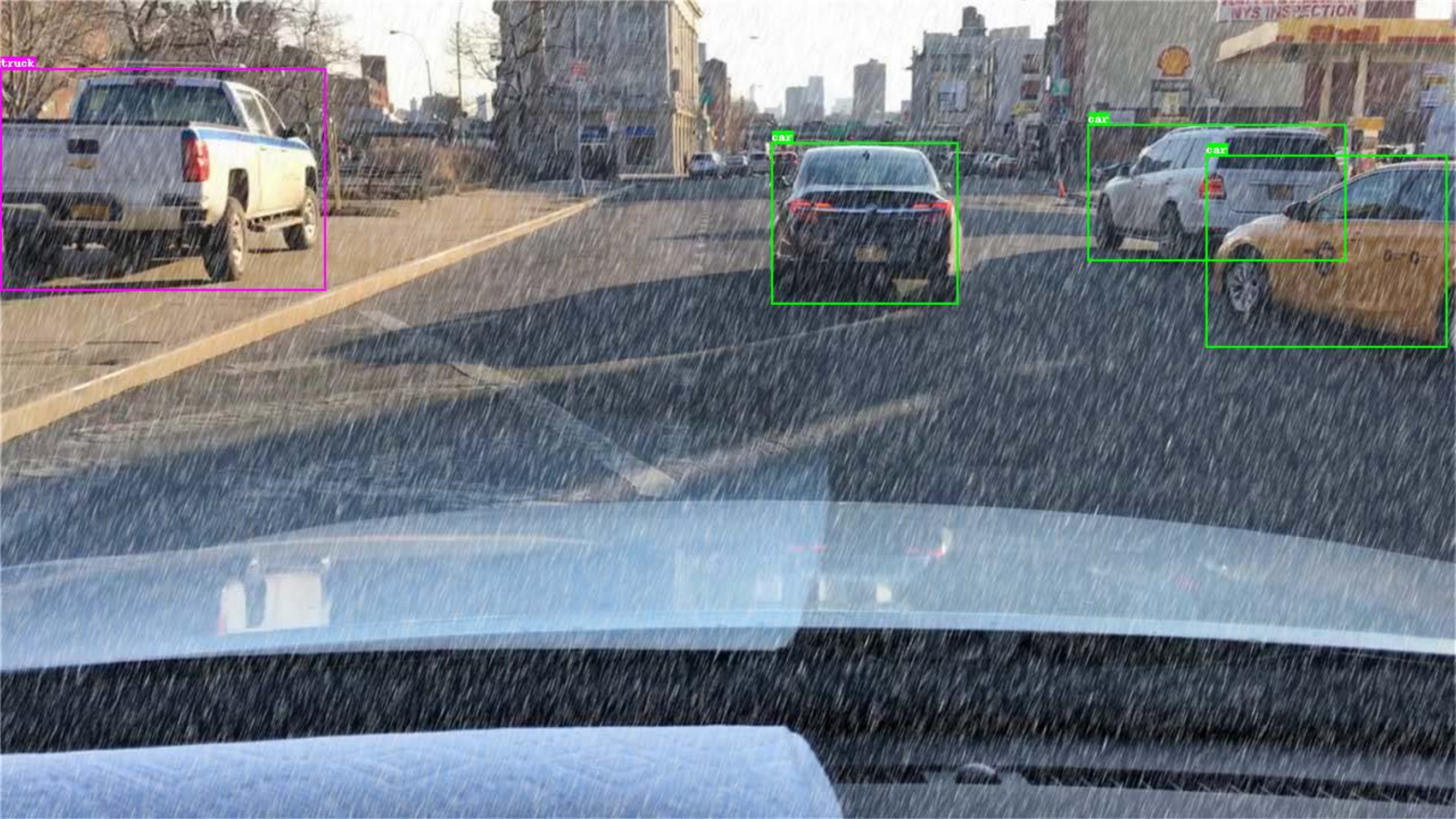}
			\caption{Rainy Input}
		\end{subfigure}
		\begin{subfigure}[t]{0.196\textwidth}
			\centering
			\includegraphics[width=\textwidth]{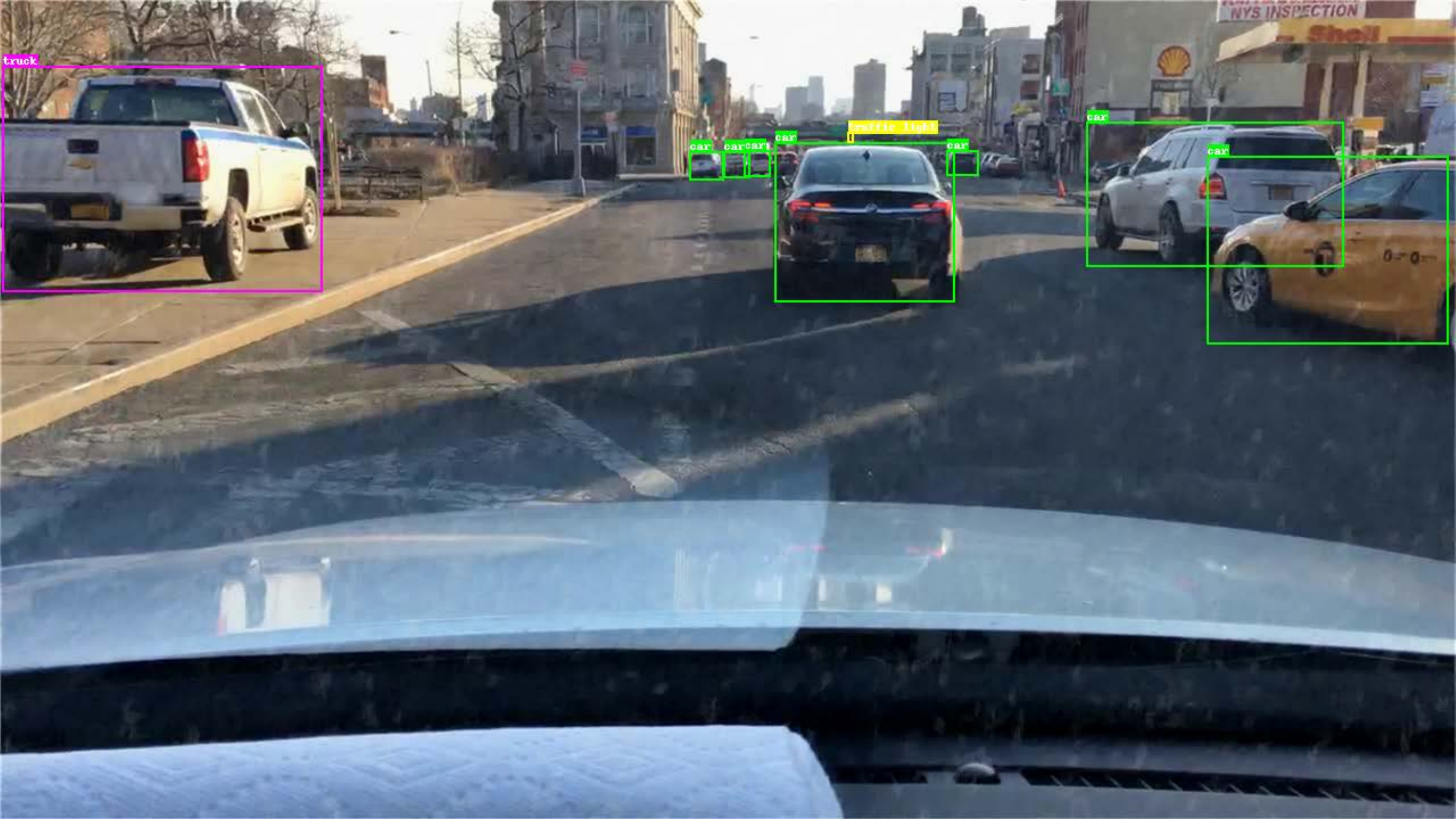}
			\caption{PReNet}
		\end{subfigure}
		\begin{subfigure}[t]{0.196\textwidth}
			\centering
			\includegraphics[width=\textwidth]{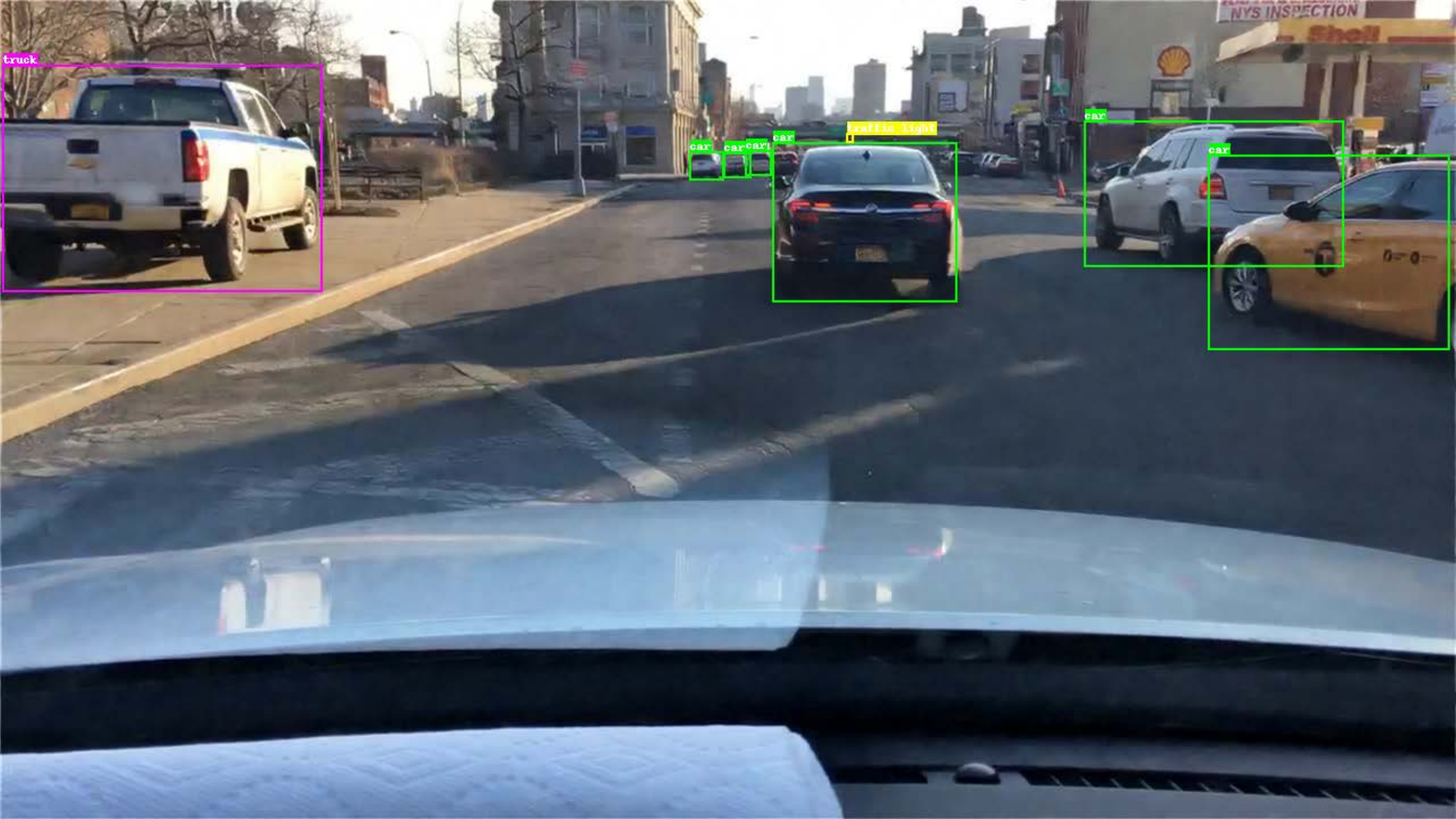}
			\caption{JORDER-E}
		\end{subfigure}	
		\begin{subfigure}[t]{0.196\textwidth}
			\centering
			\includegraphics[width=\textwidth]{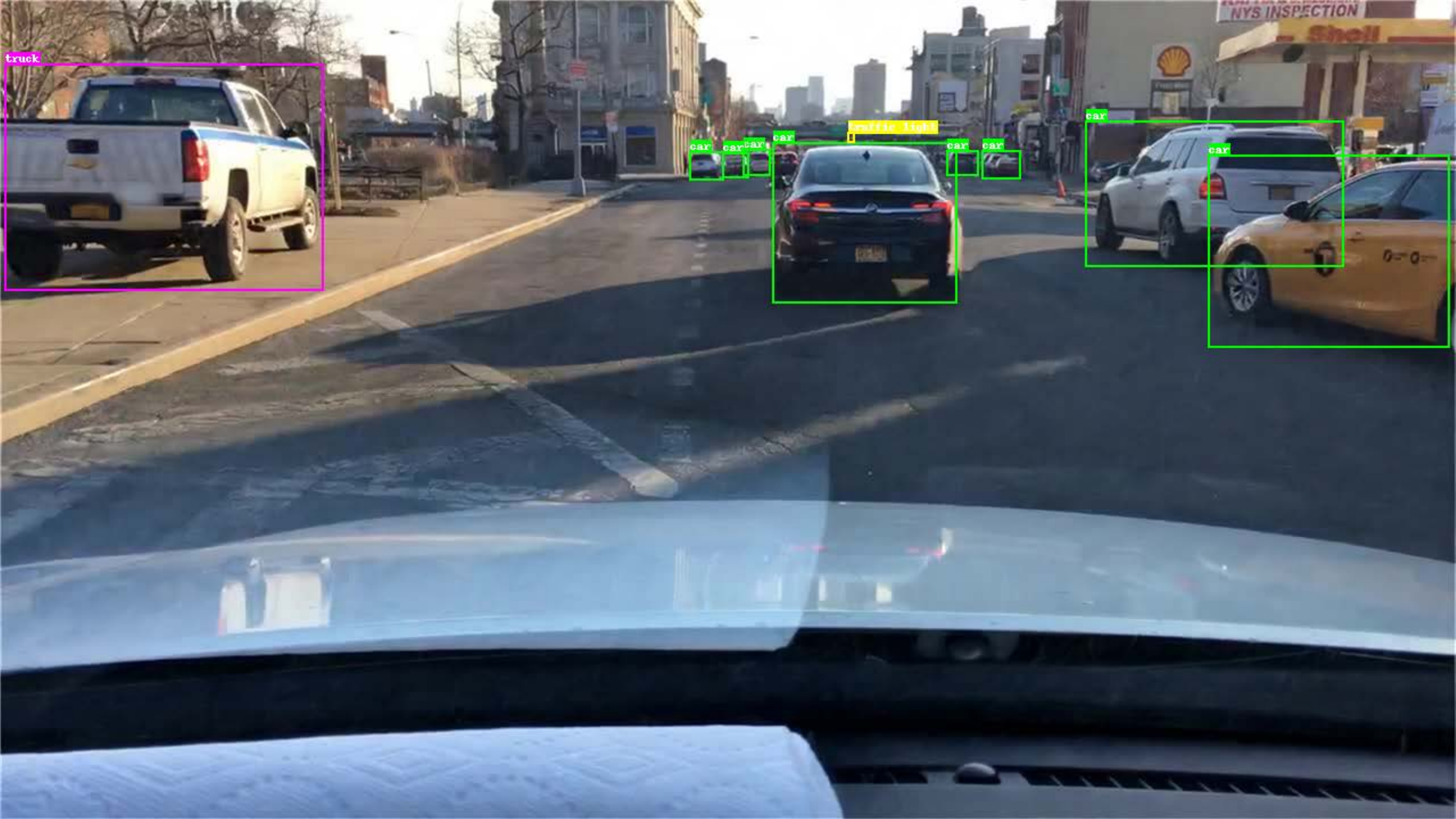}
			\caption{RCDNet}
		\end{subfigure}	
		\begin{subfigure}[t]{0.196\textwidth}
			\centering
			\includegraphics[width=\textwidth]{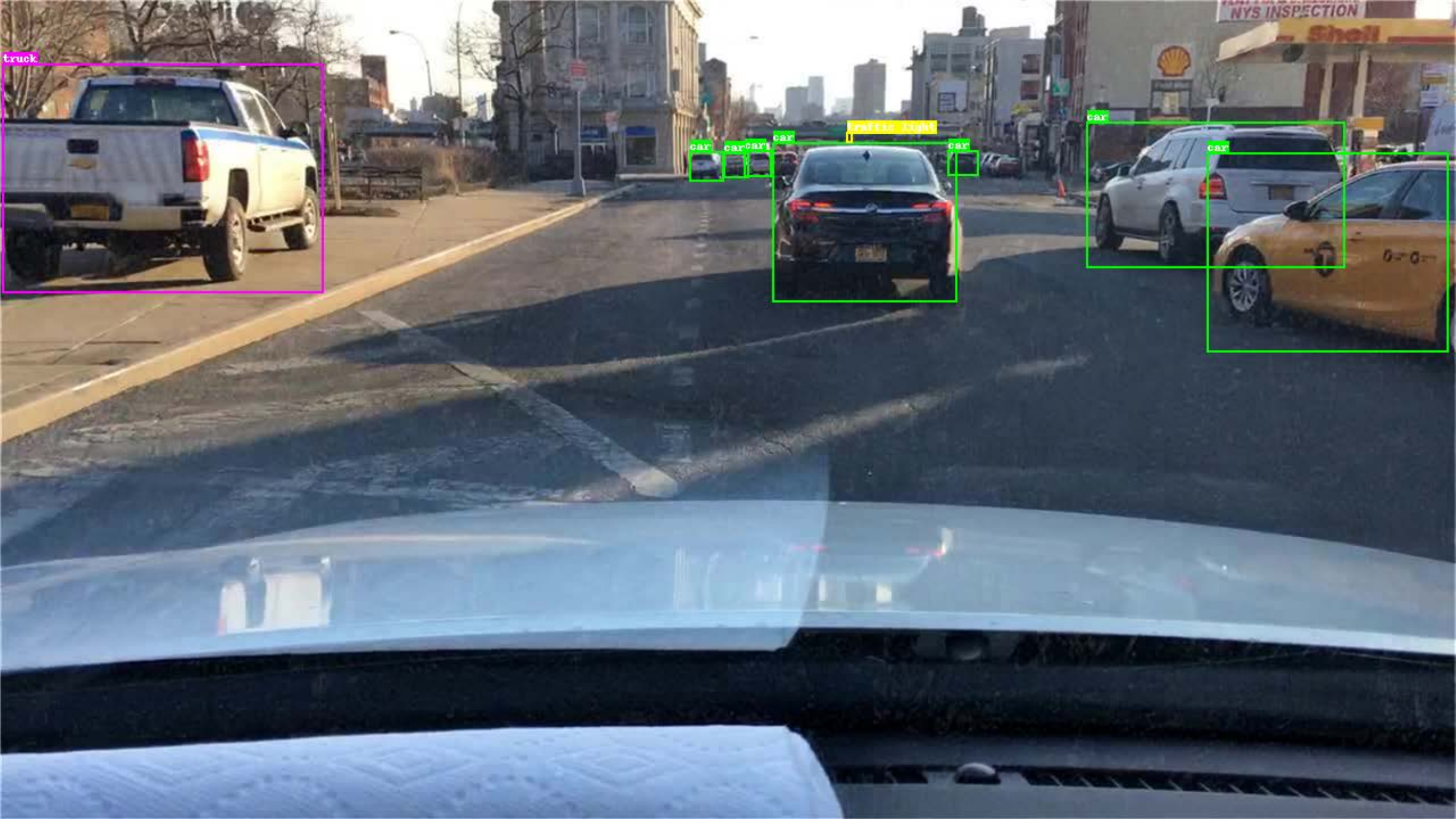}
			\caption{HINet}
		\end{subfigure}	
		\\
		\begin{subfigure}[t]{0.196\textwidth}
			\centering
			\includegraphics[width=\textwidth]{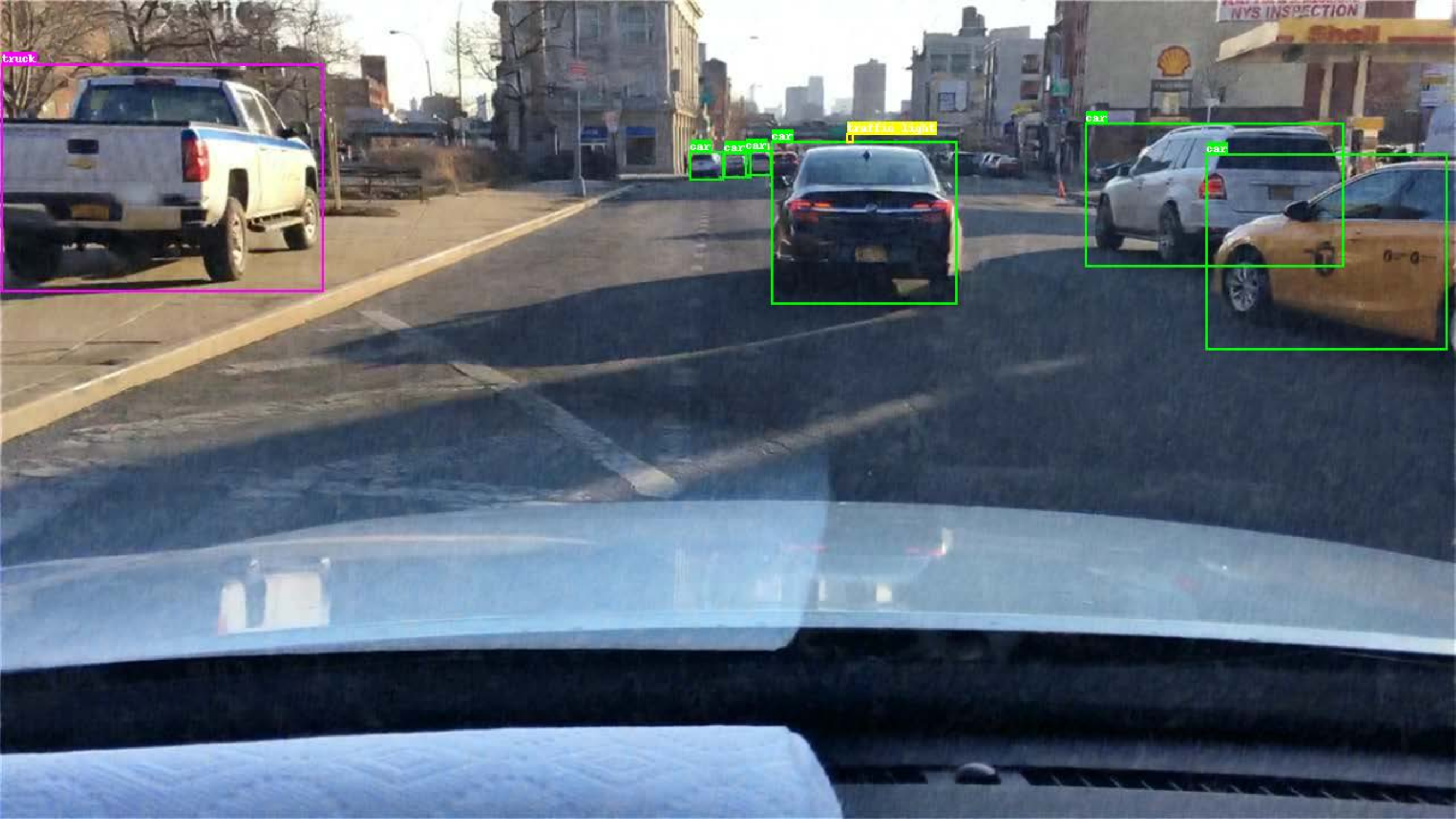}
			\caption{SPDNet}
		\end{subfigure}
		\begin{subfigure}[t]{0.196\textwidth}
			\centering
			\includegraphics[width=\textwidth]{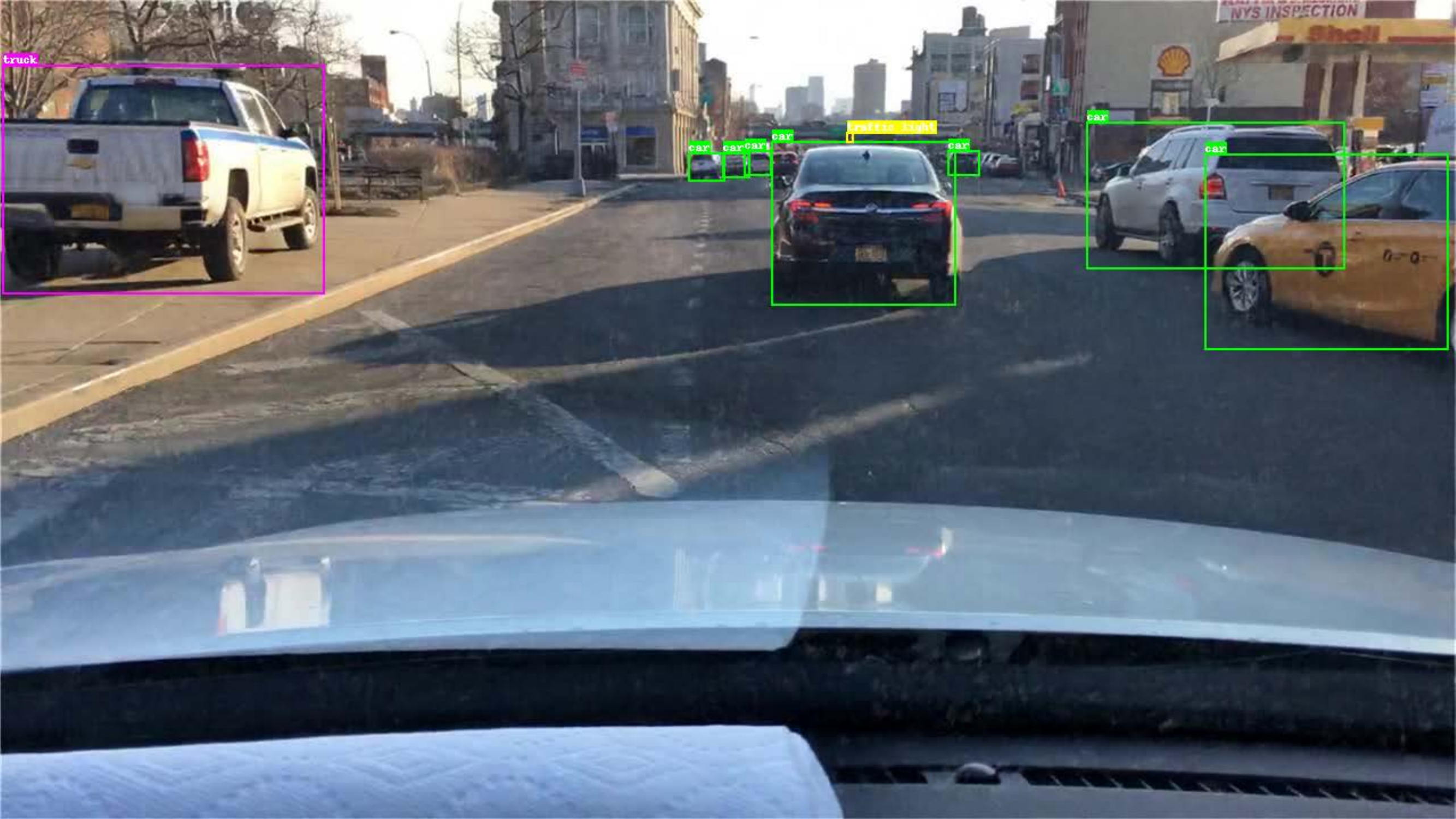}
			\caption{Uformer}
		\end{subfigure}
		\begin{subfigure}[t]{0.196\textwidth}
			\centering
			\includegraphics[width=\textwidth]{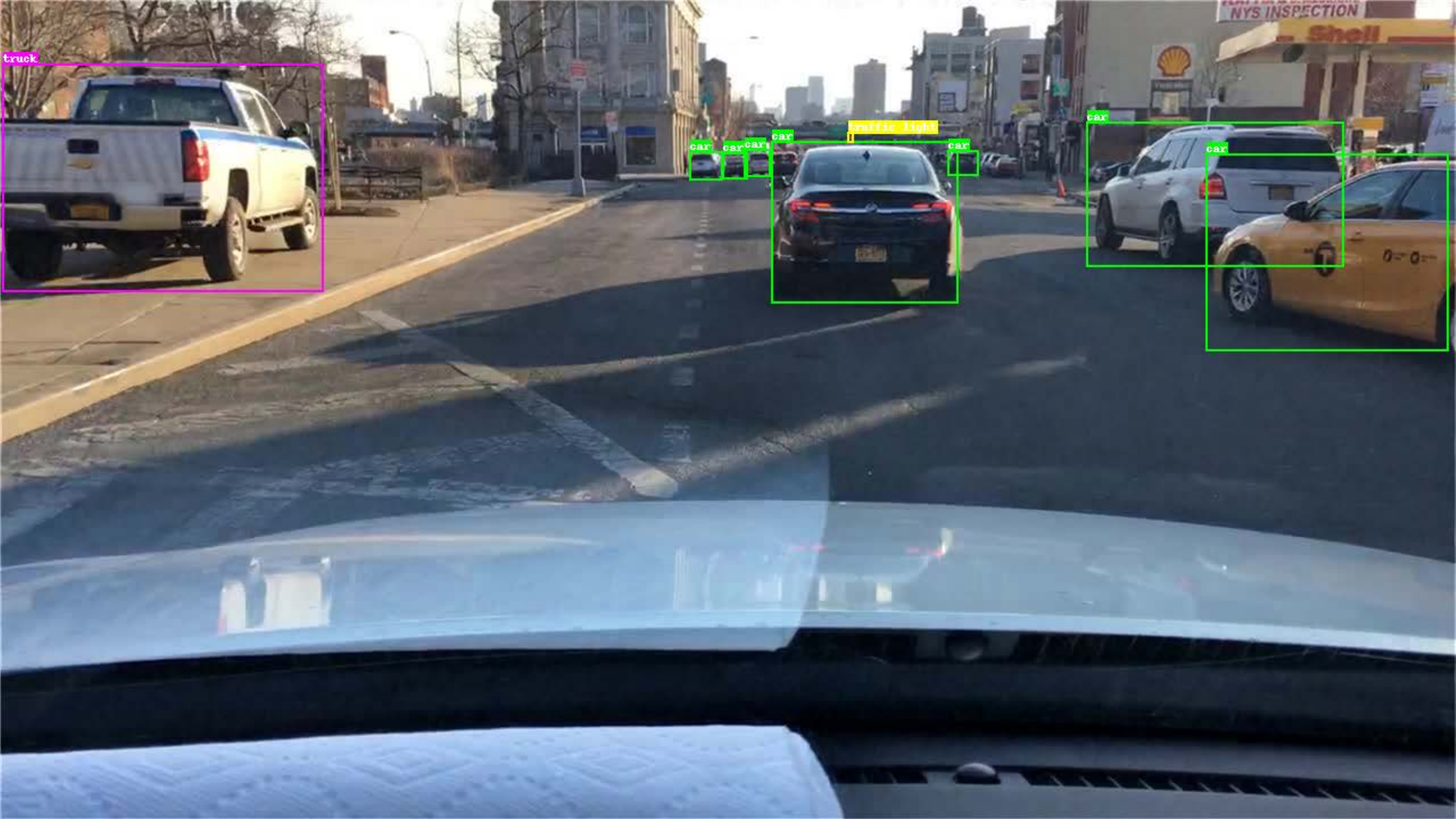}
			\caption{Restormer}
		\end{subfigure}
		\begin{subfigure}[t]{0.196\textwidth}
			\centering
			\includegraphics[width=\textwidth]{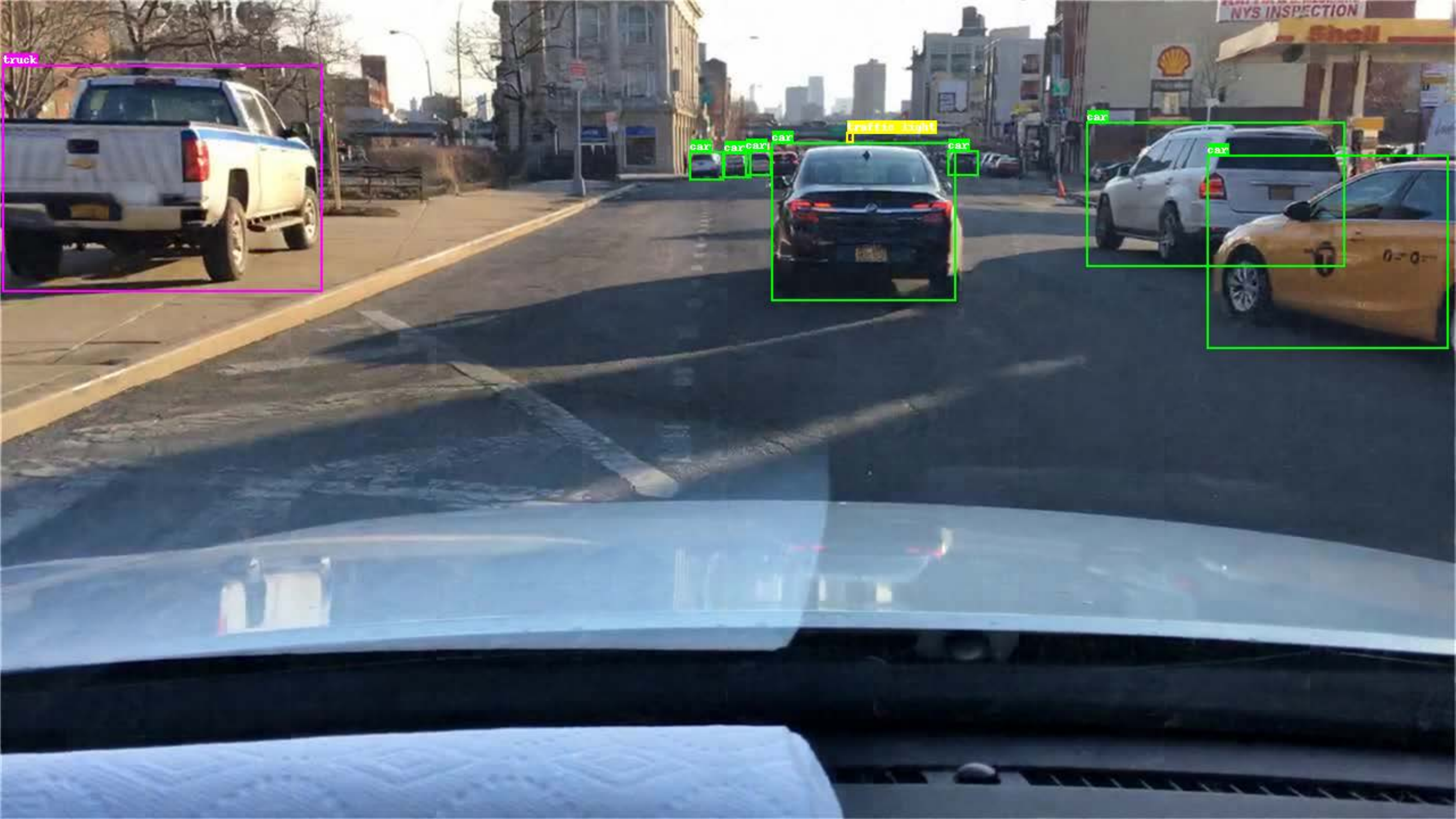}
			\caption{IDT}
		\end{subfigure}	
		\begin{subfigure}[t]{0.196\textwidth}
			\centering
			\includegraphics[width=\textwidth]{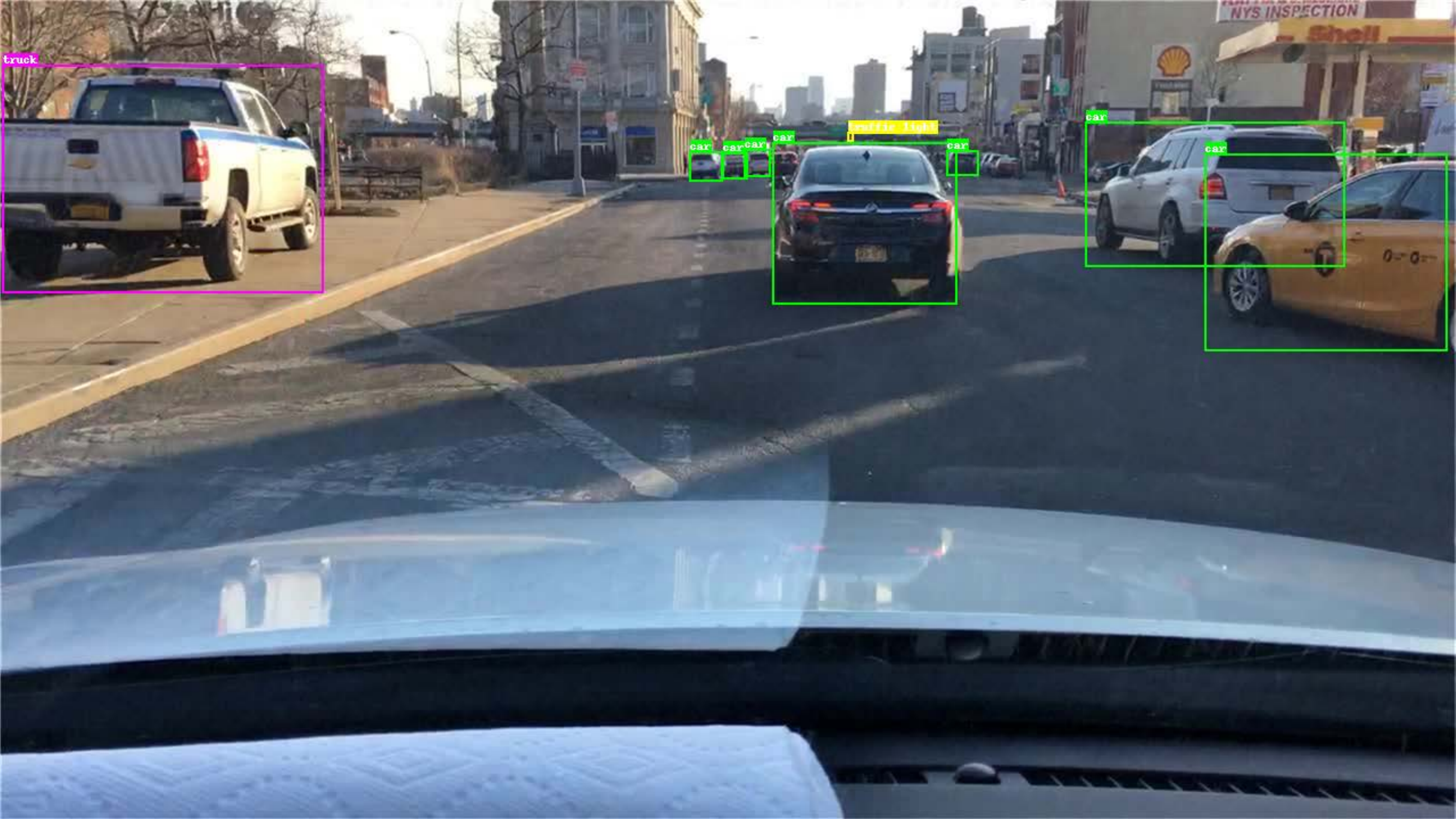}
			\caption{DRSformer}
		\end{subfigure}	
		\caption{Visual comparison of joint image deraining and object detection. Please zoom in the figures for better view of the rain removal and object detection.}
		\label{fig10}
	\end{figure*}
	
	\begin{figure*}[t]
		\centering
		\begin{subfigure}[t]{0.27\textwidth}
			\centering
			\includegraphics[width=\textwidth]{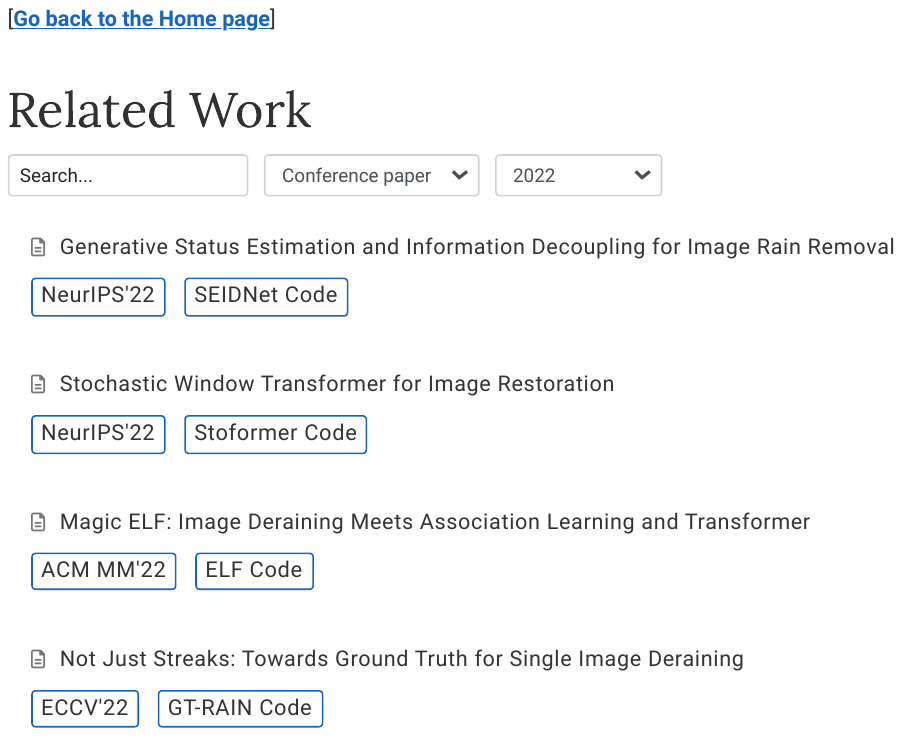}
			\caption{Survey Page}
		\end{subfigure}
		\begin{subfigure}[t]{0.33\textwidth}
			\centering
			\includegraphics[width=\textwidth]{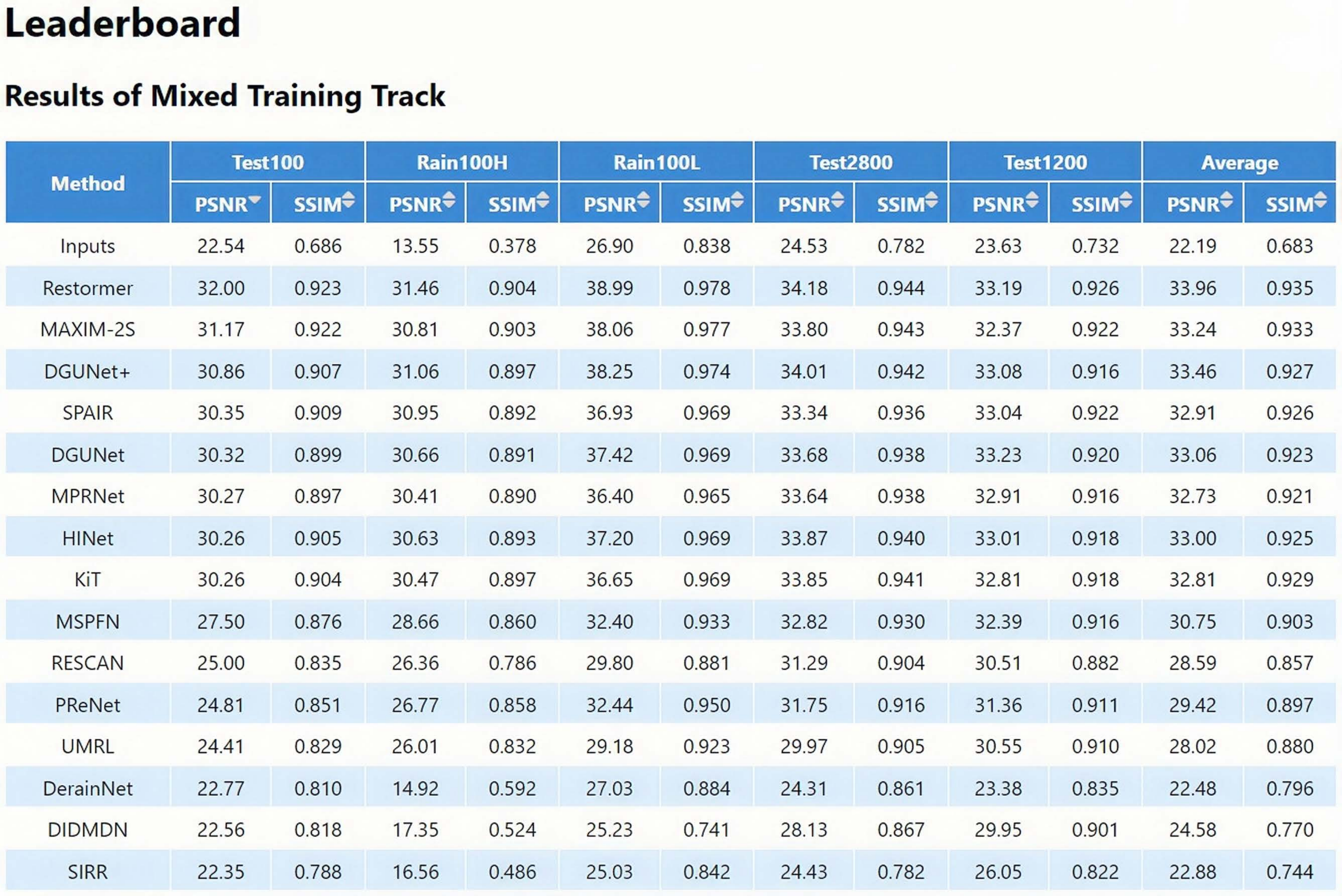}
			\caption{Leaderboard Page}
		\end{subfigure}
		\begin{subfigure}[t]{0.37\textwidth}
			\centering
			\includegraphics[width=\textwidth]{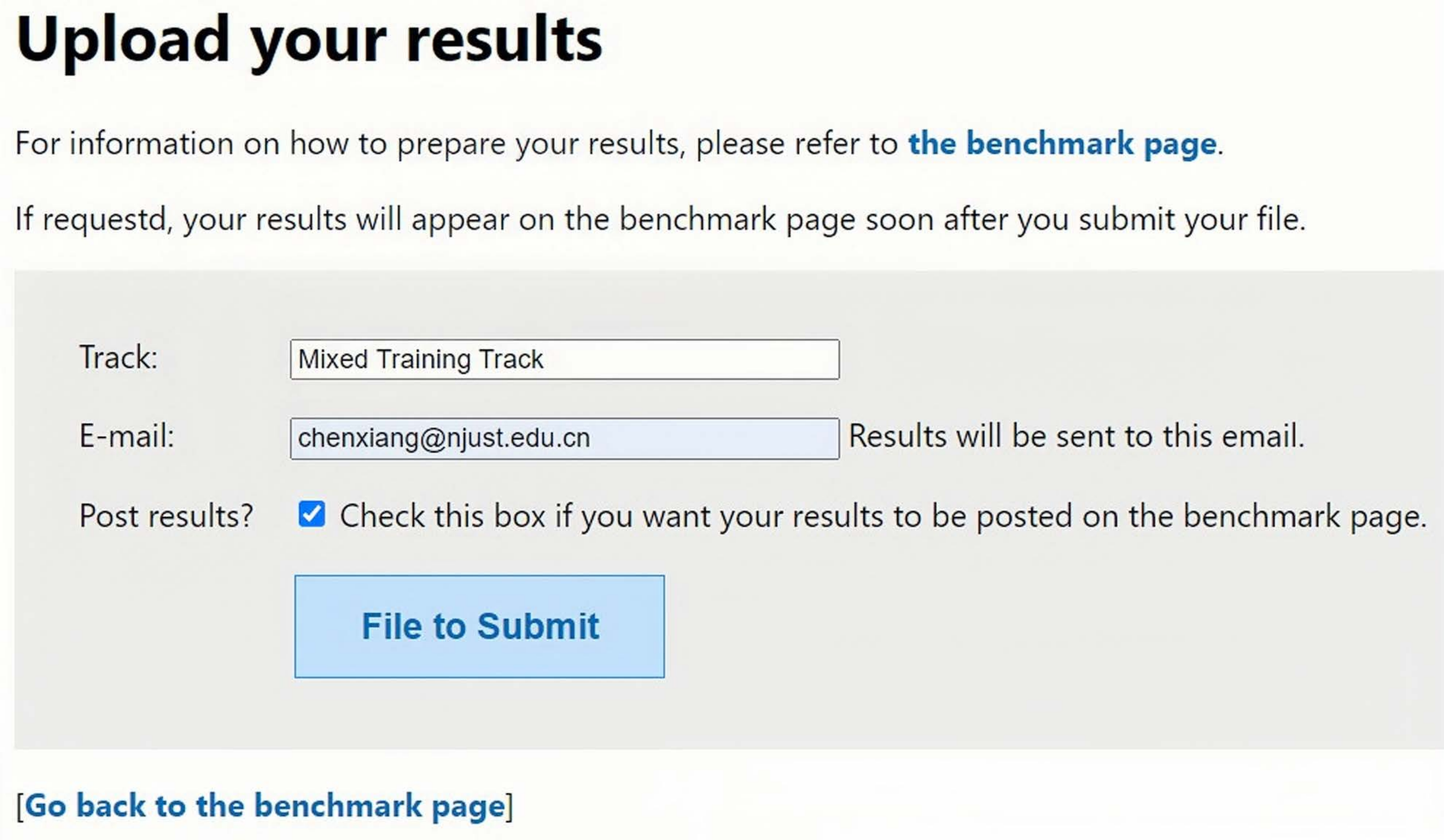}
			\caption{Submit Page}
		\end{subfigure}
		\caption{Screenshots of the developed online evaluation platform. Some design styles refer to the SIDD website \cite{abdelhamed2018high}.}
		\label{fig11}
	\end{figure*}
	
	\subsection{Application-Based Evaluation}
	To investigate whether the image deraining process benefits downstream vision-based applications, \emph{e.g.}, object detection, we apply the popular object detection pre-trained model (YOLOv3 \cite{redmon2018yolov3}) to evaluate the deraining results.
	Here, we create a high-quality version of BDD350 (joint image deraining and object detection) \cite{jiang2022magic}.
	Following \cite{li2021low}, we calculate the mean Average Precision (mAP) and mean Average Recall (mAR) under different Intersection of Union (IoU) thresholds using the available evaluation tool\footnote[2]{https://github.com/PaddlePaddle/PaddleDetection}.
	In addition, visual comparisons are presented in Figure \ref{fig10}.
	As one can see, the detection precision of the deraining results by all algorithms shows different degrees of improvement over that of original rainy inputs.
	From Table \ref{table8}, RCDNet \cite{wang2020model} surprisingly achieves relatively robust performance on the object detection.
	This may be because this method can better protect the image semantic information by using an interpretable rain dictionary model.
	However, some recent methods (\emph{e.g.}, IDT \cite{xiao2022image} and DRSformer \cite{chen2023learning}) exhibit rapid performance degradation under higher threshold setting, suggesting that there are still rooms for improvement.
	
	\subsection{Online Evaluation Platform}
	To facilitate the large-scale evaluation and tracking of the latest deraining technologies for general users, we develop an online platform\footnote[3]{The platform is available at \url{http://www.deraining.tech/}} for current research development of image deraining.
	Figure \ref{fig11} shows some screenshots of the proposed online evaluation platform.
	The proposed evaluation platform provides a comprehensive repository, including direct links and source codes for 100 representative methods, to continuously track recent developments in this fast-advancing field.
	It also provides evaluations of several popular and representative image deraining methods to better facilitate the following researches.
	In addition, the evaluation platform allows users submit the derained results and can provide direct comparisons with existing methods online.
	More visual results on the mixed training track, independent training track and new benchmark can also be downloaded on our platform.
	The proposed online evaluation platform is updated regularly to report new technology trends and benchmark results.

	\section{Future Research Prospects}
	\label{sec:challengs}
	Single image deraining has undergone rapid development in recent years, thanks to the incorporation of deep learning techniques. Researchers have made significant contributions by proposing effective models, promoting the optimization process, and establishing abundant datasets to advance the field. Despite achieving promising results, there are still several opportunities to explore in this field including how to model the degradation process, effective deraining models, and its potential applications for the following image analysis tasks.
	
	\subsection{Degradation Progress}
	The degradation progress models the generation of the rainy images. Thus, how to better model the degradation progress is critical for image deraining in the deep learning era.
	As the real degradation is complex, the image synthesization based on the linear composition model, \emph{e.g.},~\eqref{Eq1}, may not be able to model the real degradation. The methods trained on the datasets based on the linear composition model may not generalize well on real-world applications due to the domain gap of the data distributions.

	Although several approaches design kinds of hardwares to capture real-world rainy images, the diversity of images and complexity of rainy environments (\emph{e.g.}, low-light environment, hazy environment, and so on) are not sufficient to cover the realistic scenarios.
	Therefore, how to describe the degradation of clear image in rainy settings including the rainy streak distributions, rainy image properties, the laws of physics of rain and so on is a great of need but still challenging.
	
	\subsection{Model Development}
	{\flushleft\textbf{High-quality Image Deraining Models}.}
	As image deraining aims to recover realistic clear images, how to develop an effective model that describes the properties of clear image is important.
	Below we will present some promising directions for network improvements.
	(1) Integrating domain knowledge into model development. For instance, by leveraging semantic segmentation, we can identify specific regions of interest in the image such as sky or roads \cite{kirillov2023segment}. This may allow us to use unequal information from different regions to guide the model in removing rain and avoiding unnecessary information loss.
	(2) Modeling the distributions of clear images. We can use image generation approaches (\emph{e.g.},  VQGAN \cite{esser2021taming} or diffusion models \cite{ho2020denoising}) to facilitate the high-quality image restoration.
	(3) Exploring external information of clear images or distilling the knowledge from large pre-trained models to facilitate the rain removal. By prompt learning or knowledge distillation, the model can benefit from the learned representations and generalization capabilities. This allows the model to effectively utilize the specific prompts or knowledge to better understand and remove rain from images.
	
	{\flushleft\textbf{Efficient Image Deraining Models}.}
	Developing efficient image deraining models is important due to resource-limited devices in most real-world applications.
	Facing this challenge, we can adopt pruning \cite{kim2022learned}, dimension reduction and feature reuse \cite{li2023deep} to remove unnecessary connections or parameters from the network.
	In addition, knowledge distillation \cite{gou2021knowledge} can also be used to facilitate the transfer of knowledge from a complex model to a compact one, effectively compressing the model while maintaining performance.
	We can also explore neural architecture search \cite{elsken2019neural} to learn compact models by introducing various search strategies to sample architectures from the defined search space.
	
	\subsection{Potential Applications}	
	{\flushleft\textbf{Relations of Image Deraining and Other Tasks}.}
	Most existing methods mainly focuses on the individual image deraining task and usually use the image deraining task as a pre- or post-processing step for other tasks.
	Although the development of these fields are independent, different tasks can have a constructive influence in promoting each other, such as Deblur-YOLO \cite{zheng2021deblur} and YOLO-in-the-Dark \cite{sasagawa2020yolo}. We call for closer collaboration across low-level and high-level communities.
		
	\section{Concluding Remarks}
	\label{sec:conclusion}
	We have provided comprehensive overview of the existing image deraining methods.
	We first analyze the research progress, discuss the limitations, and provide empirical evaluations of image deraining methods.
	To better evaluate existing methods, we propose develop a rainy image generation approach based on an image harmonization approach and develop a realistic image deraining dataset.
	We have established two parallel tracks (\emph{i.e.}, mixed training track and independent training track) to facilitate effective and fair performance comparison among different methods.
	We carry out qualitative and quantitative analyses to investigate the open challenges and provide the future prospects.
	To provide fair evaluation protocols, we build an online platform to better evaluate the deraining methods and facilitate the following researches.
	The online evaluation platform is publicly available and will be regularly updated.

	\ifCLASSOPTIONcompsoc
	
	\ifCLASSOPTIONcaptionsoff
	\newpage
	\fi
	
	\bibliographystyle{IEEEtran}
	\bibliography{ref}

\end{document}